%% file: paper.tex
\documentclass[3p]{elsarticle}

\usepackage{amsmath, amssymb, amsthm}
\usepackage{mathtools}
\usepackage{tikz}
\usepackage{pgfplots}
\pgfplotsset{compat=1.18}
\usepgfplotslibrary{groupplots}
\usepackage{booktabs}
\usepackage{array}
\usepackage{enumitem}
\usepackage{float}
\usepackage{placeins}

\usepackage[colorlinks=true,linkcolor=blue,citecolor=blue,urlcolor=blue]{hyperref}
\usepackage{url}
\makeatletter
\let\ps@pprintTitle\ps@plain
\makeatother

\newtheorem{proposition}{Proposition}
\theoremstyle{definition}
\newtheorem{definition}{Definition}

\input{numbers.tex}
\input{figures.tex}

\begin{document}
\hypersetup{pdftitle={When does off-block Hessian curvature help an optimization step?}}
\raggedbottom

\begin{frontmatter}

\title{When does off-block Hessian curvature help an optimization step?\tnoteref{stateassignment}}
\tnotetext[stateassignment]{This work was supported by the state assignment (project FSER-2025-0004).}

\author[inst1]{Maxim Bolshim\corref{cor1}}
\cortext[cor1]{Corresponding author}
\ead{mabolshim@itmo.ru}
\ead[url]{https://orcid.org/0009-0009-3037-9852}
\author[inst1]{Alexander Kugaevskikh}
\ead{avkugaevskikh@itmo.ru}
\ead[url]{https://orcid.org/0000-0002-6676-0518}
\affiliation[inst1]{organization={ITMO University},
            addressline={49 Kronverksky Prospekt, Building A},
            city={Saint Petersburg},
            postcode={197101},
            country={Russia}}

\begin{abstract}
Testing an off-block Hessian correction alone can discard a step that lowers minibatch loss. We measure both the missed improvement and the gain from rescaling the remaining correction.
A reference based on diagonal-block norms distinguishes changes in the diagonal profile during training from changes in relative off-block mass. This reference sharply bounds the off-block fraction for positive semidefinite matrices. Regrouping fixed ResNet-18 weights from stages to layers takes the estimated fraction from below to above its partition-specific bound in all six CIFAR-100 comparisons under a conditional probe screen.
At nine ResNet-50 checkpoints on Stanford Cars, we measure the off-block contribution to the linearized gradient change along a K-FAC step. At four base-step endpoints, predicted and measured slopes along the original correction have opposite signs. Each subspace draw (two per checkpoint) supplies one Hessian-informed map for the off-block residual, the remainder and their sum. At unit scale before acceptance, the off-block correction alone worsens the base in 17/18 comparisons; the combination beats the remainder in ten, including nine of those 17. At one checkpoint only the combination improves on the base in both draws. Matching correction norms reverses seven of the ten preferences, including both at that checkpoint.
\end{abstract}

\begin{keyword}
  second-order optimization \sep block-diagonal preconditioning \sep K-FAC \sep Hessian block structure \sep quadratic model validity \sep curvature diagnostics
\end{keyword}

\end{frontmatter}

\section{Introduction}
\label{sec:intro}

Should an off-block correction be discarded if it raises minibatch loss on
its own? The same component may help when added to the rest of a correction,
even with the base step and minibatch held fixed. This question arises in
block-diagonal curvature models, which make second-order methods tractable
by retaining interactions within parameter groups and omitting interactions
between groups. A Kronecker-factored approximate curvature (K-FAC) step provides the base
candidate in our comparison. Measuring the omitted operator describes the structural error; assessing a
correction also requires its action on the base step and the resulting loss.

To describe what the grouping omits, we measure the full parameter-Hessian
mass outside a specified partition's diagonal blocks. Its squared Frobenius
fraction is the squared relative error of the best block-diagonal
approximation to that measured operator. A size reference accounts for
unequal group sizes; a profile reference uses measured diagonal-block norms.
The latter supplies a sharp upper bound for the off-block fraction of a
positive semidefinite (PSD) matrix at fixed diagonal-block norms, not a
prediction of a useful step or a claim that the measured Hessian is PSD.
Comparing both references separates differences in diagonal profiles from
differences in aggregate off-block mass relative to those profiles. In our
CIFAR-100 measurements, ResNet-18 and its skip-free counterpart, Plain-18,
have nearly equal initial discarded fractions despite markedly different
within-group curvature profiles. These quantities refer to specified weights,
inputs, coordinates and partitions, not Fisher information lost by K-FAC.

In a separate ResNet-50/Stanford Cars experiment, we examine the discarded
operator's action on a base step rather than infer correction benefit from
its norm. The saved-state comparison holds weights, a base step and a
Hessian-informed correction map fixed while varying its input: off-group
Hessian action, the remaining model residual or both together. Evaluated
minibatch losses distinguish the off-group input's standalone effect from
its incremental effect when added to the remainder. The structural
comparisons likewise hold saved weights fixed when changing the grouping,
but do not establish which partition yields a useful correction.

Our central comparison is retrospective: on a saved step we can evaluate a
combination even when its off-group component raises loss on its own. This
distinguishes a combination that merely beats the remaining correction from
one that also improves on the base step and would be lost under standalone
screening. Separate training trajectories test whether the correction
improves training; a supporting study checks signed model--loss agreement
at sampled fractions of a step.

\section{Related work}
\label{sec:related}

\textbf{Block models and structural measurements.} K-FAC approximates the
Fisher in per-module Kronecker factors \citep{martens2015optimizing};
eigenvalue-corrected K-FAC (EKFAC)
corrects eigenvalue scaling in that basis \citep{george2018fast}. Shampoo
uses per-tensor factors \citep{gupta2018shampoo}, and structured or diagonal
variants further restrict within-group curvature
\citep{lin2024singd,yao2021adahessian,liu2023sophia}. K-FAC also admits an
adjacent-layer block-tridiagonal inverse-Fisher construction
\citep{martens2015optimizing}. Our larger measurement groups contain several
modules, so their parameter-Hessian projection error is not the Fisher
approximation error of these methods. Exact Hessian blocks and module-wise
curvature backpropagation have long been available on smaller or selected
structures \citep{bishop1992exact,botev2017practical,dangel2020backpack}.
Dong et al.~\citep{dong2025quantifying} account for dimensional imbalance in
full-Hessian block-norm ratios; Hong et al.~\citep{hong2025better} measure the
relative off-block generalized Gauss--Newton (GGN) norm for influence
functions and also study exact-Hessian comparisons.
Our size and profile references and same-weights grouping comparison describe
what the grouping omits; the finite-step comparison asks whether an off-block
residual component helps a combined correction.
Activation-space analysis is reported separately by Bolshim and
Kugaevskikh~\citep{bolshim2026interlayer}.

\textbf{Corrections and finite-step evidence.} Two-level K-FAC adds coarse
Fisher interactions \citep{tselepidis2020twolevel,koroko2023analysis}, but
reducing Fisher solve error does not guarantee training gains
\citep{koroko2023analysis,benzing2022gradient}. Fisher solve accuracy,
evaluated loss at a candidate step and training-trajectory gains are distinct
endpoints. Full Hessian-vector products
are accessible without building the matrix \citep{pearlmutter1994fast,
schraudolph2002fast,martens2010deep}; randomized probes estimate its block
norms \citep{avron2011randomized}. Damping and actual-versus-predicted
reduction are standard checks on local models
\citep{martens2010deep,martens2015optimizing,nocedal2006numerical}.
We use a Hessian-informed pairwise correction as an instrument for comparing
specific candidate steps, distinguishing the quadratic-model value from
evaluated minibatch loss and from outcomes on separate training trajectories.
The supporting validity study considers a different, sampled one-step question
(\ref{app:validity}). Additional spectral, transformer-specific and
multi-step comparisons relevant to the supporting results are in
\ref{app:related-context}.

\section{Method}
\label{sec:method}

\subsection{Setting}
\label{sec:setting}

Let $L(\boldsymbol{\theta})$ be the mean data loss on a fixed minibatch, in a specified
model mode. Image tasks use mean cross-entropy over images; decoders use mean
shifted next-token cross-entropy over the batch and all predicted positions.
Optimizer weight decay is excluded from this loss; model damping enters as
a separate quadratic term. \ref{app:protocol} gives inputs,
training settings and experimental units.

Partition the parameter vector into disjoint groups $(\boldsymbol{\theta}_v)_{v\in V}$ of
sizes $p_v$, summing to $P>0$. For residual networks, stage groups collect
the stem, four stages and head;
for transformers, groups follow the projections, multilayer perceptrons (MLPs),
embedding and head.
We measure the parameter second-order operator $\mathbf{H}_{\boldsymbol{\theta}}$ returned by double
backward under the implementation's selected automatic-differentiation (AD)
conventions. The complete group ownership, nonsmooth-point conditions and
coordinate dependence are in \ref{app:extended-methods}.
Write $\mathbf{H}_{\boldsymbol{\theta}_v,\boldsymbol{\theta}_w}$ for its $(v,w)$ block and
$\mathrm{bd}(\mathbf{H}_{\boldsymbol{\theta}})=\mathrm{blkdiag}(\mathbf{H}_{\boldsymbol{\theta}_v,\boldsymbol{\theta}_v})$.

\subsection{The discarded-coupling fraction}
\label{sec:phi}

\begin{definition}[Discarded-coupling fraction]
  \label{def:phi}
  For $\|\mathbf{H}_{\boldsymbol{\theta}}\|_F>0$,
  \begin{equation}\label{eq:phi}
    \Phi(\boldsymbol{\theta}):=\frac{\sum_{v\neq w}\|\mathbf{H}_{\boldsymbol{\theta}_v,\boldsymbol{\theta}_w}\|_F^2}
    {\|\mathbf{H}_{\boldsymbol{\theta}}\|_F^2}
    =1-\frac{\sum_{v}\|\mathbf{H}_{\boldsymbol{\theta}_v,\boldsymbol{\theta}_v}\|_F^2}{\|\mathbf{H}_{\boldsymbol{\theta}}\|_F^2}
    \in[0,1].
  \end{equation}
\end{definition}

For this partition, $\Phi$ is the squared relative Frobenius error of the best
block-diagonal approximation to $\mathbf{H}_{\boldsymbol{\theta}}$
(Proposition~\ref{prop:phi-projection}). Even with exact within-group blocks, all off-block terms contribute to this
error. A genuinely nested refinement leaves the exact fraction unchanged
or increases it. Block-norm
probes estimate it without forming the Hessian; \ref{app:extended-methods}
gives their costs, zero conventions and sampling qualifications.

\subsection{Two reference values}
\label{sec:references}

For comparisons across models and checkpoints, we use two references
computed from the same run.

\begin{definition}[Reference values]
  \label{def:references}
  With the group sizes $\mathbf{p}=(p_v)_{v\in V}$ and the diagonal block norms
  $a_v:=\|\mathbf{H}_{\boldsymbol{\theta}_v,\boldsymbol{\theta}_v}\|_F$, with $\sum_v a_v>0$,
  \begin{align}
    \Phi_{\mathrm{null}}(\mathbf{p})&:=1-\frac{\sum_vp_v^2}{\bigl(\sum_vp_v\bigr)^2},
    \label{eq:phi-null}\\
    \Phi_{\mathrm{sep}}(\mathbf{a})&:=1-\frac{\sum_va_v^2}{\bigl(\sum_va_v\bigr)^2}.
    \label{eq:phi-sep}
  \end{align}
\end{definition}

The size reference $\Phi_{\mathrm{null}}$ is the value obtained when each
$\|\mathbf{H}_{\boldsymbol{\theta}_v,\boldsymbol{\theta}_w}\|_F^2$ in \eqref{eq:phi} is replaced by its expectation
under entries exchangeable across the whole matrix, which is proportional to
$p_vp_w$; it depends on the group sizes alone, and for $n_b$ groups of equal size
it is $1-1/n_b$.
The profile reference $\Phi_{\mathrm{sep}}$ is the value $\Phi$ takes when every
off-diagonal block norm equals the geometric mean of the two diagonal ones,
$\|\mathbf{H}_{\boldsymbol{\theta}_v,\boldsymbol{\theta}_w}\|_F=\sqrt{a_va_w}$. It is not a null model: the $a_v$ are
measured, so $\Phi_{\mathrm{sep}}$ moves with the weights, and
$\Phi_{\mathrm{null}}$ is the special case in which $\mathbf{a}$ is proportional to $\mathbf{p}$.
For every nonnegative, nonzero profile $\mathbf{a}$ and positive group sizes,
$\Phi_{\mathrm{sep}}$ is the sharp upper envelope for positive semidefinite
$\mathbf{H}_{\boldsymbol{\theta}}$ with these diagonal-block norms, not fixed diagonal matrices;
a rank-one matrix attains it.

The numerator of $S(\boldsymbol{\theta}):=\bigl(\sum_va_v\bigr)^2/\|\mathbf{H}_{\boldsymbol{\theta}}\|_F^2$
uses the square of the sum of the diagonal-block norms; the retained
fraction $1-\Phi$ uses the sum of their squares. The two satisfy
\begin{equation}\label{eq:s-identity}
  1-\Phi(\boldsymbol{\theta}) = S(\boldsymbol{\theta})\,\bigl(1-\Phi_{\mathrm{sep}}(\mathbf{a})\bigr),
\end{equation}
and $S=1$ exactly when the off-block squared mass
$\sum_{v\neq w}\|\mathbf{H}_{\boldsymbol{\theta}_v,\boldsymbol{\theta}_w}\|_F^2$ equals the separable-profile value
$\sum_{v\neq w}a_va_w$. Off-block mass lies below this value when $S>1$ and
above it when $S<1$. Whereas $\Phi_{\mathrm{sep}}$ depends only on diagonal
amplitudes, $S$ compares aggregate off-block mass with the separable-profile value;
it does not describe allocation among pairs or directional action.
For a positive semidefinite $\mathbf{H}_{\boldsymbol{\theta}}$, aggregate off-block mass cannot
exceed the separable-profile value, so $S\ge1$ and
$\Phi\le\Phi_{\mathrm{sep}}$. Exceeding the exact envelope implies an
indefinite operator; an estimated crossing alone is not a spectral
certificate. Proof and qualifications are in \ref{app:extended-methods}.
Given group sizes and measured diagonal-block norms, evaluating either
reference needs no additional Hessian--vector products (HVPs); computing $S$
also uses the full-norm estimate already used for $\Phi$. This excludes the
cost of obtaining the norms and of optional rank-one or spectral
(eigenvector/Lanczos) diagnostics.

$\Phi$ against the size reference then decomposes exactly:
\begin{equation}\label{eq:phi-decomposition}
  \Phi(\boldsymbol{\theta})-\Phi_{\mathrm{null}}(\mathbf{p}) = \bigl(\Phi_{\mathrm{sep}}(\mathbf{a})-\Phi_{\mathrm{null}}(\mathbf{p})\bigr)
  + \bigl(1-\Phi_{\mathrm{sep}}(\mathbf{a})\bigr)\bigl(1-S(\boldsymbol{\theta})\bigr).
\end{equation}
The first term is a pure profile comparison, positive exactly when the
diagonal amplitudes are spread more evenly than the parameter counts; the
second is the off-block excess, positive exactly when $S<1$.

\subsection{The correction and its safeguard}
\label{sec:correction}

Fix a minibatch, its gradient $\mathbf{g}=\nabla_{\boldsymbol{\theta}} L$, the selected-AD product
$\mathbf{z}\mapsto \mathbf{H}\mathbf{z}$, a damping $\lambda>0$ and the damped quadratic model
\begin{equation}\label{eq:model}
  m(\mathbf{d})=\mathbf{g}^\top \mathbf{d}+\tfrac12\,\mathbf{d}^\top(\mathbf{H}+\lambda \mathbf{I})\mathbf{d} ,
\end{equation}
where $\mathbf{d}$ is the complete parameter displacement (the curvature methods in
the correction experiment use learning rate one). Let $\mathbf{B}$ be the cached
damped Fisher-based block-diagonal system of the base provider, with
$\mathbf{B}\mathbf{d}_0=-\mathbf{g}$. Its
within-module approximation is not a Hessian block. The exact K-FAC and
EKFAC systems and their treatment of unfactored parameters are specified in
\ref{app:extended-methods} and \ref{app:stage-b-config}.
The model residual is
\begin{equation}\label{eq:residual}
  \begin{aligned}
  \mathbf{r}:=\nabla m(\mathbf{d}_0)&=\mathbf{g}+(\mathbf{H}+\lambda \mathbf{I})\mathbf{d}_0=(\mathbf{H}+\lambda \mathbf{I}-\mathbf{B})\mathbf{d}_0\\
    &=\bigl(\mathbf{H}-\mathrm{bd}(\mathbf{H})\bigr)\mathbf{d}_0+
    \bigl(\mathrm{bd}(\mathbf{H})+\lambda \mathbf{I}-\mathbf{B}\bigr)\mathbf{d}_0 .
  \end{aligned}
\end{equation}
The first term is off-group Hessian action. The second is the remaining
residual, including within-block operator and damping differences from the
Fisher-based baseline; it is not the Fisher information already retained by
K-FAC. In the saved-state comparisons, \emph{full} denotes the combined
residual input $\mathbf{r}$, \emph{off} the off-group input and \emph{rest} the remaining
residual input.

\textbf{Candidate construction.} At a refresh, the pairwise correction
selects up to three strong measured group couplings, builds a small
Hessian-informed subspace $\mathbf{U}$, and solves a damped saddle-free
problem in that subspace. For each saved state and geometry draw, this
solve defines one linear map from a residual input $j$ to a correction
$\boldsymbol{\delta}_j$. The map retains its full-Hessian information;
only the input varies. The raw candidate displacement at scale $\alpha$ is
$\mathbf{d}_0+\alpha\boldsymbol{\delta}_j$, with $\mathbf{d}_1$ denoting
the full candidate at unit scale.

The common-scale comparisons use the same $\alpha$ for all residual inputs.
Off versus base and full versus rest assess the off-input addition from
different starting displacements: $\mathbf{d}_0$ and
$\mathbf{d}_0+\alpha\boldsymbol{\delta}_{\rm rest}$, respectively. The native full
candidate is evaluated directly, not reconstructed by summing components
(\ref{app:residual-input}). A separate full/rest norm control
shortens only the longer correction to the smaller Euclidean correction
norm, leaving the shorter one unchanged. It does not equalize the norms of
the complete displacements or separate scale from direction.

A safeguard accepts the raw candidate only if its norm, damped-model value
and evaluated loss on the step's own minibatch satisfy the tests in
\eqref{eq:safeguard}; otherwise the applied step is $\mathbf{d}_0$. Here $\mathbf{d}_0$ is
the \emph{base candidate} of the Fisher-based preconditioner, not the
unchanged weights. The construction, mathematical conditions, proof of
the exact same-minibatch no-harm safeguard and its current-Hessian descent
bound are in \ref{app:extended-methods}.

\subsection{Study design}
\label{sec:protocol}

The structural measurements compare the off-block fraction with the size and
profile references for each grouping. The central step comparison is
retrospective: candidates are evaluated at the same saved state, whereas
the original correction experiment assesses training outcomes on separate
trajectories. A further study tests signed model--loss agreement along
sampled fractions of base steps (\ref{app:validity} and
\ref{sec:validity-def}).

The original study fixed four criteria internally, without public
preregistration. \ref{app:protocol} gives the configurations and
experimental units; \ref{app:criteria-verdicts} gives the four
criteria and their outcomes.

\textbf{AI-assisted code development.} Claude Opus 5.5 and Claude Sonnet 5.5
(Anthropic) helped write and debug parts of the experimental and reporting code.
Software validation includes synthetic-matrix tests of the measurement and
correction identities, protocol and reporting tests, and checks against saved
scalar records. Figures and tables are generated deterministically from those
records; the figures are rendered with PGFPlots/TikZ in LaTeX.

\section{Results}
\label{sec:results}

\subsection{Profile-aware mass and the same-state partition contrast}
\label{sec:results-a}

What do similar discarded fractions reveal about mass relative to each
model's own profile? In Table~\ref{tab:initial-profile-example}, the nearly
equal initial CIFAR-100 fractions conceal markedly different profiles.
Both fractions exceed their own profile references, but Plain-18's much
lower reference makes its excess substantially larger.

\begin{table}[htbp]
  \centering
  \caption{CIFAR-100 at initialisation, with six groups (stem, four stages and
    head), measured in evaluation mode on the fixed 32-image validation batch
    (test indices 0--31) using 30 probes per block. These are three-seed means
    from the A1 measurements in \ref{app:extended-mass}, which gives spreads, later
    checkpoints and the full architectural comparison. Each profile reference
    is computed within seed before averaging.}
  \label{tab:initial-profile-example}
  \begin{tabular}{@{}lcc@{}}
    \toprule
    Model & Discarded fraction $\Phi$ & Profile reference $\Phi_{\mathrm{sep}}$ \\
    \midrule
    ResNet-18 & $0.832$ & $0.745$ \\
    Plain-18 & $0.834$ & $0.284$ \\
    \bottomrule
  \end{tabular}
\end{table}

At initialization on ImageNet-32, ResNet-18 and Plain-18 likewise have
fractions $0.841$ and $0.844$ with profile references $0.736$ and $0.270$,
respectively. \ref{app:extended-mass} reports the full
architectural ladder and checkpoints. The final CIFAR-100 measurements did
not meet the original criterion for a larger discarded fraction in ResNet-18
than in Plain-18; a separately registered
longer, accuracy-matched comparison is described in \ref{app:protocol} and
\ref{app:extended-mass}.

The size reference addresses a different comparison. At the final six-group
CIFAR-100 ResNet-18 checkpoint, the measured fraction exceeds the size
reference by $\PhiExcessNullREighteen$, although it is
$\PhiSepMinusPhiREighteen$ below the profile reference. The
profile part of equation~\eqref{eq:phi-decomposition} is
$\PhiSepMinusNullREighteen$: curvature within groups is distributed more
evenly than parameter counts, raising the profile reference above the size
reference. Thus exceeding the size reference need not mean unusually high
off-block mass for the measured profile. Averaging across seeds within each mid and final six-group
convolutional cell gives this ordering in
\PhiCellsCnnBetweenReferences/\PhiCellsCnnMidFinal\ cells
(\ref{app:extended-mass}); one individual-seed estimate falls below the size reference.

We next hold weights fixed and change the grouping. At saved epoch-300
CIFAR-100 checkpoints, stage groups comprise stem, four stages and head,
while the alternative separates convolutional and linear layers and pools
unassigned BatchNorm parameters in a common rest group.
The latter is \emph{not} a pure refinement of the former: some parameters
previously in different stage groups are pooled together. At these same
saved ResNet-18 weights, the three-seed mean of $\Phi$ is $0.728$ for six
groups and $0.934$ for 22 layer/rest groups. Estimated $S$ moves from
$1.125$ at stages to $0.876$ at layers: aggregate off-block mass lies below
the stage grouping's profile reference but above the layer grouping's own
reference
(\ref{app:extended-mass}).

A follow-up check applies the two groupings to the same three saved checkpoints
per architecture on two disjoint fixed input batches. Under an approximate
joint screen for conditional probe variation, the ResNet-18 crossing is
resolved in all \CrossResnetResolved/\CrossCellsPerModel\ seed/input cells;
Plain-18 has \CrossPlainResolved\ resolved and \CrossPlainUnresolved\
unresolved comparisons, whose layer intervals contain one. Figure~\ref{fig:crossing-screen}
shows the conditional intervals for the fixed checkpoints and inputs;
\ref{app:crossing-check} gives all cells, thresholds and the probe budget.

\FigureCrossingScreen

\ref{app:extended-mass} extends the profile comparison to the
partially pretrained ResNet-50 and to toy and frozen decoders. These
structural measurements describe the size of the omitted operator. In a
separate ResNet-50 experiment on Stanford Cars, we examine its action on
a fixed base step and compare the resulting minibatch losses when the
off-group correction is added alone or alongside the remaining correction.

\subsection{Standalone and combined residual inputs}
\label{sec:residual-input-results}

We use nine ResNet-50/Stanford Cars states saved at steps 20, 60 and 200
on three plain K-FAC trajectories, one per seed. Within each trajectory
the checkpoints are dependent; at each checkpoint the two subspace draws are paired.
\ref{app:causal-pilot} uses the same states for different interventions;
these are not the trajectories trained with the correction. The weights, prepared minibatch and Fisher-based base
displacement $\mathbf{d}_0$ are held fixed. We use the full, off and
rest inputs of Section~\ref{sec:correction}, splitting the residual at the
six measurement groups rather than K-FAC's module blocks
(equation~\eqref{eq:residual-input-split}). Within each state and draw,
only the input to the common full-Hessian-informed map changes. We compare
raw candidates' minibatch losses before any accept/reject gate: off versus
base evaluates the standalone contribution; full versus rest evaluates the
incremental contribution alongside the remaining residual
(\ref{app:residual-input}). We also compare full and rest with
base: full can outperform rest without improving on the base step.

At unit scale, off alone raises loss relative to base in $17/18$ state/draw
comparisons. In nine of these cases, full has lower loss than rest; across
all cases it does so in $10/18$ (Figure~\ref{fig:residual-contrast}). In four of the nine
(two per draw), full still raises loss relative to base and the safeguard
rejects both full and rest. At seed 42, step 60, by contrast, each component
alone raises loss and is rejected, but their combination improves on base
and is accepted in both draws. In both draws, matching the correction norms
leaves full unchanged and shortens rest, which then has the lower loss
(Table~\ref{tab:residual-controls}).

All 486 candidate losses were measured directly and were finite. A separate
diagnostic asks what fraction of the squared norm of the off-group action
on the base step lies in the constructed subspace. Interpreting the projected
coordinates this way requires a numerically orthonormal basis. Ten of the
18 bases failed the orthonormality screen in native arithmetic, including
both draws at seed 42, step 60. For those cases we report the measured losses but no action-energy
fraction or orthonormal-basis descent guarantee. Across all 18 draws, the
largest relative spectral-norm difference between the correction maps
represented by the saved factors before and after orthonormalization was
\BasisMapDifferencePercentMax\ per cent (\ref{app:residual-input}).
This comparison concerns the maps; the reported losses are those of the
original candidates.

\FigureResidualContrast

The preference depends on scale. The compound event \emph{off raises loss
relative to base while full has lower loss than rest} occurs in $0/18$
state/draw comparisons at each of scales $1/32$, $1/16$ and $1/8$;
$1/18$ at $1/4$, $4/18$ at $1/2$ and $9/18$ at $1$.
These counts use strict recorded signs; they neither estimate a population
frequency nor count harm from off alone. Under the full/rest norm control,
seven of the ten unit-scale full-over-rest preferences reverse.
Rest beats off at matched correction norms
in all nine states in both draws (\ref{app:residual-input}). At
seed 43, step 20, the off-only loss change even switches from $+0.06428$
to $-0.04056$ between geometry draws, and the matched full/rest ordering
reverses.

\subsection{Correction proposals and training outcomes}
\label{sec:results-b}

The tested correction did not demonstrate a training gain on
ResNet-50/Stanford Cars. On the corrected K-FAC trajectories, raw proposals
lower the damped quadratic-model value relative to the base candidate on
\ModelLoweredRecords\ of \RefreshRecords\ refreshes, but the safeguard
based on evaluated loss on the same current minibatch rejects \LossRejectRecords\
($\LossRejectShare$ per cent) of these proposals; a rejected proposal leaves
the \emph{applied} displacement at $\mathbf{d}_0$. Proposition~\ref{prop:pair-descent} gives the corresponding model descent
bound when the Hessian is current and the basis orthonormal. Full acceptance and model-value
statistics are reported in \ref{app:extended-correction}.

Write $q(\mathbf{d})=m(\mathbf{d})-\lambda\|\mathbf{d}\|^2/2$ for the
undamped local quadratic model, and
$\Delta q=q(\mathbf{d}_1)-q(\mathbf{d}_0)$ for its proposal-minus-base
contrast; negative values favour the proposal.
Subtracting the explicit damping penalty retrospectively still leaves
$\Delta q<0$ on $434/438$ loss-rejected K-FAC
refresh proposals and $434/436$ EKFAC proposals, by unrounded logged
values (equation~\eqref{eq:proposal-decomposition}, \ref{app:proposal-decomposition}).
Figure~\ref{fig:proposal-model-loss} compares all raw refresh candidates,
including those not accepted. Damping still shapes candidate construction.
Unequal damping penalties can reverse the ordering of the damped model $m$
and evaluated loss even for an exact quadratic loss. They cannot explain
disagreement between $q$ and that same loss if $q$ is its exact global
quadratic model. The observed discrepancies do not by themselves identify
a cause of rejection or explain the separate model-validity results.

\FigureProposalModelLoss

Mean final accuracies for K-FAC and its corrected version are $84.0$ and
$83.2$ per cent; for EKFAC and its corrected version they are $80.5$ and
$78.9$ per cent on separate three-seed baseline and corrected fine-tuning
trajectories. Full protocol criteria and verdicts are in
\ref{app:criteria-verdicts}; baseline comparisons, costs and stopping
reasons are in \ref{app:extended-correction}.

\section{Discussion}
\label{sec:discussion}

The standalone and incremental effects of a correction can differ even
for a quadratic loss. What matters for an optimizer is whether that
difference changes the step it would take. At seed 42, step 60, a screen
based on the off-group correction alone would have discarded the combined
candidate at unit scale. In both subspace draws the combination lowered
minibatch loss relative to the base step and passed the safeguard; neither
component alone did so. Matching the correction norms left the combination
unchanged but shortened the remainder, which then gave a lower loss still.
The two comparisons distinguish the cost of screening a component at the
proposed scale from the benefit of rescaling the alternative.

For studies of how interactions evolve during training, the choice of
reference matters even when the parameter groups remain fixed. By
\eqref{eq:s-identity}, a change in $\Phi$ can reflect a change in the
diagonal-block profile $\Phi_{\rm sep}$, a change in off-block mass relative
to that profile (through $S$), or both. At the final six-group CIFAR-100
ResNet-18 checkpoint, the discarded fraction is above the size reference
but below the profile reference. Tracking the profile and $S$ separately
therefore gives a study of interlayer dynamics a specific quantity to
explain: how the mass between groups changes relative to the curvature
within them. A change of partition requires a new profile comparison,
as the same-weights stage/layer contrast shows.

For studies of gradient evolution, the corresponding quantity is a vector:
$(\mathbf{H}-\mathrm{bd}(\mathbf{H}))\mathbf{d}_0$ is the part of the
linearized gradient change along the base step that the block-diagonal
Hessian omits. On the same nine saved Cars states, a separate diagnostic
compares the slope along the original correction at the base-step endpoint
using the gradient predicted by the undamped model and the gradient
measured there (\ref{app:causal-pilot}). The two slopes have opposite signs
in $\CausalPilotSlopeSignFlips/\CausalPilotStates$ states. The state-linked
records thus support two distinct checks: whether a model predicts the
directional slope at the endpoint, and whether its proposed correction
actually lowers minibatch loss relative to the base step.

At correction refreshes, most proposals rejected on minibatch loss had
lower undamped quadratic values than their base steps. A separate validity
study traces signed model--loss agreement along sampled fractions of base
steps and shows how
that agreement varies with the optimizer and the inclusion of damping
(\ref{app:validity}). Over the full fine-tuning trajectories,
mean final accuracy was lower for corrected K-FAC and EKFAC than for
their respective baselines.

\textbf{Limits.} All structural results depend on the stated weights, fixed
inputs, selected-AD operator, coordinate system, groupings and probe budgets.
The approximate crossing screen describes conditional Monte Carlo variation,
not input-population or total numerical uncertainty; four Plain-18
comparisons remain unresolved and interval-width goals are not uniformly
met. Decoder seeds vary probes on fixed models and inputs. The correction
covers one fine-tuning task, three separate trajectories per method, and
saved states from three seeds with paired geometry draws. Historical loader
and precision provenance, shared noise, and missing restoration and coverage
measurements further constrain inference about training efficacy
(\ref{app:protocol}). The candidate comparisons show how including
the off-group component changes minibatch loss at a given scale. A natural
next question is whether information available before evaluating those
losses can guide that choice.

\section{Conclusion}
\label{sec:conclusion}

The off-block fraction measures what a grouping omits, not whether a
correction will lower loss. Similar fractions can conceal different diagonal
profiles, and changing the grouping at fixed weights can move the fraction
across its corresponding profile reference. The original correction did not
demonstrate a training gain.

For a fixed base step, however, rejecting a component on its standalone
loss can discard a useful combination. At one checkpoint, neither component
alone improved on the base step at unit scale, but the combined candidate
did so in both subspace draws. Matching correction norms favoured the
remainder instead: the benefit was specific to how the candidates were scaled.

\section*{Declaration of competing interests}
The authors declare that they have no known competing financial interests
or personal relationships that could have appeared to influence the work
reported in this paper.

\section*{CRediT authorship contribution statement}
\textbf{Maxim Bolshim}: Conceptualization, Methodology, Software,
Formal analysis, Writing -- original draft, Visualization.
\textbf{Alexander Kugaevskikh}: Validation, Writing -- review \& editing.

\section*{Data availability}
The code and retained scalar result files are archived at Zenodo
\citep{bolshim2026curvblocksoftware}:
\url{https://doi.org/10.5281/zenodo.24535427}. The archive includes paired
stage/layer probe measurements, predicted and measured endpoint slopes,
and candidate losses indexed by training seed, step and geometry draw for
the scale and norm-matched comparisons. Historical provenance limitations
are described in \ref{app:protocol}.

\section*{Declaration of Generative AI and AI-assisted technologies
  in the writing process}
Claude Opus 5.5 and Claude Sonnet 5.5 (Anthropic) helped write and debug
parts of the experimental and reporting code. For the manuscript, AI assistance
was limited to formatting tables, editing selected descriptions of results
dictated by the authors, and LaTeX typesetting.
The authors checked the manuscript for originality and accuracy, verified
the references and confirmed that the tools' terms of use are suitable for
publication. They take full responsibility for the work.

\FloatBarrier
\bibliographystyle{elsarticle-num-names}
\bibliography{references}

\appendix
\raggedbottom
\counterwithin{equation}{section}
\counterwithin{figure}{section}
\counterwithin{table}{section}
\renewcommand{\theequation}{\Alph{section}.\arabic{equation}}
\renewcommand{\theHequation}{appendix.\Alph{section}.\arabic{equation}}
\renewcommand{\thefigure}{\Alph{section}.\arabic{figure}}
\renewcommand{\thetable}{\Alph{section}.\arabic{table}}

\let\preprintsection\section
\def\section{\FloatBarrier\preprintsection}
\input{appendices.tex}

\end{document}

%% file: numbers.tex
\newcommand{\AccPlainFinalInet}{9.9}
\newcommand{\AccResnetFinalInet}{20.5}
\newcommand{\AcceptShare}{14.8}
\newcommand{\AcceptShareOther}{15.4}
\newcommand{\AcceptShareOtherEkfac}{33.6}
\newcommand{\AcceptShareRefresh}{2.7}
\newcommand{\AcceptShareRefreshEkfac}{3.1}
\newcommand{\AcceptedEarlyRecords}{9}
\newcommand{\AcceptedRecords}{12}
\newcommand{\CouplingBelowTau}{0}
\newcommand{\CouplingFieldMax}{7.326}
\newcommand{\CouplingFieldMedian}{0.852}
\newcommand{\CouplingFieldMinEkfac}{0.202}
\newcommand{\CouplingFieldMinKfac}{0.375}
\newcommand{\CouplingFourthMinEkfac}{0.618}
\newcommand{\CouplingFourthMinKfac}{0.726}
\newcommand{\CouplingLayerOneHeadRfifty}{7.9}
\newcommand{\CouplingObsBothArms}{27000}
\newcommand{\CouplingStemHeadRfifty}{13.7}
\newcommand{\CurvRatioAdamHi}{1.3}
\newcommand{\CurvRatioAdamLo}{0.6}
\newcommand{\CurvRatioLast}{7.6}
\newcommand{\CurvRatioLastHi}{14.2}
\newcommand{\CurvRatioLastLo}{2.3}
\newcommand{\CurvRatioMedianAdam}{0.8}
\newcommand{\CurvRatioMedianKfac}{3.8}
\newcommand{\CurvRatioMedianSgd}{3.3}
\newcommand{\CurvRatioSpanHi}{14.4}
\newcommand{\CurvRatioSpanLo}{1.7}
\newcommand{\DampSharePctAdamHi}{87}
\newcommand{\DampSharePctAdamLo}{71}
\newcommand{\DiscardedBetweenStages}{92.8}
\newcommand{\DiscardedWithinStages}{7.2}
\newcommand{\EpsAdditiveShare}{95.3}
\newcommand{\EpsLayerThreeStem}{0.029}
\newcommand{\EpsStemLayerThree}{32.6}
\newcommand{\FinalDeltaLnSLayer}{-0.0395}
\newcommand{\FinalDeltaLnSLayerHi}{0.0678}
\newcommand{\FinalDeltaLnSLayerLo}{-0.1467}
\newcommand{\FinalDeltaLnSStage}{-0.0284}
\newcommand{\FinalDeltaLnSStageHi}{0.0348}
\newcommand{\FinalDeltaLnSStageLo}{-0.0916}
\newcommand{\FinalDeltaPhiLayer}{0.0130}

\newcommand{\FinalDeltaPhiStage}{0.0804}
\newcommand{\FinalDeltaPhiStageHi}{0.1167}
\newcommand{\FinalDeltaPhiStageLo}{0.0440}
\newcommand{\FinalDeltaSepLayer}{0.0111}
\newcommand{\FinalDeltaSepLayerHi}{0.0152}
\newcommand{\FinalDeltaSepLayerLo}{0.0070}
\newcommand{\FinalDeltaSepStage}{0.0625}
\newcommand{\FinalDeltaSepStageHi}{0.0795}
\newcommand{\FinalDeltaSepStageLo}{0.0456}
\newcommand{\GammaAboveOneShare}{0.67}
\newcommand{\GammaEligibleRecordsEkfac}{449}
\newcommand{\GammaEligibleRecordsKfac}{450}
\newcommand{\GammaSkippedRecordsEkfac}{1}
\newcommand{\GammaSkippedRecordsKfac}{0}
\newcommand{\GammaZeroShare}{97.3}
\newcommand{\GateAMeanDelta}{0.011}
\newcommand{\GateAMidMeanDelta}{0.081}
\newcommand{\GateAMidPooledHi}{0.107}
\newcommand{\GateAMidPooledLo}{0.068}
\newcommand{\GateAMidPooledMean}{0.088}
\newcommand{\GateASeedHi}{0.070}
\newcommand{\GateASeedLo}{-0.054}
\newcommand{\GranGroupsBlock}{17}
\newcommand{\GranGroupsLayer}{55}
\newcommand{\GranGroupsStage}{6}
\newcommand{\HeadShareRfiftyHi}{83}
\newcommand{\HeadShareRfiftyLo}{82}
\newcommand{\InetDeltaFinalOne}{0.116}
\newcommand{\InetDeltaFinalThree}{0.129}
\newcommand{\InetDeltaFinalTwo}{0.174}
\newcommand{\KeptShareLayer}{0.0005}
\newcommand{\KeptShareStage}{0.073}
\newcommand{\KinkShareInit}{1.4}
\newcommand{\KinkShareLate}{0.1}
\newcommand{\LambdaMinCellsBelowDamping}{52}
\newcommand{\LambdaMinCellsTotal}{54}
\newcommand{\LongAccPlainFinal}{61.5}
\newcommand{\LongAccResnetFinal}{63.3}
\newcommand{\LongDeltaPhi}{0.086}
\newcommand{\LongDeltaPhiHi}{0.112}
\newcommand{\LongDeltaPhiLo}{0.060}
\newcommand{\LongDeltaPhiTolHi}{0.090}
\newcommand{\LongDeltaPhiTolLo}{0.067}
\newcommand{\LongPairs}{19}
\newcommand{\LongPhiResnetFinal}{0.728}
\newcommand{\LongTolHi}{0.100}
\newcommand{\LongTolLo}{0.005}
\newcommand{\LongTolWorking}{0.020}
\newcommand{\LossRejectRecords}{438}
\newcommand{\LossRejectShare}{97.3}
\newcommand{\MedianModelBase}{1.15}
\newcommand{\MedianModelProposal}{-0.002}
\newcommand{\ModelLoweredRecords}{450}
\newcommand{\NewtonDivergeStep}{1800}
\newcommand{\NewtonNegCurvShare}{80.9}
\newcommand{\NewtonNegCurvShareFiniteLo}{93.3}
\newcommand{\OffOverSepGranBlock}{1.8}
\newcommand{\OffOverSepGranLayer}{248}
\newcommand{\OffOverSepGranStage}{5.7}
\newcommand{\OracleNegCurvShare}{66.7}
\newcommand{\PhiBlockBelowStageSeeds}{1}
\newcommand{\PhiCellsCnnBetweenReferences}{16}
\newcommand{\PhiCellsCnnMidFinal}{16}
\newcommand{\PhiCellsRandomInit}{11}
\newcommand{\PhiCellsTotal}{30}
\newcommand{\PhiCellsTrainedBelowSep}{18}
\newcommand{\PhiCellsTrainedWeights}{18}
\newcommand{\PhiExcessNullREighteen}{0.236}
\newcommand{\PhiGranBlockVThree}{0.928}
\newcommand{\PhiGranLayerVThree}{0.9995}
\newcommand{\PhiGranStageVThree}{0.927}
\newcommand{\PhiLlama}{0.814}
\newcommand{\PhiNullLlama}{0.928}
\newcommand{\PhiNullQwen}{0.907}
\newcommand{\PhiNullResNetHi}{0.517}
\newcommand{\PhiNullResNetLo}{0.400}
\newcommand{\PhiNullRfifty}{0.517}
\newcommand{\PhiNullToyHi}{0.930}
\newcommand{\PhiNullToyLo}{0.803}
\newcommand{\PhiPlainFinalCifar}{0.629}
\newcommand{\PhiPlainFinalInet}{0.508}
\newcommand{\PhiPretrainedRfifty}{0.927}
\newcommand{\PhiQwen}{0.679}
\newcommand{\PhiResnetFinalCifar}{0.640}
\newcommand{\PhiResnetFinalInet}{0.648}
\newcommand{\PhiSepGranBlock}{0.876}
\newcommand{\PhiSepGranLayer}{0.892}
\newcommand{\PhiSepGranStage}{0.693}
\newcommand{\PhiSepMinusNullREighteen}{0.294}
\newcommand{\PhiSepMinusPhiREighteen}{0.058}
\newcommand{\PhiSepRfifty}{0.695}
\newcommand{\PhiToyHi}{0.651}
\newcommand{\PhiToyLo}{0.583}
\newcommand{\PhiTrainedResNetHi}{0.648}
\newcommand{\PhiTrainedResNetLo}{0.606}
\newcommand{\ProposalsTotal}{9000}
\newcommand{\RefreshRecords}{450}
\newcommand{\RejectByLoss}{6766}
\newcommand{\RejectByLossShare}{75.2}
\newcommand{\RejectByModel}{14}
\newcommand{\RejectByRatio}{892}
\newcommand{\RestMassLayerPpm}{0.9}
\newcommand{\SGranBlock}{0.584}
\newcommand{\SGranLayer}{0.005}
\newcommand{\SGranStage}{0.236}
\newcommand{\SNoHeadRfiftyHi}{0.95}
\newcommand{\SNoHeadRfiftyLo}{0.92}
\newcommand{\SecantBelowAdam}{3}
\newcommand{\SecantBelowKfac}{12}
\newcommand{\SecantBelowSgd}{14}
\newcommand{\SecantCellsArm}{18}
\newcommand{\SpearmanEkfac}{0.04}
\newcommand{\SpearmanEkfacP}{0.00004}
\newcommand{\SpearmanKfac}{0.01}
\newcommand{\SpearmanKfacP}{0.105}
\newcommand{\SpearmanObs}{13500}
\newcommand{\SpearmanPairsPerRecord}{30}
\newcommand{\TstarCellsDefinedTolFive}{25}
\newcommand{\TstarCellsOk}{27}
\newcommand{\TstarCellsOkCoarse}{44}
\newcommand{\TstarCellsOkDefinedTolFive}{25}
\newcommand{\TstarCellsOkHundredth}{7}
\newcommand{\TstarCellsOkTolFifty}{14}
\newcommand{\TstarCellsOkTolTwenty}{24}
\newcommand{\TstarCellsOkTwentieth}{17}
\newcommand{\TstarCellsPerArm}{9}
\newcommand{\TstarCellsTotal}{27}
\newcommand{\TstarCellsTotalCoarse}{45}
\newcommand{\TstarDropKfac}{5.8}
\newcommand{\TstarDropSgd}{25}
\newcommand{\TstarMedianAdam}{0.075}
\newcommand{\TstarMedianEarlyKfac}{0.175}
\newcommand{\TstarMedianLateKfac}{0.03}
\newcommand{\TstarMedianTolFifty}{0.1}
\newcommand{\TstarMedianTolFive}{0.01}
\newcommand{\TstarOriginalStepSixHundred}{0.01}
\newcommand{\TstarOriginalStepTwoHundred}{0.1}
\newcommand{\TstarReplicateStepSixHundred}{0.05}
\newcommand{\TstarReplicateStepTwoHundred}{0.25}
\newcommand{\TstarReplicateUndeterminedStep}{1500}
\newcommand{\TstarTol}{0.1}
\newcommand{\TstarUndampedHalfAdam}{8}
\newcommand{\TstarUndampedOkKfac}{9}
\newcommand{\TstarUndampedOkSgd}{8}
\newcommand{\TstarUndeterminedTolFive}{2}
\newcommand{\ValidityShareEkfac}{98.0}
\newcommand{\ValidityShareEkfacOverlay}{98.7}
\newcommand{\ValidityShareKfac}{86.4}
\newcommand{\ValidityShareKfacOverlay}{95.3}

\newcommand{\CausalPilotStates}{9}
\newcommand{\CausalPilotSeeds}{3}

\newcommand{\CausalPilotGradientScaleWins}{4}

\newcommand{\CausalPilotSlopeSignFlips}{4}
\newcommand{\CausalPilotReluFlips}{3}
\newcommand{\CausalPilotPoolFlips}{0}
\newcommand{\CausalPilotReluBothLossesRiseFlips}{2}
\newcommand{\CausalPilotGainFloor}{0.0001}

\newcommand{\CrossResnetResolved}{6}
\newcommand{\CrossPlainResolved}{2}
\newcommand{\CrossCellsPerModel}{6}
\newcommand{\CrossPlainUnresolved}{4}
\newcommand{\CrossSGoalsUnmet}{5}
\newcommand{\CrossFamilyGoalsUnmet}{10}
\newcommand{\CrossSHalfWidthLo}{0.0075}
\newcommand{\CrossSHalfWidthHi}{0.0440}
\newcommand{\CrossCriticalValue}{3.170180}
\newcommand{\CrossNarrowLayerUpper}{0.9984}

\newcommand{\AblationExpected}{486}

\newcommand{\AblationValid}{486}

\newcommand{\BasisNativeErrorMin}{1.30{\cdot}10^{-5}}
\newcommand{\BasisNativeErrorMax}{6.22{\cdot}10^{-4}}
\newcommand{\BasisGramEigenMin}{0.999876}
\newcommand{\BasisGramEigenMax}{1.000623}
\newcommand{\BasisGramConditionMax}{1.000622}
\newcommand{\BasisMapDifferencePercentMax}{0.025}

%% file: appendices.tex
\section{Extended measurement and correction methods}
\label{app:extended-methods}

\subsection{Projection and stochastic estimation}
For a fixed partition, the operator and fraction of Section~\ref{sec:phi}
have the following exact projection property.

\begin{proposition}[$\Phi$ is the block-diagonal approximation error]
  \label{prop:phi-projection}
  For every block-diagonal matrix $\mathbf{B}$ whose blocks conform to the partition,
  $\|\mathbf{H}_{\boldsymbol{\theta}}-\mathbf{B}\|_F^2=\|\mathbf{H}_{\boldsymbol{\theta}}-\mathrm{bd}(\mathbf{H}_{\boldsymbol{\theta}})\|_F^2
  +\|\mathrm{bd}(\mathbf{H}_{\boldsymbol{\theta}})-\mathbf{B}\|_F^2$. Hence
  $\Phi(\boldsymbol{\theta})=\min_{\mathbf{B}}\|\mathbf{H}_{\boldsymbol{\theta}}-\mathbf{B}\|_F^2/\|\mathbf{H}_{\boldsymbol{\theta}}\|_F^2$ over block-diagonal
  $\mathbf{B}$, attained at $\mathbf{B}=\mathrm{bd}(\mathbf{H}_{\boldsymbol{\theta}})$.
\end{proposition}

\noindent\emph{Proof.} The matrix $\mathbf{H}_{\boldsymbol{\theta}}-\mathrm{bd}(\mathbf{H}_{\boldsymbol{\theta}})$ has zero
diagonal blocks and $\mathrm{bd}(\mathbf{H}_{\boldsymbol{\theta}})-\mathbf{B}$ has zero off-diagonal blocks, so
the two are orthogonal in the Frobenius inner product and their squared norms add.
Minimizing over $\mathbf{B}$ sets the second term to zero.

Residual networks use the stem, four stages and
head. Transformers use the embedding, four attention projections per layer
(with pre-attention normalization assigned to the query group), the multi-layer
perceptron with its pre-norm, and the head. Each tied tensor belongs to its first group only. In Qwen and Llama the
shared output embedding belongs to the embedding group, leaving only the
final norm in the head. The toy
output embedding is untied. Details are in \ref{app:protocol}.

These groups combine multiple linear or convolutional modules. K-FAC and EKFAC
factor one Kronecker pair per such module and use the identity on unfactored
normalization parameters. A nested per-module partition therefore refines these groups. The projection
bound in Proposition~\ref{prop:phi-projection} quantifies Hessian approximation error.
Section~\ref{sec:results-a} treats the
non-nested partitions used in the granularity comparison.

Throughout, $\mathbf{H}_{\boldsymbol{\theta}}$ denotes the parameter second-order operator returned by
double backward at fixed weights, inputs and mode, using the implementation's
selected automatic-differentiation (AD) conventions for rectifiers and max-pool
winners. It represents a classical Hessian $\nabla^2L(\boldsymbol{\theta})$ only when all
differentiated primitives follow compatible locally twice continuously
differentiable branches; smoothness of the final scalar loss alone is
insufficient. Zero activations and tied pooling winners are not excluded. The exact
symmetric-matrix results below apply to this selected operator, including
where a classical Hessian is not established. Similarly, $\mathbf{g}$ denotes the
selected-AD gradient, which is the operational meaning of $\mathbf{g}=\nabla_{\boldsymbol{\theta}} L$
at nonsmooth points. The Frobenius diagnostics are coordinate-dependent:
a rescaling of parameter coordinates need not preserve their values.
Write $\mathbf{H}_{\boldsymbol{\theta}_v,\boldsymbol{\theta}_w}$ for its $(v,w)$ block and
$\mathrm{bd}(\mathbf{H}_{\boldsymbol{\theta}})=\mathrm{blkdiag}(\mathbf{H}_{\boldsymbol{\theta}_v,\boldsymbol{\theta}_v})$.

\textbf{Estimator.} With $\mathbf{g}=\nabla_{\boldsymbol{\theta}} L$,
$\mathbf{H}_{\boldsymbol{\theta}_v,\boldsymbol{\theta}_w}\mathbf{z}_w=\nabla_{\boldsymbol{\theta}_v}\langle \mathbf{g}_{\boldsymbol{\theta}_w},\mathbf{z}_w\rangle$ is
one double backward pass \citep{pearlmutter1994fast}, and
$\|\mathbf{H}_{\boldsymbol{\theta}_v,\boldsymbol{\theta}_w}\|_F^2=\mathbb{E}_{\mathbf{z}}\|\mathbf{H}_{\boldsymbol{\theta}_v,\boldsymbol{\theta}_w}\mathbf{z}_w\|^2$ for
Rademacher $\mathbf{z}_w\in\{-1,+1\}^{p_w}$ \citep{avron2011randomized}. With $m$ probes
per block,
$\hat\Phi=1-\sum_v\hat N_{vv}/\bigl(\sum_v\hat N_{vv}+2\sum_{v<w}\hat N_{vw}\bigr)$,
where each $\hat N_{vw}\ge0$ is unbiased for $\|\mathbf{H}_{\boldsymbol{\theta}_v,\boldsymbol{\theta}_w}\|_F^2$ and
$\|\mathbf{H}_{\boldsymbol{\theta}_v,\boldsymbol{\theta}_w}\|_F=\|\mathbf{H}_{\boldsymbol{\theta}_w,\boldsymbol{\theta}_v}\|_F$. With complete pair measurements, the numerator and denominator use the
same component estimates. The cost is $\tfrac12n_b(n_b{+}1)\,m$ double backward passes for
$n_b$ groups. When only a band of pairs is measured, as for the deep
transformers, $\|\mathbf{H}_{\boldsymbol{\theta}}\|_F^2$ comes from a separate full-parameter probe:
$\hat\Phi$ still estimates $\Phi$ and only the coupling field is restricted to
the band. The same probes give that coupling field,
$\mathcal{C}(v,w):=\|\mathbf{H}_{\boldsymbol{\theta}_v,\boldsymbol{\theta}_w}\|_F/ \sqrt{\|\mathbf{H}_{\boldsymbol{\theta}_v,\boldsymbol{\theta}_v}\|_F\|\mathbf{H}_{\boldsymbol{\theta}_w,\boldsymbol{\theta}_w}\|_F}$,
at no extra cost where both diagonal norms are positive; it is the symmetric
variable on which the gate of Section~\ref{sec:correction} selects unordered
pairs, and it is not bounded by one in general. Nonzero $\mathbf{H}$ alone does not
ensure a nonzero diagonal profile. Ratios with a zero denominator are undefined;
non-finite observations are not interpreted as measured correlations.
The estimator treats an estimated squared full norm below $10^{-12}$,
or a coupling/action denominator below $10^{-12}$, as undefined. These
numerical-zero conventions exclude no association observations in the
reported inputs.

Unbiasedness of $\hat N_{vw}$ does not imply unbiasedness of their square
roots or of nonlinear quantities such as $\hat\Phi$, estimated $\mathcal C$
and $S$.
The exact bounds concern the target matrix; stochastic estimates need not
satisfy them sample by sample. Unless stated otherwise, nonlinear quantities
are computed within seed before averaging. Three-seed intervals describe
those realizations rather than all estimator or training uncertainty.

\subsection{Profile-reference properties}
The references and their interpretation are introduced in Section~\ref{sec:references}.

$\Phi_{\mathrm{null}}$ and $\Phi_{\mathrm{sep}}$ are both instances of one
functional. Writing $n_{\mathrm{eff}}(\mathbf{x}):=\bigl(\sum_vx_v\bigr)^2/\sum_vx_v^2$
for a non-negative, nonzero profile $\mathbf{x}=(x_v)_{v\in V}$, $\Phi_{\mathrm{null}}(\mathbf{p})=1-1/n_{\mathrm{eff}}(\mathbf{p})$
and $\Phi_{\mathrm{sep}}(\mathbf{a})=1-1/n_{\mathrm{eff}}(\mathbf{a})$.

If $\mathbf{H}_{\boldsymbol{\theta}}$ is positive semidefinite, every pair with
$a_va_w>0$ has $\mathcal{C}(v,w)\le1$:
writing $\mathbf{H}_{\boldsymbol{\theta}}=\mathbf{X}^\top \mathbf{X}$ with $\mathbf{X}_v$ the columns of group $v$, the
Cauchy--Schwarz inequality for the trace inner product gives
$\|\mathbf{H}_{\boldsymbol{\theta}_v,\boldsymbol{\theta}_w}\|_F^2=\mathrm{tr}(\mathbf{X}_v\mathbf{X}_v^\top \mathbf{X}_w\mathbf{X}_w^\top)\le\|\mathbf{H}_{\boldsymbol{\theta}_v,\boldsymbol{\theta}_v}\|_F\,\|\mathbf{H}_{\boldsymbol{\theta}_w,\boldsymbol{\theta}_w}\|_F$.
Then $S\ge1$ and $\Phi\le\Phi_{\mathrm{sep}}$, with equality attainable by
a rank-one matrix for every such diagonal-norm profile. If $S<1$ or any
$\mathcal{C}(v,w)>1$, neither $\mathbf{H}_{\boldsymbol{\theta}}$ nor $-\mathbf{H}_{\boldsymbol{\theta}}$ can be positive
semidefinite, since both have the same block norms; hence $\mathbf{H}_{\boldsymbol{\theta}}$ is
indefinite. The converse does not hold, and estimated crossings alone do
not certify indefiniteness (Section~\ref{sec:references}).

\subsection{Correction construction and reference solve}
The damped model and residual are defined in Section~\ref{sec:correction}.
In K-FAC, a module's block is the factored-damped
$(\mathbf{A}+\lambda \mathbf{I})\otimes(\mathbf{G}+\lambda \mathbf{I})$, not $\mathbf{A}\otimes \mathbf{G}+\lambda \mathbf{I}$; in EKFAC
it is the cached eigenbasis-corrected Fisher approximation plus damping.
On unfactored parameters $\mathbf{B}=\mathbf{I}$. The base system is Fisher-based, not the
full or block Hessian.

Positive definiteness of $\mathbf{H}+\lambda \mathbf{I}$ gives
a unique minimizer. A singular positive semidefinite system can have nonunique
minimizers if $\mathbf{g}$ is in its range; otherwise it has no finite minimum.

A negative block direction transfers to the full parameter Hessian: if a
diagonal block $\mathbf{H}_{\boldsymbol{\theta}_v,\boldsymbol{\theta}_v}$ has an eigenvalue below $-\lambda$, so
does $\mathbf{H}$ (extend the eigenvector with zeros outside the block), and along that
direction the damped model~\eqref{eq:model} is unbounded below. The implication
runs one way only: it forces a negative direction of $\mathbf{H}$, not necessarily
indefiniteness (which also requires a positive direction). An indefinite $\mathbf{H}$
need not put any one diagonal block below $-\lambda$.
\ref{app:conditioning} gives a separate single-seed ResNet-18/CIFAR-10
illustration of block spectra, not evidence about the blocks of the
ResNet-50 correction experiment on Stanford Cars.

With
$\mathbf{d}_{0,v}$ denoting the group-$v$ component of $\mathbf{d}_0$, the ordered cross-term
ratio compares cross-block and diagonal action along the base step:
\begin{equation}\label{eq:cross-term}
  \varepsilon(v,w):=\frac{\|\mathbf{H}_{\boldsymbol{\theta}_v,\boldsymbol{\theta}_w}\mathbf{d}_{0,w}\|}
  {\|\mathbf{H}_{\boldsymbol{\theta}_v,\boldsymbol{\theta}_v}\mathbf{d}_{0,v}\|} .
\end{equation}
This ordered step-action diagnostic is compared with $\mathcal C$, the symmetric
quantity used for pair selection, in \ref{app:coupling-sensitivity}.
It is defined only for a positive denominator: $v$ is the
receiver and $w$ the source of $\mathbf{H}_{\boldsymbol{\theta}_v,\boldsymbol{\theta}_w}\mathbf{d}_{0,w}$.
A stored label $v\to w$ denotes the ordered entry, not a physical action
from $v$ into $w$.

\textbf{The correction.} COUPLE-FAC uses a refresh period $T$ and four operations.
\begin{enumerate}[label=(\roman*),leftmargin=1.6em,itemsep=2pt,topsep=2pt]
  \item \emph{Gate.} Every $T$ steps the coupling field is re-measured with
    $m_c$ probes and the unordered set $\mathcal{E}=\{\{v,w\}:\mathcal{C}(v,w)>\tau\}$ is formed,
    strongest pair first, with at most $s_{\max}$ pairs.
  \item \emph{Subspace}, every $T$ steps. For each selected pair the leading $\ell$
    singular-vector approximations $(\mathbf{L}_{vw},\mathbf{R}_{vw})$ of $\mathbf{H}_{\boldsymbol{\theta}_v,\boldsymbol{\theta}_w}$ are obtained by
    $n_{\mathrm{it}}$ rounds of matrix-free subspace iteration, and one joint basis
    is formed over all selected
    pairs,
    \begin{equation}\label{eq:subspace}
      \begin{aligned}
        \mathbf{U}&=\mathrm{orth}\bigl[\mathbf{P}_v\mathbf{d}_{0,v},\,\mathbf{P}_w\mathbf{d}_{0,w},\,\mathbf{P}_v\mathbf{L}_{vw},\,\mathbf{P}_w\mathbf{R}_{vw}:(v,w)\in\mathcal{E}\bigr],\\
        \hat{\mathbf{H}}&=\mathbf{U}^\top \mathbf{H}\mathbf{U}=\mathbf{V}\boldsymbol{\Lambda} \mathbf{V}^\top ,
      \end{aligned}
    \end{equation}
    with $\mathbf{P}_v\in\mathbb{R}^{P\times p_v}$ the coordinate embedding of group $v$,
    $k\le|\mathcal{E}|(2+2\ell)$ columns,
    $k$ Hessian-vector products and one $k\times k$ eigendecomposition.
  \item \emph{Step}, every step. From $\mathbf{r}=\mathbf{g}+\mathbf{H}\mathbf{d}_0+\lambda \mathbf{d}_0$, at one
    Hessian-vector product,
    \begin{equation}\label{eq:saddle-free}
      \mathbf{c}^*=-\mathbf{V}\,\mathrm{diag}\bigl(|\lambda_i|+\lambda\bigr)^{-1}\mathbf{V}^\top \mathbf{U}^\top \mathbf{r},
      \qquad \mathbf{d}_1=\mathbf{d}_0+\mathbf{U}\mathbf{c}^* ,
    \end{equation}
    where the $\lambda_i$ are the eigenvalues in $\boldsymbol{\Lambda}$. This is the
    saddle-free solve \citep{dauphin2014identifying} with positive-definite
    metric $|\hat{\mathbf{H}}|+\lambda \mathbf{I}$ and right-hand side $-\mathbf{U}^\top \mathbf{r}$.
    It is not Newton on $\hat{\mathbf{H}}+\lambda \mathbf{I}$.
    The damping parameter is shared with the base provider, although the
    systems damped are different.
  \item \emph{Safeguard}, every step, at one Hessian-vector product and two forward
    passes. With $L_{\mathrm{mb}}(\mathbf{d}):=L(\boldsymbol{\theta}+\mathbf{d})$ on the step's own minibatch,
    in the same mode and with the same objective,
    \begin{equation}\label{eq:safeguard}
      \mathbf{d}=\begin{cases}
        \mathbf{d}_1,&\|\mathbf{d}_1\|\le c_{\mathrm{step}}\|\mathbf{d}_0\|,\ \ m(\mathbf{d}_1)\le m(\mathbf{d}_0)
        \ \ \text{and}\ \ L_{\mathrm{mb}}(\mathbf{d}_1)\le L_{\mathrm{mb}}(\mathbf{d}_0),\\
        \mathbf{d}_0,&\text{otherwise}.
      \end{cases}
    \end{equation}
\end{enumerate}
An empty gate returns the base displacement. Because the proposal uses both
components of $\mathbf{r}$, it combines cross-block and within-block Hessian information
rather than isolating an off-block intervention. Between refreshes, $\mathbf{U}$ and
its reduced eigensystem are stale; the residual and safeguard are evaluated
at the current state.
\ref{app:protocol} specifies the shared augmented minibatch and isolated
candidate-trial buffer state used in the loss comparison.

\begin{proposition}[No-harm]
  \label{prop:no-harm}
  For finite candidate norms, model values and losses, exact comparisons on
  the same current objective, mode and minibatch give, for any cache staleness,
  $m(\mathbf{d})\le m(\mathbf{d}_0)$,
  $\|\mathbf{d}\|\le\max(1,c_{\mathrm{step}})\,\|\mathbf{d}_0\|$ and
  $L_{\mathrm{mb}}(\mathbf{d})\le L_{\mathrm{mb}}(\mathbf{d}_0)$ on the minibatch of the step.
\end{proposition}

\noindent\emph{Proof.} The three inequalities are the acceptance test
\eqref{eq:safeguard}, evaluated on the exact model and the exact minibatch loss at
the current weights; the fallback $\mathbf{d}_0$ satisfies them trivially.

\begin{proposition}[Descent of the saddle-free correction]
  \label{prop:pair-descent}
  At a refresh using the current state, let $\mathbf{U}$ have orthonormal columns and
  $\hat{\mathbf{H}}=\mathbf{U}^\top \mathbf{H}\mathbf{U}=\mathbf{V}\boldsymbol{\Lambda} \mathbf{V}^\top$ for the current $\mathbf{H}$, with
  eigenvalues $\lambda_i$, $s_i=|\lambda_i|+\lambda$ with $\lambda>0$,
  $\mathbf{q}=\mathbf{U}^\top \mathbf{r}$ with $\mathbf{r}$ as in \eqref{eq:residual}, and
  $\mathbf{c}^*=-\mathbf{V}\,\mathrm{diag}(s_i)^{-1}\mathbf{V}^\top \mathbf{q}$. Then, for the damped
  model~\eqref{eq:model},
  \begin{equation*}
    m(\mathbf{d}_0+\mathbf{U}\mathbf{c}^*)-m(\mathbf{d}_0)
    =-\sum_i\frac{2s_i-\lambda_i-\lambda}{2s_i^2}\,(\mathbf{V}^\top \mathbf{q})_i^2
    \le-\frac{\|\mathbf{U}^\top \mathbf{r}\|^2}{2\bigl(\|\hat{\mathbf{H}}\|_2+\lambda\bigr)},
  \end{equation*}
  and $\|\mathbf{c}^*\|\le\|\mathbf{U}^\top \mathbf{r}\|/\lambda$.
\end{proposition}

\noindent\emph{Proof.} Since $\mathbf{U}^\top \mathbf{U}=\mathbf{I}$,
$m(\mathbf{d}_0+\mathbf{U}\mathbf{c})-m(\mathbf{d}_0)=\mathbf{r}^\top \mathbf{U}\mathbf{c}+\tfrac12\mathbf{c}^\top(\hat{\mathbf{H}}+\lambda \mathbf{I})\mathbf{c}$. In the eigenbasis,
with $\tilde{\mathbf{q}}=\mathbf{V}^\top \mathbf{q}$ and $\tilde c_i=-\tilde q_i/s_i$, the $i$-th term is
$-\tilde q_i^2/s_i+(\lambda_i+\lambda)\tilde q_i^2/(2s_i^2)$, which is
$-\tilde q_i^2(2s_i-\lambda_i-\lambda)/(2s_i^2)$. Since
$2s_i-\lambda_i-\lambda=2|\lambda_i|-\lambda_i+\lambda\ge|\lambda_i|+\lambda=s_i$,
the factor is at least $1/(2s_i)$, and $\max_is_i=\|\hat{\mathbf{H}}\|_2+\lambda$. The norm
bound follows from $|\tilde c_i|\le|\tilde q_i|/\lambda$.

The reference $\mathbf{d}_{\mathrm{N}}$ comes from a separate truncated Newton-CG
solve of $(\mathbf{H}+\lambda \mathbf{I})\mathbf{d}=-\mathbf{g}$ for model~\eqref{eq:model} at the same weights,
using the stopping rules in \ref{app:stage-b-config}. It is labelled an
oracle in the outputs. On non-positive search
direction curvature, including zero, it returns $-\mathbf{g}$ if the current iterate
is zero, otherwise that iterate. Thus $\mathbf{d}_{\mathrm{N}}$ is a truncated
reference, not a computed optimum. An indefinite $\mathbf{H}+\lambda \mathbf{I}$ makes the
quadratic unbounded below.

\begin{definition}[Closed gap]
  \label{def:gamma}
  On a record with $m(\mathbf{d}_0)-m(\mathbf{d}_{\mathrm{N}})>
  10^{-8}\max(1,|m(\mathbf{d}_{\mathrm{N}})|)$,
  the closed gap is
  \begin{equation}\label{eq:gamma}
    \gamma=\bigl(m(\mathbf{d}_0)-m(\mathbf{d})\bigr)/\bigl(m(\mathbf{d}_0)-m(\mathbf{d}_{\mathrm{N}})\bigr) ,
  \end{equation}
  with $\mathbf{d}$ the step \eqref{eq:safeguard} actually takes, reported unclipped.
  If the safeguard retains $\mathbf{d}_0$, then $\gamma=0$.
\end{definition}

\textbf{Roles of the three tests.} The norm bound keeps the correction on
the scale of the base step. The model test penalizes length using the base
step's damping $\lambda$. The loss test is necessary because
$m(\mathbf{d}_1)\le m(\mathbf{d}_0)$ need not order evaluated minibatch losses, even for an exact
global quadratic when damping penalties differ. Like the actual-reduction ratio of
trust-region methods \citep{nocedal2006numerical}, it evaluates loss, but
compares two candidates rather than actual and predicted reduction.


\section{Protocol and pre-specified criteria}
\label{app:protocol}

The runs used a 24\,GB Quadro RTX~6000. Timings are indicative because the
card was shared. Training uses seeds 42--44, except for the single conditioning
run at seed 42. The five partially pretrained ResNet-50 snapshots use seeds
42--46, each drawing a new head and probes. Decoder seeds 42--44 vary only
Hutchinson probes on fixed models and inputs, so their variation describes
probe sampling rather than independent model or corpus replication.

\subsection{Structural measurement configurations (Stage A)}

Table~\ref{tab:stage-a-config} specifies the six Stage A runs, their datasets,
training settings and probe budgets. \ref{app:extended-mass} reports the
corresponding $\Phi$ measurements.

\begin{table}[htbp]
  \centering
  \caption{Stage A runs.  $\Phi$ and the coupling field are measured in
    evaluation mode on one fixed minibatch per run, of validation images in the
    first four rows, with $m$ Rademacher probes per block, at the listed
    checkpoints.}
  \label{tab:stage-a-config}
  \scriptsize
  \setlength{\tabcolsep}{3pt}
  \begin{tabular}{@{}l>{\raggedright\arraybackslash}p{3.1cm}>{\raggedright\arraybackslash}p{2.5cm}>{\raggedright\arraybackslash}p{3.4cm}ll@{}}
    \toprule
    Run & Model & Data & Training & Checkpoints & $m$ / batch \\
    \midrule
    A1 & ResNet-18 vs.\ plain & CIFAR-100 & SGD $0.1$, cosine, 100 ep, batch 128 & init, mid, final & 30 / 32 \\
    A1-width & ResNet-18 $\times\{0.5,1,2\}$; ResNet-34 & CIFAR-100 & as A1 & init, mid, final & 30 / 32 \\
    A1-IN32 & ResNet-18 vs.\ plain & ImageNet-32 & as A1, 30 ep & init, mid, final & 30 / 32 \\
    A2 & ResNet-50, ImageNet weights, new random head & Stanford Cars & none & init & 30 / 32 \\
    A3-toy & decoder-only, tiny/small/base & uniform random tokens, seq.\ 64 & none (fixed init) & init & 8 / 8 \\
    A3-LLM & Qwen2.5-0.5B; Llama-3.2-1B & fixed author text, repeated, seq.\ 128 & none & init & 8 / 4 (Llama: 2), band 8 \\
    \bottomrule
  \end{tabular}
\end{table}

For the image measurements, the protocol reconstructed from the saved source
selects the first listed number of examples from an unshuffled validation
loader, fixed across checkpoints, architectures and seeds within each dataset. For CIFAR-100 this is the official
test split used as validation, ordered zero-based indices 0--31: tensor
conversion and channel normalization only, without validation augmentation or
shuffle. CIFAR uses 32-pixel images; training alone adds random crops with
four-pixel padding and horizontal flips. Channel means/standard deviations are
$(0.5071,0.4865,0.4409)/(0.2673,0.2564,0.2762)$ for CIFAR-100 and
$(0.4914,0.4822,0.4465)/(0.2470,0.2435,0.2616)$ for the conditioning
CIFAR-10 run. ImageNet-32 uses analogous augmentation, ImageNet channel
means/standard deviations $(0.485,0.456,0.406)/(0.229,0.224,0.225)$,
and 1001 labels in the retained mirror. Its measurement batch is the first
32 rows of the validation split of
\texttt{benjamin-paine/imagenet-1k-32x32}, in retained mirror order, with
normalization but no augmentation. Stage A's SGD uses momentum 0.9, weight decay
$5\times10^{-4}$, gradient-norm clipping at one and cosine annealing. These
optimizer operations are excluded from the measured data-loss Hessian. Image maps use
evaluation mode and stored BatchNorm statistics; step diagnostics instead use
training-mode batch statistics.

\paragraph{A1 architecture and partitions} On \(32\times32\)
RGB images, ResNet-18 retains the unmodified stem: a bias-free \(7\times7\)
convolution, 3 to 64 channels, stride 2, padding 3, followed by BatchNorm,
ReLU and \(3\times3\) max-pooling, stride 2, padding 1. Its four stages have
widths 64/128/256/512 and two BasicBlocks each, with two bias-free \(3\times3\)
main-path convolutions (padding 1) per block. The first block of stages 2--4
uses stride 2 in its first convolution and a \(1\times1\), stride-2
convolution--BatchNorm shortcut projection; other shortcuts are identities.
Adaptive \(1\times1\) average pooling, flattening and a biased 512-to-100 head
complete the model. Plain-18 removes every residual addition and all three
projection submodules and their parameters, while retaining the main-path
strides and Conv--BN--ReLU--Conv--BN--ReLU blocks. It is not a CIFAR-specific
\(3\times3\)-stem/no-pool variant. The measured objective is mean 100-class
cross-entropy, without optimizer decay or damping.

The partition comparison in \ref{app:extended-mass} uses the final saved weights of the 300-epoch
schedule, 3000 probes and the same ordered 32-image batch, not the original
100-epoch/30-probe map. Its fine A1 rule traverses named modules in preorder,
assigning one group per \texttt{Conv2d} or \texttt{Linear} module; trainable
parameters are assigned by identity to the first module that contains them,
and all unassigned parameters form one pooled \emph{rest} group. Thus ResNet has
17 main-path convolution groups, 3 projection-convolution groups, 1 classifier
and 1 rest, totalling 22; Plain has \(17+1+1=19\). All trainable BatchNorm
scales/shifts are in rest, including projection BatchNorm in ResNet; running
statistics and counters are buffers, not parameter coordinates. The six coarse
groups instead own the stem convolution and its BatchNorm, each complete stage
with its BatchNorm and any projections, and the head. This pooled cross-stage
rest makes the fine partition non-nested. The fine-group names, order and counts are recorded. Their parameter sizes
are derived from the construction because the original records do not contain
these sizes. The specifications describe the saved implementations; their
historical torchvision version is unrecorded.

For the original A1 comparison of the six-group and layer partitions,
restore paths and matching validation summaries support reuse of saved
checkpoints, but weight identity cannot be verified because measurement-time
checksums were not recorded. \ref{app:extended-mass} reports the independently
preserved 3000-probe comparison. Additional layer-budget and redraw analyses
are omitted because their primary scalar inputs are unavailable.

The toy decoders are pre-norm causal transformers with learned token and
position embeddings, GELU MLPs and no training. Tiny/small/base have respectively
$(d,\text{layers},\text{heads},d_{\mathrm{ff}})=(64,2,2,256)$,
$(128,4,4,512)$ and $(256,6,8,1024)$, vocabulary 512 and position capacity 512.
Their output weights are untied. Model construction uses seed zero; a separate
seed-zero generator draws independent uniform token IDs, eight sequences of
length 64. The loss averages next-token cross-entropy over the batch and the
63 shifted positions, not over unshifted input targets.

For \texttt{Qwen/Qwen2.5-0.5B} and
\texttt{meta-llama/Llama-3.2-1B}, the tokenizer uses the same repository ID as
the model. The recorded decoder protocol specifies evaluation mode, float32, eight probes,
and respectively four and two sequences of length 128, without training.
The input consists of the following four author-written sentences:
\begin{enumerate}[label=(\arabic*),leftmargin=1.6em,itemsep=2pt,topsep=2pt]
  \item The curvature of a deep network describes how its loss bends as the
    weights move.
  \item A block-diagonal preconditioner keeps each layer's own curvature and
    discards the coupling between layers.
  \item Attention ties queries, keys, and values together, so the curvature
    shared across those projections is exactly what a layer-wise approximation
    throws away.
  \item Measuring this discarded fraction turns an architectural intuition
    into a number one can track as a model grows.
\end{enumerate}
The sentences are joined with a single ASCII space and tokenized once.
The resulting token-ID stream is repeated only if necessary, and its first
\(B\cdot128\) IDs are reshaped into \(B\) sequences of length 128 without
padding (\(B=4\) for Qwen, \(2\) for Llama). The implementation uses no chat
template or explicit special-token override. Historical token IDs, resolved
model/tokenizer revisions and the tokenizer version are unavailable, so the
recipe does not identify the exact original token batch. The fixed repeated
text is not a representative language corpus. Mean shifted next-token
cross-entropy covers \(B\cdot127\) prediction positions. Shared input/output
weights belong only to the embedding group, leaving the final normalization
in the head: 896 parameters for Qwen and 2048 for Llama. This ownership rule
is part of the measured partition.

On the two frozen language models the coupling field is measured on a subset of
the pairs, written \emph{band 8} in Table~\ref{tab:stage-a-config}: every pair
inside one layer, together with every pair at most eight parameter groups apart
in group order. The restriction is on the field alone. $\Phi$ and the diagonal
block norms cover every group, and for these two rows $\|\mathbf{H}_{\boldsymbol{\theta}}\|_F^2$ comes
from the separate full-parameter probe described in \ref{app:extended-methods}.

\subsection{Correction experiment configuration (Stage B)}
\label{app:stage-b-config}

Table~\ref{tab:stage-b-config} lists the Stage B hyperparameters.

\begin{table}[htbp]
  \centering
  \caption{Stage B hyperparameters.  The correction row gives the parameters of
    the correction of Section~\ref{sec:correction}; EKFAC is the
    eigenvalue-corrected variant of K-FAC.}
  \label{tab:stage-b-config}
  \scriptsize
  \begin{tabular}{@{}lp{0.8\textwidth}@{}}
    \toprule
    Task & ResNet-50 (ImageNet weights) on Stanford Cars, $196$ classes, $224$px, flip/crop augmentation \\
    Budget / first order & 30 epochs, batch 64; SGD $\eta=10^{-2}$, momentum $0.9$; Adam $\eta=10^{-3}$; first-order baselines use cosine, weight decay $10^{-4}$ and gradient-norm clipping at one \\
    Second order & step $\eta=1$; K-FAC / EKFAC (curvlinops, Monte-Carlo Fisher,
    one sampled label per example$^{\dagger}$), damping
    $\lambda=1$$^{\ddagger}$, added to each Kronecker factor in K-FAC,
    $(\mathbf{A}+\lambda \mathbf{I})\otimes(\mathbf{G}+\lambda \mathbf{I})$, and to the product in EKFAC,
    $\hat{\mathbf{K}}+\lambda \mathbf{I}$; $T_k=20$ (K-FAC), $10$ (EKFAC); parameters
    without Kronecker factors (BatchNorm affine) take the gradient step,
    so the actual $\mathbf{B}$ of~\eqref{eq:residual} is the identity on them.
    The second-order hooks use no optimiser decay, gradient clipping or
    learning-rate scheduler \\
    Correction & $\tau=0.1$, $\ell=4$, $n_{\mathrm{it}}=8$, $s_{\max}=3$, $T=20$, $m_c=10$, $c_{\mathrm{step}}=2$; damping $\lambda$ shared with the base step \\
    Oracle & Newton-CG on $(\mathbf{H}+\lambda \mathbf{I})\mathbf{d}=-\mathbf{g}$, stopped at 20 iterations, at an
    absolute residual norm below $10^{-6}$, or on a direction of non-positive
    curvature (including zero); at this stop it returns $-\mathbf{g}$ if the iterate
    is zero, otherwise the current iterate, not an optimum \\
    Measurements & every $T$ steps, in training mode on the step's own 64 training images: $m(\mathbf{d}_0)$, raw $m(\mathbf{d}_1)$, applied $m(\mathbf{d})$, $m(\mathbf{d}_{\mathrm{N}})$, $L_{\mathrm{mb}}(\mathbf{d}_0)$, $L_{\mathrm{mb}}(\mathbf{d}_1)$, $\varepsilon(v,w)$, $\mathcal{C}(v,w)$; safeguard verdicts by test; $\Phi$ at init, mid, final, in evaluation mode ($m=30$, 32 validation images) \\
    Target & epochs to $90\%$ validation accuracy \\
    \bottomrule
    \multicolumn{2}{@{}p{0.97\textwidth}}{$^{\dagger}$The exact expectation over the
      model's predictive distribution costs one backward pass per class on Stanford
      Cars, $196$ per refresh; it is therefore replaced throughout by the
      one-sample Monte-Carlo Fisher of \citet{martens2015optimizing}.
      $^{\ddagger}$The historical protocol records a 30-step K-FAC damping screen
      at batch $64$, seed 42, over $\lambda\in\{10^{-1},1,10\}$, fixed before
      the validity radius was defined. A choice was rejected if $m(\mathbf{d}_0)>0$ on
      more than a fifth of the records or the actual-to-predicted reduction left
      $[\tfrac14,4]$; only $\lambda=1$ passed this screening rule. It is a
      historical proxy, not a model-agreement test: $m(\mathbf{d}_0)>0$ alone merely
      predicts a loss increase. Primary records of the historical damping
      screen are unavailable. The same $\lambda$ is used by
      the correction and by the Newton-CG oracle.} \\
  \end{tabular}
\end{table}

The pretrained ResNet-50 in A2 and Stage B uses
\path{torchvision.models.ResNet50_Weights.IMAGENET1K_V2}, with a fresh
196-class head; no measurement-time weight checksum or complete older
library-version metadata was recovered. Stanford Cars uses a deterministic
validation subset of the 8144 labelled training images, not the official test
set. In sorted class order, one \texttt{random.Random(0)} stream shuffles the
indices in \texttt{cars\_train\_annos.mat} record order. For each non-singleton
class, the first \(\min(\max(1,\lceil0.2n_c\rceil),n_c-1)\) entries form
validation; selected original indices are then sorted. The first 32 entries
of that subset supply A2 and Stage B's evaluation-mode \(\Phi\) batch. Training
uses 224-pixel random resized crops (scale 0.5--1) and horizontal flips;
validation resizes to 256 then centre-crops to 224, with ImageNet normalization.
The selection rules were reconstructed from the retained source. The image
tensors, dataset revision and checksums, and a Cars filename list are unavailable.

\paragraph{Candidate trial state}
Within each Stage B record, curvature and candidate losses use the same
augmented, device-resident tensors of 64 images and labels
(\texttt{ctx.x}, \texttt{ctx.y}) before the live update. The trials therefore
share their inputs without refetching or re-augmenting them. Native loss is
mean cross-entropy without damping. Each functional trial disables gradient
recording, uses detached displaced parameter copies \(\boldsymbol{\theta}+\mathbf{d}\), and clones
all live buffers while preserving training mode. Both candidates start from
the same buffers; BatchNorm uses candidate-specific batch statistics, and
trial changes to running means, variances and counters are discarded. Buffer
isolation is specific to these trials and does not cover every live gradient
or provider forward. The helper does not replay RNG state, although the
configured ResNet-50 has no dropout layer. Under the finite same-state and
exact-comparison conditions of Proposition~\ref{prop:no-harm}, the safeguard tests
native loss after the norm and damped-model checks and otherwise uses \(\mathbf{d}_0\).
Complete historical buffer and RNG snapshots for individual records are unavailable.

\paragraph{Cohorts and historical provenance}
Stage B's corrected and baseline methods follow separate training trajectories.
Only the two candidate displacements within a record share the same state;
this pairing does not extend to comparisons of training outcomes. The historical loader
layout affects K-FAC and EKFAC baseline seeds 43/44, SGD seeds 43/44 and all
Adam baseline seeds. Persistent worker augmentation state carried across
seed/method loops, while the shuffling sampler received a new draw after
reseeding. In the validity arms the
corresponding later-in-process cells are seed 44 of gradient descent and Adam.
Original K-FAC validity seeds and the corrected Stage B shards were recorded
in separate seed processes. Extra probe draws can also alter later data order
relative to a baseline. Thus sharing a seed does not mean sharing an entire
trajectory or all randomness; fixed Monte-Carlo Fisher label sampling is
another shared noise source.

The process assignments and older configurations were reconstructed from
scalar files, source descriptions, job files and contemporary notes; job files
alone do not confirm execution or inclusion in the reported results. Several
older sets of results lack source and environment metadata, and
archived EKFAC implementations differ in eigenbasis precision. The exact
source and precision used at measurement time are unknown. These limitations
restrict
cross-optimizer efficacy comparisons.

\paragraph{Registration and sensitivity}
The four pre-specified criteria use their original thresholds. The validity
and profile analyses and the later frozen-state pilot are separate from these
confirmatory endpoints. \ref{app:validity} reports sensitivity
to the sampled tolerance and extent threshold; \ref{app:coupling-sensitivity}
examines the gate threshold and retrospective coupling associations. The
long-schedule replication was registered separately. Sensitivity to patch
rank, pair cap, refresh period, batch size, Lanczos depth, seed count and
bootstrap resample count was not assessed. The damping calibration supports
the stated configuration rather than a universal damping choice.

\subsection{Retrospective proposal decomposition}
\label{app:proposal-decomposition}

To separate the local quadratic from its damping penalty, define
retrospectively for the same candidate pair
\begin{equation}\label{eq:proposal-decomposition}
  \begin{aligned}
    q(\mathbf{d})&=\mathbf{g}^\top \mathbf{d}+\tfrac12\mathbf{d}^\top \mathbf{H}\mathbf{d},\qquad
    m(\mathbf{d})=q(\mathbf{d})+\tfrac\lambda2\|\mathbf{d}\|^2,\\
    \Delta m&=\Delta q+\Delta_{\mathrm{penalty}},\qquad
    \Delta_{\mathrm{penalty}}=\tfrac\lambda2(\|\mathbf{d}_1\|^2-\|\mathbf{d}_0\|^2).
  \end{aligned}
\end{equation}
Each difference is proposal minus base within a record; the evaluated
minibatch-loss difference is $L(\boldsymbol{\theta}+\mathbf{d}_1)-L(\boldsymbol{\theta}+\mathbf{d}_0)$.
Table~\ref{tab:proposal-decomposition} reports the decomposition in
equation~\eqref{eq:proposal-decomposition}
by seed and arm; its column $\Delta p$ denotes the penalty difference
$\Delta_{\mathrm{penalty}}$. Among loss-rejected raw refresh proposals,
$434/438$ for K-FAC and $434/436$ for EKFAC lower $q$ according to the
unrounded logged scalars.

Figure~\ref{fig:proposal-model-loss} displays the same 450 raw refresh records
per method, 150 per seed, regardless of eligibility for the closed gap $\gamma$.
Refreshes within a training trajectory are dependent. Damping remains part of
candidate construction when its penalty is subtracted from the comparison.

\begin{table}[htbp]
  \centering
  \caption{Retrospective same-record proposal decomposition on the
    corrected K-FAC and EKFAC trajectories (ResNet-50, Stanford Cars).
    $N$ counts every retained raw refresh candidate, without gamma-eligibility
    filtering; $R$ counts all rejected refresh candidates, not all training steps.
    $Q^-/R_L$ counts strict $\Delta q<0$ among the $R_L$ loss-rejected candidates.
    The last four columns are medians over those $R_L$ pairs, using the same
    pre-update weights and minibatch within each pair:
    $\Delta m=m(\mathbf{d}_1)-m(\mathbf{d}_0)$ for the damped model,
    $\Delta p=\lambda(\|\mathbf{d}_1\|^2-\|\mathbf{d}_0\|^2)/2$,
    $\Delta q=\Delta m-\Delta p$ for the undamped local quadratic, and
    $\Delta L=L(\boldsymbol{\theta}+\mathbf{d}_1)-L(\boldsymbol{\theta}+\mathbf{d}_0)$ for the native data loss, not a model value.
    Here $\mathbf{d}_1$ is the proposal even when rejected. K-FAC's legacy records omit $\lambda$: its stated fixed historical protocol supplies $\lambda=1$, not current software defaults; EKFAC uses each record's damping.
    Subtracting the penalty does not remove damping from candidate construction.
    Counts use unrounded logged-scalar signs; rounded medians do not define them.}
  \label{tab:proposal-decomposition}
  \scriptsize
  \setlength{\tabcolsep}{3pt}
  \begin{tabular}{@{}llrrccccc@{}}
    \toprule
    Arm & Seed & $N$ & $R$ & $Q^-/R_L$ & median $\Delta m$ & median $\Delta p$ & median $\Delta q$ & median $\Delta L$ \\
    \midrule
    K-FAC & 42 & 150 & 146 & 144/146 & $-2.22{\cdot}10^{0}$ & $-3.90{\cdot}10^{-2}$ & $-2.08{\cdot}10^{0}$ & $4.00{\cdot}10^{-3}$ \\
    K-FAC & 43 & 150 & 147 & 147/147 & $-9.32{\cdot}10^{-1}$ & $-3.00{\cdot}10^{-2}$ & $-9.19{\cdot}10^{-1}$ & $3.43{\cdot}10^{-3}$ \\
    K-FAC & 44 & 150 & 145 & 143/145 & $-1.16{\cdot}10^{0}$ & $-3.50{\cdot}10^{-2}$ & $-1.09{\cdot}10^{0}$ & $3.39{\cdot}10^{-3}$ \\
    K-FAC & pooled & 450 & 438 & 434/438 & $-1.19{\cdot}10^{0}$ & $-3.52{\cdot}10^{-2}$ & $-1.15{\cdot}10^{0}$ & $3.54{\cdot}10^{-3}$ \\
    \midrule
    EKFAC & 42 & 150 & 146 & 145/146 & $-3.72{\cdot}10^{0}$ & $-3.48{\cdot}10^{-1}$ & $-3.06{\cdot}10^{0}$ & $7.06{\cdot}10^{-3}$ \\
    EKFAC & 43 & 150 & 143 & 143/143 & $-4.14{\cdot}10^{0}$ & $-4.10{\cdot}10^{-1}$ & $-3.42{\cdot}10^{0}$ & $6.59{\cdot}10^{-3}$ \\
    EKFAC & 44 & 150 & 147 & 146/147 & $-4.41{\cdot}10^{0}$ & $-3.12{\cdot}10^{-1}$ & $-3.95{\cdot}10^{0}$ & $7.25{\cdot}10^{-3}$ \\
    EKFAC & pooled & 450 & 436 & 434/436 & $-4.19{\cdot}10^{0}$ & $-3.63{\cdot}10^{-1}$ & $-3.74{\cdot}10^{0}$ & $7.06{\cdot}10^{-3}$ \\
    \bottomrule
  \end{tabular}
\end{table}

\subsection{Pre-specified criteria and verdicts}
\label{app:criteria-verdicts}

Table~\ref{tab:gates} lists the four pre-specified criteria, their thresholds
and the observed decisions.

\begin{table*}[t]
  \centering
  \caption{Pre-specified acceptance criteria (a)--(d) and verdicts
    (Section~\ref{sec:results-b}). Criterion~(b) is evaluated on the K-FAC pair;
    criterion~(d) compares the frozen language models with the largest toy
    decoder.}
  \label{tab:gates}
  \scriptsize
  \setlength{\tabcolsep}{4pt}
  \begin{tabular}{@{}llp{3.6cm}p{6.8cm}l@{}}
    \toprule
    Criterion & Name & Requirement & Numbers & Verdict \\
    \midrule
    (a) & Inversion (Stage A) & $\Delta_s>0$ all seeds, mean $\ge0.05$ & mean $\Delta=0.011$ ($\Delta_s$: $0.066$, $-0.025$, $-0.009$); replication (same config, seeds 42--44): $\Delta_s$: $0.007$, $0.020$, $-0.011$; pooled $n=6$: mean $0.008$, 95\,\% CI $[-0.026,\,0.042]$; per-seed means $n=3$: CI $[-0.054,\,0.070]$ & \textbf{fail} \\
    (b) & Payoff (Stage B) & $\gamma\ge0.2$, CI $>0$; epochs to target $\le$ K-FAC (median); $\Delta\mathrm{acc}\ge-0.5$\,pp every seed; $\rho_s>0$, $p<0.01$ & $\gamma=0.00\,[0.00,\,0.00]$ (450 records, $0.67\,\%$ above 1); epochs to target $\infty$ vs $\infty$; $\min_s\Delta\mathrm{acc}=-2.4$\,pp; $\rho_s$: COUPLE-FAC (K-FAC) $0.01$ ($p{=}0.105$), COUPLE-FAC (EKFAC) $0.04$ ($p{<}10^{-3}$), $n=13500$ each & \textbf{fail} \\
    (c) & EKFAC (Stage B) & failed eigenbasis refreshes $<5\,\%$; no NaN val.\ loss & max failed-refresh share $0.000$; NaN in val.\ loss: no & \textbf{pass} \\
    (d) & Transformers (Stage A) & $J\ge2/3$, top-3 pairs matched by role & toy base: $v\to o$ $\times3$; Qwen/Qwen2.5-0.5B: $J=3/3$ ($v\to o$); meta-llama/Llama-3.2-1B: $J=3/3$ ($v\to o$) & \textbf{pass} \\
    \bottomrule
  \end{tabular}
\end{table*}

For criterion~(b), the 95 per cent percentile-bootstrap interval uses $10^4$
resamples of records within seed, with reporting random seed zero. Serial dependence and new trajectories are
not fully represented; the zero interval is conditional, not zero population
uncertainty. Historical Spearman p-values are nominal (\ref{app:coupling-sensitivity}). Criterion~(c) did not
test non-divergence itself, and the registered consequence of its failure was
that the eigenvalue-corrected variant would be reported apart from the main
comparison. The roles registered for criterion~(d) were, within a layer,
$v\to o$, $q\leftrightarrow k$ and $o\to{}$multi-layer perceptron, and, across
layers, multi-layer perceptron${}\to{}$next $q$, or the same role in adjacent
layers.

Criterion~(d) was registered as
$J:=|T_{\mathrm{toy}}\cap T_{\mathrm{LM}}|/3\ge2/3$, where $T_{\mathrm{toy}}$
and $T_{\mathrm{LM}}$ are the three strongest coupled pairs of the largest toy
decoder and of a pretrained language model, and the pairs of the two triples are
matched one to one by structural role. On the toy decoder and on both language
models the three strongest measured pairs are $v\to o$, in three different
layers, so all three are matched within this one role, $J=3/3$ for each
language model, and the criterion is met.

\subsection{Long-schedule replication}

\begin{table}[t]
  \centering
  \caption{The A1 inversion (ResNet-18 vs plain ResNet-18, CIFAR-100) on one
    300-epoch cosine schedule, both architectures, checkpoints \texttt{init}, every
    25~epochs, and \texttt{final} = epoch~300 with the learning rate annealed to 0;
    accuracy and $\Phi$ (300 Hutchinson probes) are means over $n=3$ seeds.
    Each architecture's $\Phi$ is shown with its descriptive standard deviation
    (divisor $n$). For $\Delta\Phi=\Phi_{\mathrm{ResNet}}-\Phi_{\mathrm{Plain}}$
    at the same checkpoint label, the mean and sample standard deviation
    (divisor $n-1$) are taken over the paired within-seed differences. This is a matched-accuracy replication of the inversion: the
    footer pairs each plain checkpoint after initialisation with the ResNet checkpoint
    of the \emph{same seed} nearest in accuracy, kept when
    $|\Delta\mathrm{acc}|\le0.02$, rather than comparing at the same label. The final ResNet $\Phi$ is higher in these runs: $\Phi_{\mathrm{final}}(\mathrm{ResNet})=0.73$ (mean over 3~seeds) against $0.61$--$0.67$ across the individual seeds of the 100-epoch run and its replication (\ref{app:extended-mass}).}
  \label{tab:phi-long}
  \small
  \begin{tabular}{@{}lrccccc@{}}
    \toprule
    Checkpoint & Epoch & Acc.\ ResNet & Acc.\ Plain & $\Phi_{\mathrm{ResNet}}$ & $\Phi_{\mathrm{Plain}}$ & $\Delta\Phi$ \\
    \midrule
    init & 0 & $0.011$ & $0.010$ & $0.832 \pm 0.016$ & $0.837 \pm 0.009$ & $-0.005 \pm 0.011$ \\
    25 & 25 & $0.449$ & $0.307$ & $0.608 \pm 0.015$ & $0.569 \pm 0.016$ & $0.039 \pm 0.035$ \\
    50 & 50 & $0.477$ & $0.319$ & $0.635 \pm 0.020$ & $0.566 \pm 0.007$ & $0.070 \pm 0.016$ \\
    75 & 75 & $0.484$ & $0.361$ & $0.668 \pm 0.009$ & $0.574 \pm 0.007$ & $0.094 \pm 0.018$ \\
    100 & 100 & $0.507$ & $0.355$ & $0.673 \pm 0.007$ & $0.573 \pm 0.011$ & $0.100 \pm 0.007$ \\
    125 & 125 & $0.522$ & $0.398$ & $0.693 \pm 0.006$ & $0.591 \pm 0.012$ & $0.103 \pm 0.022$ \\
    mid & 150 & $0.538$ & $0.429$ & $0.701 \pm 0.001$ & $0.581 \pm 0.006$ & $0.120 \pm 0.008$ \\
    175 & 175 & $0.556$ & $0.467$ & $0.715 \pm 0.008$ & $0.576 \pm 0.010$ & $0.139 \pm 0.020$ \\
    200 & 200 & $0.572$ & $0.500$ & $0.726 \pm 0.006$ & $0.582 \pm 0.008$ & $0.144 \pm 0.015$ \\
    225 & 225 & $0.583$ & $0.549$ & $0.732 \pm 0.006$ & $0.605 \pm 0.008$ & $0.128 \pm 0.003$ \\
    250 & 250 & $0.603$ & $0.585$ & $0.737 \pm 0.002$ & $0.622 \pm 0.011$ & $0.115 \pm 0.016$ \\
    275 & 275 & $0.626$ & $0.610$ & $0.729 \pm 0.004$ & $0.640 \pm 0.017$ & $0.090 \pm 0.025$ \\
    final & 300 & $0.633$ & $0.615$ & $0.728 \pm 0.004$ & $0.655 \pm 0.016$ & $0.072 \pm 0.015$ \\
    \bottomrule
    \multicolumn{7}{@{}p{0.95\linewidth}}{\footnotesize Accuracy-matched pairs ($|\Delta\mathrm{acc}|\le0.02$): $19$ pairs, mean $\Delta\Phi=0.086$, 95\,\% $t$-CI $[0.060,\,0.112]$ over 3~seeds.}
  \end{tabular}
\end{table}

Both architectures of the A1 pair were trained again on one three-hundred-epoch
cosine schedule, with $\Phi$ measured at thirteen checkpoints and on the same
three seeds (Table~\ref{tab:phi-long}). The analysis pairs each
skip-free checkpoint after initialization with the residual checkpoint of the
\emph{same seed} nearest to it in accuracy, keeps a pair only when the two
accuracies agree within the tolerance the caption states, and averages the
difference over pairs inside a seed before taking an interval over the three
seeds. The tolerance was fixed at $\LongTolWorking$ before the run.
Sweeping it from $\LongTolLo$ to $\LongTolHi$ through the same accuracy-matched
difference keeps the mean between $\LongDeltaPhiTolLo$ and $\LongDeltaPhiTolHi$,
and the sign never changes.

A cosine schedule set for three hundred epochs has not annealed its
learning rate at epoch one hundred, where the hundred-epoch schedule reported in
\ref{app:extended-mass} has, so the checkpoints that share a label are different
points of training.

The replication was registered on 21 September 2026, before execution and
independently of criterion~(a). That criterion remains evaluated on its
specified hundred-epoch schedule, with the decision in Table~\ref{tab:gates}.

\subsection{Retrospective coupling sensitivity}
\label{app:coupling-sensitivity}

The raw ordered association between coupling $\mathcal{C}(v,w)$ and the
cross-term ratio $\varepsilon(v,w)$ is small: Spearman's $\rho_s$ is
$\SpearmanKfac$ ($p=\SpearmanKfacP$) for K-FAC and $\SpearmanEkfac$
($p=\SpearmanEkfacP$) for EKFAC. The $\SpearmanObs$ observations comprise
$\SpearmanPairsPerRecord$ ordered pairs at each of $\RefreshRecords$
refreshes. They are dependent, making the reported historical p-values
nominal rather than valid iid inference. EKFAC's nominal p-value is below the
registered threshold; criterion~(b) concerns K-FAC. Figure~\ref{fig:eps-vs-C}
also shows the difference in targets: $\mathcal C$ is symmetric, whereas
$\varepsilon$ is ordered and spans three orders of magnitude. For
$(v,w)=(\mathrm{stem},\mathrm{third\ stage})$, the ratio is
$\EpsStemLayerThree$; the reverse entry is $\EpsLayerThreeStem$.
Here $v$ receives action from source $w$.

Retrospective comparisons find positive associations on every seed in both
arms, using either the symmetric target
$\sqrt{\varepsilon(v,w)\varepsilon(w,v)}$ or residuals of both log variables
after receiver/source adjustment within each refresh. The per-seed results
appear beside the raw ordered associations in
Table~\ref{tab:coupling-sensitivity}, without new
p-values or changes to the registered criteria. Shared block magnitude can
contribute to these associations, so their positivity establishes neither
useful top-three selection nor training gain.

The $\SpearmanObs$ K-FAC pair measurements, made in training mode on each
step's own minibatch, have median coupling $\CouplingFieldMedian$ and maximum
$\CouplingFieldMax$. The maximum occurs at step zero on a pair involving
the new head. These pooled pair statistics describe the selector's inputs;
they do not evaluate the weighted mean of $\mathcal C^2$ for a single operator
in \ref{app:extended-mass}.

The threshold excludes no measured pair. Among $\CouplingObsBothArms$
ordered couplings across both arms, $\CouplingBelowTau$ are at or below it;
the minima are $\CouplingFieldMinKfac$ for K-FAC and
$\CouplingFieldMinEkfac$ for EKFAC. Selection therefore operates through
ranking and the cap: a rank-$\ell$ patch is applied to at most
$s_{\max}=3$ of fifteen group pairs, in descending order of $\mathcal C$.
Because this ranking normalizes by diagonal norms, it need not select the
pairs carrying most off-block mass. Restored mass coverage was not recorded
at refreshes. With the cap fixed, the threshold can affect selection only
above the fourth-largest coupling at a refresh. Its minimum across refreshes
is $\CouplingFourthMinKfac$ for K-FAC and $\CouplingFourthMinEkfac$ for
EKFAC; every threshold below $\CouplingFourthMinEkfac$ leaves all selections
unchanged.

\FigureEpsVsC

A separate pooled fit describes the scale of $\log\varepsilon$ alone.
The ratio contains the receiver-specific term
$-\log\|\mathbf{H}_{\boldsymbol{\theta}_v,\boldsymbol{\theta}_v}\mathbf{d}_{0,v}\|$, independent of $w$. On K-FAC,
least squares with an intercept and row and column effects explains
$\EpsAdditiveShare$ per cent of its pooled log-spread. Unlike the
within-refresh two-variable adjustment, this fit does not separate
denominator scale from systematic numerator structure associated with the
same receiver or source. Its explained share therefore does not bound all
information in cross-block structure.

Table~\ref{tab:coupling-sensitivity} presents a retrospective analysis of the
Stage B refresh records. Raw associations retain all finite
ordered $(\mathcal C,\varepsilon)$ entries without a positivity filter.
For the symmetric target, one observation per unordered pair at each refresh is retained only
when both orientations are present, finite and non-negative with matching
$\mathcal C$; the target is
$\sqrt{\varepsilon(v,w)\varepsilon(w,v)}$, including zero. The adjusted domain uses the same strictly
positive finite cells for both variables. Separate OLS fits of $\log\mathcal C$
and $\log\varepsilon$ use an intercept and categorical receiver/source effects
within each refresh. Residual observations are pooled only within seed before
Spearman ranks are calculated: neither refresh correlations nor seeds are
averaged into this statistic. The pooled fit above adjusts only $\log\varepsilon$.

The table lists observations and exclusions for each domain. No observations
are excluded in these data, and the gamma eligibility filter is not applied.
A correlation is undefined, and printed as ---, when there are fewer than
three observations or a constant variable. The adjustment uses the recorded
observation order and least-squares convention, setting numerically exact-fit
residuals to zero. Equivalent floating-point projections or reordered inputs
can affect ranks, although the checked alternatives agree at the reported
three-decimal precision.

\begin{table}[t]
  \centering
  \caption{Retrospective coupling sensitivity, per seed: Spearman associations
    pooled within seed.
    Raw: finite ordered $(\mathcal{C},\varepsilon)$, including zeros.
    Symmetric: one observation per unordered pair at each refresh, comparing $\mathcal{C}$ with
    $\sqrt{\varepsilon(v,w)\varepsilon(w,v)}$; both orientations must be present
    with finite non-negative values and matching $\mathcal{C}$.
    Adjusted: finite strictly positive $(\mathcal{C},\varepsilon)$; both log variables
    are fitted by OLS on intercept + receiver + source per refresh, followed by pooled residual ranks
    within seed. No gamma eligibility filter is used. $R$ counts refresh records;
    $N$ and excluded counts are raw/symmetric/adjusted observations.
    Undefined correlations (fewer than three observations or constant values) read ---.}
  \label{tab:coupling-sensitivity}
  \footnotesize
  \setlength{\tabcolsep}{3pt}
  \begin{tabular}{@{}llrcccll@{}}
    \toprule
    Method & Seed & $R$ & Raw & Symmetric & Adjusted & $N$ & Excluded \\
    \midrule
    COUPLE-FAC (K-FAC) & 42 & 150 & $0.013$ & $0.201$ & $0.266$ & 4500/2250/4500 & 0/0/0 \\
    COUPLE-FAC (K-FAC) & 43 & 150 & $0.015$ & $0.209$ & $0.259$ & 4500/2250/4500 & 0/0/0 \\
    COUPLE-FAC (K-FAC) & 44 & 150 & $0.014$ & $0.229$ & $0.248$ & 4500/2250/4500 & 0/0/0 \\
    COUPLE-FAC (EKFAC) & 42 & 150 & $0.036$ & $0.357$ & $0.521$ & 4500/2250/4500 & 0/0/0 \\
    COUPLE-FAC (EKFAC) & 43 & 150 & $0.040$ & $0.408$ & $0.554$ & 4500/2250/4500 & 0/0/0 \\
    COUPLE-FAC (EKFAC) & 44 & 150 & $0.030$ & $0.365$ & $0.536$ & 4500/2250/4500 & 0/0/0 \\
    \bottomrule
  \end{tabular}
\end{table}

\section{Extended profile and partition measurements}
\label{app:extended-mass}

Figure~\ref{fig:phi-map} summarizes the architectural ladder;
Table~\ref{tab:phi-map} gives the fractions and both references at each checkpoint.

\FigurePhiMap

\begin{table}[t]
  \centering
  \caption{Stage A: discarded-coupling fraction $\Phi$ per run and architecture at the
    three checkpoints (mean$\,{\pm}\,1\sigma$ over $n$~seeds), against
    $\Phi_{\mathrm{null}}$ - the group-size reference, fixed per architecture -
    and, in parenthesis next to each measured $\Phi$,
    $\Phi_{\mathrm{sep}}$: the matched separable reference computed from that
    checkpoint's own diagonal block norms, which moves with the weights and is not a
    null. Both are computed per seed before averaging. For the band-estimated
    decoder rows, the full norm is estimated separately. For frozen transformers, weights and inputs are fixed;
    only the Hutchinson probes vary across seeds. Checkpoints not applicable to a run read ---.}
  \label{tab:phi-map}
  \footnotesize
  \setlength{\tabcolsep}{2pt}
  \begin{tabular}{@{}llccccc@{}}
    \toprule
    Run & Model & $\Phi_{\mathrm{null}}$ & $\Phi_{\mathrm{init}}$ & $\Phi_{\mathrm{mid}}$ & $\Phi_{\mathrm{final}}$ & $n$ \\
    \midrule
    A1 CIFAR-100 & ResNet-18 & $0.404$ & $0.832 \pm 0.016$ ($0.745$) & $0.662 \pm 0.011$ ($0.771$) & $0.640 \pm 0.025$ ($0.698$) & 3 \\
     & Plain-18 & $0.404$ & $0.834 \pm 0.011$ ($0.284$) & $0.581 \pm 0.015$ ($0.765$) & $0.629 \pm 0.024$ ($0.683$) & 3 \\
    \midrule
    A1 width/depth & ResNet-18 ($0.5\times$) & $0.410$ & $0.833 \pm 0.025$ ($0.757$) & $0.666 \pm 0.027$ ($0.781$) & $0.640 \pm 0.004$ ($0.712$) & 3 \\
     & ResNet-18 & $0.404$ & $0.833 \pm 0.016$ ($0.744$) & $0.645 \pm 0.007$ ($0.769$) & $0.632 \pm 0.003$ ($0.689$) & 3 \\
     & ResNet-18 ($2\times$) & $0.400$ & $0.835 \pm 0.011$ ($0.736$) & $0.614 \pm 0.012$ ($0.752$) & $0.606 \pm 0.009$ ($0.690$) & 3 \\
     & ResNet-34 & $0.517$ & $0.667 \pm 0.014$ ($0.730$) & $0.604 \pm 0.017$ ($0.769$) & $0.636 \pm 0.011$ ($0.715$) & 3 \\
    \midrule
    A1 ImageNet-32 & ResNet-18 & $0.448$ & $0.841 \pm 0.002$ ($0.736$) & $0.646 \pm 0.008$ ($0.789$) & $0.648 \pm 0.006$ ($0.748$) & 3 \\
     & Plain-18 & $0.449$ & $0.844 \pm 0.006$ ($0.270$) & $0.476 \pm 0.020$ ($0.710$) & $0.508 \pm 0.020$ ($0.738$) & 3 \\
    \midrule
    A2 pretrained & ResNet-50, new random head & $0.517$ & $0.927 \pm 0.003$ ($0.695$) & --- & --- & 5 \\
    \midrule
    A3 toy ladder & tiny (198k) & $0.803$ & $0.583 \pm 0.002$ ($0.649$) & --- & --- & 3 \\
     & small (988k) & $0.902$ & $0.625 \pm 0.000$ ($0.794$) & --- & --- & 3 \\
     & base (5.1M) & $0.930$ & $0.651 \pm 0.000$ ($0.855$) & --- & --- & 3 \\
    \midrule
    A3 LLM ladder & Qwen/Qwen2.5-0.5B (0.5B) & $0.907$ & $0.679 \pm 0.034$ ($0.802$) & --- & --- & 3 \\
     & meta-llama/Llama-3.2-1B (1.2B) & $0.928$ & $0.814 \pm 0.011$ ($0.887$) & --- & --- & 3 \\
    \bottomrule
    \multicolumn{7}{@{}p{0.95\linewidth}}{\footnotesize A1 $\Delta_s=\Phi_{\mathrm{ResNet}}-\Phi_{\mathrm{Plain}}$ per seed final: $0.066$, $-0.025$, $-0.009$ (mean $0.011$); replication: $0.007$, $0.020$, $-0.011$; pooled $n=6$: mean $0.008$, 95\,\% CI $[-0.026,\,0.042]$; per-seed means $n=3$: CI $[-0.054,\,0.070]$; mid: $0.080$, $0.099$, $0.064$ (mean $0.081$); replication: $0.087$, $0.116$, $0.079$; pooled $n=6$: mean $0.088$, 95\,\% CI $[0.068,\,0.107]$; per-seed means $n=3$: CI $[0.042,\,0.133]$. ImageNet-32 A1 $\Delta_s$ per seed, mid: $0.164$, $0.197$, $0.149$ (mean $0.170$).}
  \end{tabular}
\end{table}

At the saved final checkpoints of the 300-epoch schedule, the measurements
in Table~\ref{tab:two-partitions} separate the factors in
$1-\Phi=S\,(1-\Phi_{\mathrm{sep}})$, equation~\eqref{eq:s-identity}. Their
differences are on different scales and do not add to $\Delta\Phi$. At the
six-group partition, the residual-minus-Plain difference is
$\Delta\Phi=\FinalDeltaPhiStage$, with 95 per cent three-seed interval
$[\FinalDeltaPhiStageLo,\ \FinalDeltaPhiStageHi]$.
The difference in $\Phi_{\mathrm{sep}}$ is $\FinalDeltaSepStage$, interval
$[\FinalDeltaSepStageLo,\ \FinalDeltaSepStageHi]$, while that in $\ln S$ is
$\FinalDeltaLnSStage$, interval
$[\FinalDeltaLnSStageLo,\ \FinalDeltaLnSStageHi]$, which contains zero.
At the layer partition, $\Delta\Phi=\FinalDeltaPhiLayer$ is several times
smaller. The profile difference is $\FinalDeltaSepLayer$, interval
$[\FinalDeltaSepLayerLo,\ \FinalDeltaSepLayerHi]$, and the $\ln S$ difference
is $\FinalDeltaLnSLayer$, interval
$[\FinalDeltaLnSLayerLo,\ \FinalDeltaLnSLayerHi]$, again containing zero.

The largest $\Phi$ in the map belongs to ResNet-50 before fine-tuning, with
ImageNet weights in the body and a fresh random classification head per seed.
Its value, $\PhiPretrainedRfifty$, exceeds both references,
$\PhiNullRfifty$ and $\PhiSepRfifty$. Across five snapshots, head-incident
pairs carry $\HeadShareRfiftyLo$ to $\HeadShareRfiftyHi$ per cent of off-block
squared mass; on the five body groups, estimated $S$ ranges from
$\SNoHeadRfiftyLo$ to $\SNoHeadRfiftyHi$. In evaluation mode on thirty-two
validation images, the largest couplings involve the head:
$\CouplingStemHeadRfifty$ with the stem and $\CouplingLayerOneHeadRfifty$
with the first stage. These incident-pair shares account for mass in the snapshots;
without matched random and trained heads, they do not identify an effect of
head randomness or certify a positive semidefinite body.

The middle checkpoints distinguish the two families more clearly. The
difference $\Phi_{\mathrm{ResNet}}-\Phi_{\mathrm{Plain}}$ is positive on every
seed on both datasets and in the CIFAR-100 replication. On CIFAR-100 its
three-seed mean is $\GateAMidMeanDelta$; pooling the run and replication gives
$\GateAMidPooledMean$ with a nominal 95 per cent interval
$[\GateAMidPooledLo,\ \GateAMidPooledHi]$ over six values. This calculation
treats those values as independent, although same-seed runs are not six
independent training units. Table~\ref{tab:phi-map} gives the per-seed values
and the more conservative interval over three seed means.

The final CIFAR-100 checkpoint does not meet criterion~(a). The residual and
skip-free means are $\PhiResnetFinalCifar$ and $\PhiPlainFinalCifar$,
respectively; the differences are not positive on every seed and their mean,
$\GateAMeanDelta$, is below the pre-specified threshold. The interval over
three seed means, each averaging the run and its replication, is
$[\GateASeedLo,\ \GateASeedHi]$. The verdict follows from the registered
criterion rather than this interval. On ImageNet-32 the final residual mean
is $\PhiResnetFinalInet$, against $\PhiPlainFinalInet$ for Plain-18, with
positive differences $\InetDeltaFinalOne$, $\InetDeltaFinalTwo$ and
$\InetDeltaFinalThree$ in seed order. At this thirty-epoch budget, top-one
accuracy is $\AccPlainFinalInet$ per cent for Plain-18 and
$\AccResnetFinalInet$ per cent for ResNet-18. This fixed-budget comparison is
distinct from the criterion registered for CIFAR-100.

A separately registered replication compares the architectures at matched
accuracy on a three-hundred-epoch schedule, measured at thirteen checkpoints.
The matching protocol is described in \ref{app:protocol}. Final top-one
accuracies are
$\LongAccResnetFinal$ and $\LongAccPlainFinal$ per cent. Across $\LongPairs$
matched pairs, the mean difference is $\LongDeltaPhi$, with a 95 per cent
interval $[\LongDeltaPhiLo,\ \LongDeltaPhiHi]$ over three seeds
(Table~\ref{tab:phi-long}, \ref{app:protocol}). Thus a component of the
inversion survives accuracy matching. In these runs, the residual network
finishes at $\LongPhiResnetFinal$ (mean over three seeds), above every
individual seed of the hundred-epoch run and its replication, whose range is
given in that table's caption.

Width has little effect on initial $\Phi$ across the three eighteen-layer
widths, but a larger effect at the later checkpoints. The final residual rows
of the map---ResNet-18 on both datasets, its three widths and ResNet-34---span
$\PhiTrainedResNetLo$ to $\PhiTrainedResNetHi$, against size references from
$\PhiNullResNetLo$ to $\PhiNullResNetHi$. The depth comparison is clearer
relative to that reference: final excess over $\Phi_{\mathrm{null}}$ is about
twice as large for ResNet-18 as for ResNet-34. Parameters are distributed more
evenly over ResNet-34's six groups, raising its reference without a comparable
rise in $\Phi$ (Table~\ref{tab:phi-map}).

The decoder measurements lie below both references. At random initialization,
the three toy decoders have $\Phi$ between $\PhiToyLo$ and $\PhiToyHi$, against
$\Phi_{\mathrm{null}}$ between $\PhiNullToyLo$ and $\PhiNullToyHi$. The frozen
pretrained models have values $\PhiQwen$ and $\PhiLlama$, against
$\PhiNullQwen$ and $\PhiNullLlama$. All five also lie below their profile
references at the grouping specified in Section~\ref{sec:setting}. A
per-module partition with separate normalization groups refines this grouping.
For the exact operator, the coarser partition's $\Phi$ is therefore a lower
bound on that of the refinement (Section~\ref{sec:phi}). The size of the granularity effect is measured
on ResNet-50 rather than on these decoders. In all five transformers the
strongest within-layer pair is the value and output projections. The three
strongest measured pairs of each pretrained model and the largest toy model
are instances of this pair in three layers. Criterion~(d) is therefore met
with $J=3/3$ for both pretrained models, entirely within one structural role
(\ref{app:protocol}).

The profile references put these observations into a common accounting.
Of the map's $\PhiCellsTotal$ cells of seed means,
$\PhiCellsTrainedWeights$ have trained weights: the middle and final
convolutional checkpoints and the two pretrained language models.
Another $\PhiCellsRandomInit$ have random weights: the initial convolutional
checkpoints and three toy decoders. The remaining cell combines a pretrained
ResNet-50 body with a random head. Mean $\Phi$ lies below mean
$\Phi_{\mathrm{sep}}$ in $\PhiCellsTrainedBelowSep$ of the
$\PhiCellsTrainedWeights$ trained-weight cells. For these arithmetic seed
means, equation~\eqref{eq:s-identity} gives
$\overline{\Phi_{\mathrm{sep}}}-\overline{\Phi}
=\overline{(1-\Phi_{\mathrm{sep}})(S-1)}$.
The sign of this weighted gap is not equivalent to the arithmetic mean of
$S$ exceeding one, nor does it establish $S>1$ on every seed. For an individual
matrix, $\Phi<\Phi_{\mathrm{sep}}$ places aggregate off-block squared mass
below the positive semidefinite envelope at the same diagonal amplitudes;
it neither establishes positive semidefiniteness nor constrains every pair.
For an exact full matrix with defined couplings, $\Phi>\Phi_{\mathrm{sep}}$
means that the unordered-pair mean of $\mathcal C(v,w)^2$, weighted by
$\|\mathbf{H}_{vv}\|_F\|\mathbf{H}_{ww}\|_F$, exceeds one; the reverse inequality means it
falls below one. For estimates, this relation requires complete pair components
using the same norm decomposition. It is not a sample-wise identity for the
band-estimated decoder rows, whose full norm is estimated separately.

\begin{table}[t]
  \centering
  \caption{The A1 pair (ResNet-18 and its skip-free twin, Plain-18, CIFAR-100) at
    the six-group partition and at the layer partition: group count, $\Phi$,
    $\Phi_{\mathrm{sep}}$ and $S$ of~\eqref{eq:s-identity} (mean over
    3~seeds), with $S$'s 95\,\% $t$-interval over the three per-seed values.
    Measured on the saved checkpoints of the 300-epoch schedule, not
    retrained. 3000~Hutchinson probes.}
  \label{tab:two-partitions}
  \footnotesize
  \setlength{\tabcolsep}{4pt}
  \begin{tabular}{@{}lccccc@{}}
    \toprule
    Model & Groups & $\Phi$ & $\Phi_{\mathrm{sep}}$ & $S$ & $S$ 95\,\% CI \\
    \midrule
    ResNet-18 (six groups) & $6$ & $0.728$ & $0.758$ & $1.125$ & $[1.115,\,1.135]$ \\
    ResNet-18 (layer) & $22$ & $0.934$ & $0.924$ & $0.876$ & $[0.788,\,0.964]$ \\
    Plain-18 (six groups) & $6$ & $0.647$ & $0.695$ & $1.158$ & $[1.075,\,1.241]$ \\
    Plain-18 (layer) & $19$ & $0.921$ & $0.913$ & $0.913$ & $[0.729,\,1.097]$ \\
    \bottomrule
  \end{tabular}
\end{table}

\paragraph{Granularity on the partially pretrained ResNet-50}
\label{par:granularity}
For ResNet-50 before fine-tuning, the six-group, residual-block and
individual-layer measurements over three seeds are summarized in
Table~\ref{tab:granularity}. The body has ImageNet weights and the head is
newly random. Alongside group count, $\Phi$, $\Phi_{\mathrm{sep}}$ and $S$,
the final column gives off-block mass relative to its separable-profile value.

\begin{table}[t]
  \centering
  \caption{Curvature profiles across three partitions of ResNet-50 with an ImageNet-pretrained body and a new random 196-class head, before fine-tuning. Measurements use evaluation mode on one fixed 32-image Stanford Cars validation batch. Curvature quantities are computed within seed before averaging over three seeds (42--44), which vary the new head and probes. The off-block/sep column estimates off-block squared Hessian mass divided by its separable-profile value $\sum_{v\ne w}a_va_w$, where $a_v=\|\mathbf{H}_{\boldsymbol{\theta}_v,\boldsymbol{\theta}_v}\|_F$; it is a mean of per-seed ratios, not a ratio of means.}
  \label{tab:granularity}
  \begin{tabular}{@{}lccccc@{}}
    \toprule
    Partition & Groups & $\Phi$ & $\Phi_{\mathrm{sep}}$ & $S$ & off-block/sep \\
    \midrule
    Stage & \GranGroupsStage & \PhiGranStageVThree & \PhiSepGranStage & \SGranStage & \OffOverSepGranStage \\
    Block & \GranGroupsBlock & \PhiGranBlockVThree & \PhiSepGranBlock & \SGranBlock & \OffOverSepGranBlock \\
    Layer & \GranGroupsLayer & \PhiGranLayerVThree & \PhiSepGranLayer & \SGranLayer & \OffOverSepGranLayer \\
    \bottomrule
  \end{tabular}
\end{table}

Neither finer partition refines the six groups: parameters outside its leaf
modules are pooled into a single rest group. For each seed, exact rest-group
diagonal mass bounds the possible change from resolving this missing nesting.
Its mean estimate at the layer partition is about $\RestMassLayerPpm$ parts
per million of $\|\mathbf{H}_{\boldsymbol{\theta}}\|_F^2$, not a certified maximum across seeds.
Splitting the pooled group can only increase exact $\Phi$, so the layer target
still bounds a genuinely nested per-module partition below. At the block
partition, the pooled group combines stem and head; $\Phi$ is lower than
at six groups on $\PhiBlockBelowStageSeeds$ of the three seeds.

From stage to layer, $\Phi$ rises from $\PhiGranStageVThree$ to
$\PhiGranLayerVThree$. The six-group model retains $\KeptShareStage$ of
squared curvature mass, compared with $\KeptShareLayer$ for the layer
partition, which groups each convolutional or linear module separately and
pools the parameters untouched by the preconditioners. These shares average
the per-seed fractions $1-\Phi$. Approximate between/within accounting uses
$\overline{\Phi}_{\mathrm{stage}}/\overline{\Phi}_{\mathrm{layer}}$ and its
complement, giving $\DiscardedBetweenStages$ and
$\DiscardedWithinStages$ per cent. They are ratios of seed-mean discarded
fractions, not ratios of unnormalized masses or means of per-seed ratios.
Because the partitions are non-nested, the complement is net reclassification
rather than all newly exposed within-stage mass. For an exact operator, let
$M_{\mathrm{new}}$ denote newly off-block mass and $M_{\mathrm{merge}}$
denote old cross-stage mass merged into the pooled rest group. Then
$O_{\mathrm{layer}}-O_{\mathrm{stage}}=M_{\mathrm{new}}-M_{\mathrm{merge}}$,
where $O$ is off-block squared mass. Exact layer-rest diagonal mass bounds
$M_{\mathrm{merge}}$; its small mean estimate suggests a small correction,
without establishing equality. These static shares do
not measure coverage of the correction, which selects pairs between six
coarse groups during training.

\section{Extended correction outcomes}
\label{app:extended-correction}

Stage B compares the two corrected methods with their base preconditioners,
two first-order baselines and truncated Newton-CG on Stanford Cars
(Table~\ref{tab:stage-b}). Criterion~(b), evaluated on the K-FAC pair,
fails three of its four parts (Table~\ref{tab:gates}).

The raw proposal lowers the damped model~\eqref{eq:model} on
$\ModelLoweredRecords$ of $\RefreshRecords$ K-FAC refresh records across
three seeds. Proposition~\ref{prop:pair-descent} gives an ideal descent
bound only under its current-Hessian and orthonormality conditions; freshness
alone does not certify it for these proposals. Median model value changes from $\MedianModelBase$ at the
base step to $\MedianModelProposal$ at the proposal. These are comparisons
of $m(\mathbf{d}_1)$ with $m(\mathbf{d}_0)$ at the same weights, mode and minibatch.
Nevertheless, the loss test rejects $\LossRejectRecords$ proposals
($\LossRejectShare$ per cent), leaving $\AcceptedRecords$ accepted. On those
rejected records the applied displacement remains $\mathbf{d}_0$.
Across all $\ProposalsTotal$ proposals, one per step in the same runs,
$\RejectByLoss$ fail the loss test ($\RejectByLossShare$ per cent),
$\RejectByRatio$ the ratio test and $\RejectByModel$ the model test;
$\AcceptShare$ per cent are accepted.

\begin{table*}[t]
  \centering
  \caption{Stage B: per method, final validation accuracy
    (mean$\,{\pm}\,1\sigma$ over 3~seeds), epochs to the target
    accuracy of $90\,\%$ (mean over the seeds that reached it;
    reached/total), gap closed $\gamma$ (applied step; median over eligible refresh records,
    95\,\% stratified record-bootstrap CI; eligible/total: COUPLE-FAC (K-FAC) $450/450$; COUPLE-FAC (EKFAC) $449/450$), safeguard accept rate (mean over epochs and
    seeds), and Spearman $\rho_s$ between $\mathcal{C}(v,w)$ and
    $\varepsilon(v,w)$ over all finite ordered pairs in the refresh records,
    without the gamma eligibility filter. Eligibility requires
    $m(\mathbf{d}_0)-m(\mathbf{d}_N)>10^{-8}\max(1,|m(\mathbf{d}_N)|)$; the truncated oracle is not an optimum.
    The historical $p$-values are nominal and the record bootstrap does not account
    fully for serial dependence or new trajectories. Accuracy means describe separate
    trajectories, not paired causal effects. Proposal model and native-loss comparisons
    within each refresh are same-state comparisons. $m(\mathbf{d}_0)>0$ is the share of
    refresh records, pooled over seeds, on which the damped model predicts an
    increase along the base step $\mathbf{d}_0$. Verdicts are the safeguard's decisions pooled over seeds:
    the share of proposals accepted, then the numbers rejected by the ratio, the
    model and the loss test, in that order.}
  \label{tab:stage-b}
  \scriptsize
  \setlength{\tabcolsep}{3pt}
  \begin{tabular}{@{}lccccccc@{}}
    \toprule
    Method & Final acc.\,(\%) & Ep.\ to target & $\gamma$ [95\,\% CI]
    & $m(\mathbf{d}_0)>0$ & Accept & \begin{tabular}[c]{@{}c@{}}Verdicts\\acc.; rej.\ by\\ratio/model/loss\end{tabular} & $\rho_s$ \\
    \midrule
    SGD & $75.9 \pm 0.4$ & --- (0/3) & --- & --- & --- & --- & --- \\
    Adam & $86.4 \pm 0.6$ & --- (0/3) & --- & --- & --- & --- & --- \\
    K-FAC & $84.0 \pm 0.8$ & --- (0/3) & --- & $86.4\,\%$ & --- & --- & --- \\
    COUPLE-FAC (K-FAC) & $83.2 \pm 0.9$ & --- (0/3) & $0.00\,[0.00,\,0.00]$ & $95.3\,\%$ & $0.15$ & $15\,\%$; $892/14/6766$ & $0.01$ ($p{=}0.105$) \\
    EKFAC & $80.5 \pm 1.3$ & --- (0/3) & --- & $98.0\,\%$ & --- & --- & --- \\
    COUPLE-FAC (EKFAC) & $78.9 \pm 0.7$ & --- (0/3) & $0.00\,[0.00,\,0.00]$ & $98.7\,\%$ & $0.32$ & $32\,\%$; $54/2/6056$ & $0.04$ ($p{<}10^{-3}$) \\
    Newton-CG & $0.8 \pm 0.2$ & --- (0/3) & --- & --- & --- & --- & --- \\
    \bottomrule
  \end{tabular}
\end{table*}

The applied closed gap has an atom at zero: $\GammaZeroShare$ per cent of
K-FAC values are exactly zero and $\GammaAboveOneShare$ per cent exceed one.
Consequently, both the median and its record-bootstrap interval in
Table~\ref{tab:stage-b} are zero. Values above one are possible because the
truncated reference is not an optimum. The reference is evaluated only at
refreshes, of which $\GammaEligibleRecordsKfac$ K-FAC records and
$\GammaEligibleRecordsEkfac$ EKFAC records meet the positive-gap rule in
Definition~\ref{def:gamma}; $\GammaSkippedRecordsKfac$ and
$\GammaSkippedRecordsEkfac$, respectively, are excluded. EKFAC seed 43 at
step zero has a negative reference gap rather than a rounding tie, but its
ordered coupling/action observations remain in the association analysis.
The within-seed record-bootstrap interval is conditional on the observed
trajectories and does not fully account for serial dependence.

At refreshes, both the gate and reduced subspace are rebuilt, yet acceptance
is lower than at other steps: $\AcceptShareRefresh$ against
$\AcceptShareOther$ per cent for K-FAC, and $\AcceptShareRefreshEkfac$
against $\AcceptShareOtherEkfac$ per cent for EKFAC. Of the
$\AcceptedRecords$ accepted K-FAC refresh records,
$\AcceptedEarlyRecords$ occur in the first three refreshes of the three
seeds. Acceptance is concentrated early.

No method reaches the registered ninety per cent validation accuracy target
within thirty epochs on any seed. Each corrected method has lower three-seed
mean final accuracy than its baseline on separate training trajectories
(Table~\ref{tab:stage-b}), so the correction did not demonstrate a training
gain in these runs. Baseline provenance is described in \ref{app:protocol}.
Table~\ref{tab:payoff-budget} reports the cost in Hessian-vector products
and wall-clock seconds per step; its caption specifies which products are
counted.

\begin{table}[t]
  \centering
  \caption{Cost per optimiser step by method: exact Hessian-vector products (HVPs) per
    step (the diagnostics included, the Hutchinson probes of the coupling field not)
    and training wall-clock seconds per step, means over seeds. HVP/solve and
    non-pos.\ curv.\ describe the truncated Newton-CG oracle's solve on each method's own
    trajectory (for the Newton-CG method, its own step): the number of
    conjugate-gradient iterations (one exact HVP each) per solve and the share of
    solves that stopped on non-positive curvature (including zero), a non-PD diagnostic.
    Counts retain solves on non-finite states,
    where a non-positive-curvature stop need not be flagged.}
  \label{tab:payoff-budget}
  \small
  \begin{tabular}{@{}lcccc@{}}
    \toprule
    Method & HVP/step & HVP/solve & Non-pos.\ curv. & s/step \\
    \midrule
    SGD & $0.0$ & --- & --- & $0.463$ \\
    Adam & $0.0$ & --- & --- & $0.250$ \\
    K-FAC & $1.8$ & $9.3$ & $67\,\%$ & $2.732$ \\
    COUPLE-FAC (K-FAC) & $14.3$ & $9.1$ & $74\,\%$ & $23.483$ \\
    EKFAC & $1.9$ & $10.7$ & $59\,\%$ & $3.648$ \\
    COUPLE-FAC (EKFAC) & $14.5$ & $10.9$ & $63\,\%$ & $24.086$ \\
    Newton-CG & $5.8$ & $5.7$ & $81\,\%$ & $7.722$ \\
    \bottomrule
  \end{tabular}
\end{table}

As an optimizer at the same $\lambda$, truncated Newton-CG finishes near
chance accuracy (Table~\ref{tab:stage-b}). It stops on non-positive search
direction curvature on $\NewtonNegCurvShare$ per cent of steps. One run
becomes non-finite from step $\NewtonDivergeStep$, where no such stop can be
flagged; the share among finite-state solves is at least
$\NewtonNegCurvShareFiniteLo$ per cent. Along the K-FAC trajectory,
$\OracleNegCurvShare$ per cent of reference solves stop on non-positive
curvature, including zero. This diagnoses failure of positive definiteness,
not necessarily strict negative curvature or indefiniteness.

The correction selects at most three pairs by normalized coupling, not by
off-block mass. Its residual combines cross-group Hessian action and
within-group operator/damping differences from the Fisher-based baseline
(equation~\eqref{eq:residual}). Neither historical mass coverage nor the
additional utility of the selected $\mathbf{U}$ was measured. Small raw ordered
associations between coupling and the action ratio coexist with positive
retrospective symmetric and adjusted per-seed associations (\ref{app:coupling-sensitivity},
Table~\ref{tab:coupling-sensitivity}, Figure~\ref{fig:eps-vs-C}).
The frozen-state controls in \ref{app:causal-pilot} compare endpoint refresh,
scale and surrogate objectives on separate plain K-FAC trajectories.
Gradient refresh improves on the best sampled scaling in some states,
whereas curvature-only and joint refresh do not. The summary in
Table~\ref{tab:causal-pilot} is selective: \ref{app:causal-pilot}
identifies the archived scalar records for all refresh variants, the scale
control and the pooling intervention.
These controls do not isolate a causal off-block contribution or establish
a training gain.

\clearpage

\section{The validity measurement}
\label{app:validity}

The validity study tests signed loss/model agreement along fractions of base
updates on trajectories separate from the correction study.

\subsection{Validity of the damped model along a step}
\label{sec:validity-def}

For the validity diagnostic, $\mathbf{d}_0$ is the full applied displacement,
including Adam's learning rate where applicable. We write
$\Delta L(t)=L(\boldsymbol{\theta}+t\mathbf{d}_0)-L(\boldsymbol{\theta})$ and compare it with the model value $m(t\mathbf{d}_0)$ of \eqref{eq:model} on a grid $G$ of step fractions
$t$ between $10^{-4}$ and $1$.

\begin{definition}[Sampled validity extent]
  \label{def:tstar}
  With a relative tolerance $\mathrm{tol}$,
  \begin{equation}\label{eq:tstar}
    t^*:=\max\bigl\{t\in G:\ |\Delta L(u)-m(u\mathbf{d}_0)|\le\mathrm{tol}\,|\Delta L(u)|
    \ \text{for every}\ u\in G\ \text{with}\ u\le t\bigr\} .
  \end{equation}
\end{definition}

The inequality remains meaningful at $\Delta L(u)=0$, where it requires an
exactly zero model value.

This sampled prefix requires
agreement at every grid point up to and including $t^*$. It is not a continuous
segment or ball radius. Failure at the smallest grid point leaves an empty
prefix and undefined $t^*$ (reported as undetermined), even if later points
pass. The protocol below specifies the tolerance and grid. Each record also
contains four quantities: the pointwise curvature along the step,
$\mathbf{d}_0^\top \mathbf{H}\mathbf{d}_0/\|\mathbf{d}_0\|^2$; the secant curvature the loss realizes over the whole
step, $\kappa_{\mathrm{sec}}=2\,(\Delta L(1)-\mathbf{g}^\top \mathbf{d}_0)/\|\mathbf{d}_0\|^2$; the extreme
eigenvalues $\lambda_{\max}$ and $\lambda_{\min}$ by Lanczos; and the fraction
of rectified-linear pre-activations whose sign changes between $\boldsymbol{\theta}$ and
$\boldsymbol{\theta}+\mathbf{d}_0$. Curvature quotients require $\|\mathbf{d}_0\|>0$; Lanczos values are
finite-iteration spectral estimates, not exact eigendecompositions.

\subsection{Observed validity and sensitivity}
\label{sec:results-validity}

In the Stage B records, the damped model~\eqref{eq:model} predicts an
increase along the base step while minibatch loss decreases on
$\ValidityShareKfac$ per cent of K-FAC records and
$\ValidityShareKfacOverlay$, $\ValidityShareEkfac$ and
$\ValidityShareEkfacOverlay$ per cent for COUPLE-FAC (K-FAC), EKFAC and
COUPLE-FAC (EKFAC), respectively.
The base loss decreases on all 450 refresh records of each arm, so the joint
counts coincide with the model-positive counts.
The separate validity trajectories examine agreement over fractions of a
step; they do not test the mechanism of correction rejection.
Table~\ref{tab:validity} gives $t^*$ for the original K-FAC cohort:
agreement covers the whole sampled step at initialization and shorter
prefixes at later probes. Table~\ref{tab:validity-arms} compares the other
two trajectories.

At probes 200, 600 and 1500, $t^*$ is at most one tenth of the displacement
in $\TstarCellsOk$ of $\TstarCellsTotal$ original arm/seed cells at relative
tolerance $\TstarTol$. Probe 200 is the earliest sampled step from which all
later sampled cells meet this bound; intervening steps are unobserved.
At probes 20, 60, 200, 600 and 1500, $\TstarCellsOkCoarse$ of
$\TstarCellsTotalCoarse$ cells have extent at most a quarter. Each extent
is a grid point: $t^*=0.1$ means that the next point, $0.25$, fails, rather
than locating a continuous threshold between them.

Tolerance sensitivity distinguishes the agreement rule from the reported
extent threshold. At relative-error tolerance $0.05$,
$\TstarCellsOkDefinedTolFive$ of $\TstarCellsDefinedTolFive$ cells with a
defined extent meet the one-tenth bound; $\TstarUndeterminedTolFive$ have an
empty prefix. At tolerances $0.2$ and $0.5$, the counts are
$\TstarCellsOkTolTwenty$ and $\TstarCellsOkTolFifty$ of
$\TstarCellsTotal$, respectively, with median $t^*$ moving from
$\TstarMedianTolFive$ to $\TstarMedianTolFifty$ across the range. Thus the
one-tenth bound holds for defined cells at tolerances between one twentieth
and one tenth, but becomes less frequent from one fifth onwards. The extent
threshold is a separate reporting choice: among the same
$\TstarCellsTotal$ cells, $\TstarCellsOkTwentieth$ are at or below a
twentieth and $\TstarCellsOkHundredth$ at or below a hundredth of the step.

The trajectories differ in their temporal pattern. Between probes 20 and 60
and probes 600 and 1500, the K-FAC median falls from
$\TstarMedianEarlyKfac$ to $\TstarMedianLateKfac$, a factor of
$\TstarDropKfac$. The gradient-descent median falls by a factor of
$\TstarDropSgd$, whereas Adam's remains $\TstarMedianAdam$ in both periods.

\begin{table}[htbp]
  \centering
  \caption{Validity of the damped quadratic model
    $m(t \mathbf{d}_0)=t\,\mathbf{g}^{\top}\mathbf{d}_0+\tfrac{t^2}{2}\,\mathbf{d}_0^{\top}(\mathbf{H}+\lambda \mathbf{I})\mathbf{d}_0$ along the
    K-FAC step $\mathbf{d}_0$ (ResNet-50 on Stanford Cars, batch~64,
    $\lambda=1$, $\eta=1$, MC~Fisher;
    3~seeds; mean $\pm\,1\sigma$ over seeds, $t^*$ listed in ascending order), on the step's own minibatch before the step is applied.
    Curvatures per unit $\|\mathbf{d}_0\|^2$: the pointwise $\mathbf{d}_0^{\top}\mathbf{H}\mathbf{d}_0/\|\mathbf{d}_0\|^2$ and the
    secant $\kappa_{\mathrm{sec}}=2\,(\Delta L(1)-\mathbf{g}^{\top}\mathbf{d}_0)/\|\mathbf{d}_0\|^2$ realised by the
    loss over the whole step; $\lambda_{\max}$ and $\lambda_{\min}$ from the same
    Lanczos run. Sampled validity extent $t^*$: the largest fraction of the
    step up to which $|\Delta L-m|\le0.1\,|\Delta L|$ holds at every
    sampled grid point in the initial prefix, separately per seed; an empty prefix
    reads ${<}\,t_{\min}$.
    This does not assert failure at every later grid point. Kinks: the fraction of ReLU units whose pre-activation changes sign between
    $\boldsymbol{\theta}$ and $\boldsymbol{\theta}+\mathbf{d}_0$.}
  \label{tab:validity}
  \footnotesize
  \setlength{\tabcolsep}{4pt}
  \begin{tabular}{@{}rcccccccc@{}}
    \toprule
    Step & $\|\mathbf{d}_0\|$ & $\mathbf{g}^{\top}\mathbf{d}_0$ & $\mathbf{d}_0^{\top}\mathbf{H}\mathbf{d}_0/\|\mathbf{d}_0\|^2$ & $\kappa_{\mathrm{sec}}$ & $\lambda_{\max}$ & $\lambda_{\min}$ & $t^*$ & kinks \\
    \midrule
    0 & $0.773 \pm 0.008$ & $-1.130 \pm 0.020$ & $0.1 \pm 0.1$ & $0.9 \pm 0.1$ & $26.5 \pm 13.6$ & $-26.0 \pm 21.8$ & $1/1/1$ & $1.4 \pm 0.2\,\%$ \\
    20 & $1.659 \pm 0.033$ & $-5.223 \pm 0.166$ & $1.9 \pm 0.2$ & $2.5 \pm 0.1$ & $40.8 \pm 14.8$ & $-15.4 \pm 3.4$ & $0.1/0.1/0.1$ & $2.2 \pm 0.1\,\%$ \\
    60 & $2.473 \pm 0.154$ & $-11.638 \pm 1.432$ & $6.7 \pm 0.7$ & $3.1 \pm 0.0$ & $57.1 \pm 12.9$ & $-19.5 \pm 4.4$ & $0.25/0.25/0.25$ & $2.5 \pm 0.2\,\%$ \\
    200 & $1.268 \pm 0.122$ & $-3.081 \pm 0.599$ & $15.5 \pm 2.9$ & $3.4 \pm 0.1$ & $57.3 \pm 8.8$ & $-7.0 \pm 1.8$ & $0.05/0.1/0.1$ & $1.0 \pm 0.1\,\%$ \\
    600 & $0.942 \pm 0.134$ & $-1.734 \pm 0.472$ & $17.7 \pm 7.8$ & $3.5 \pm 0.1$ & $65.7 \pm 19.0$ & $-7.1 \pm 1.4$ & $0.01/0.01/0.05$ & $0.8 \pm 0.2\,\%$ \\
    1500 & $0.184 \pm 0.142$ & $-0.101 \pm 0.124$ & $23.4 \pm 19.7$ & $3.1 \pm 0.4$ & $47.4 \pm 39.9$ & $-1.5 \pm 1.3$ & $0.01/0.05/0.1$ & $0.1 \pm 0.1\,\%$ \\
    \bottomrule
  \end{tabular}
\end{table}

\begin{table}[htbp]
  \centering
  \caption{The sampled validity extent $t^*$ of the damped quadratic model of
    Table~\ref{tab:validity} ($|\Delta L-m|\le0.1\,|\Delta L|$ at every
    sampled grid point in the initial prefix, separately per seed)
    for three training arms (ResNet-50 on Stanford Cars, batch~64;
    $\lambda=1$ in the model; outer displacement multiplier $\eta=1$ in every arm;
    K-FAC/GD learning rate one, Adam learning rate $10^{-3}$ already included in $\mathbf{d}_0$),
    each run separately on the same 3~seeds. $t^*$ is listed in ascending
    order.}
  \label{tab:validity-arms}
  \footnotesize
  \begin{tabular}{@{}rccc@{}}
    \toprule
    Step & K-FAC & GD & Adam \\
    \midrule
    0 & $1/1/1$ & $0.01/0.25/0.5$ & $0.1/0.1/0.1$ \\
    20 & $0.1/0.1/0.1$ & $0.25/0.25/0.5$ & $0.05/0.05/0.05$ \\
    60 & $0.25/0.25/0.25$ & $0.05/0.05/0.25$ & $0.1/0.1/0.1$ \\
    200 & $0.05/0.1/0.1$ & $0.1/0.1/0.1$ & $0.05/0.05/0.05$ \\
    600 & $0.01/0.01/0.05$ & $0.01/0.01/0.1$ & $0.05/0.1/0.1$ \\
    1500 & $0.01/0.05/0.1$ & $0.01/0.01/0.05$ & $0.05/0.05/0.1$ \\
    \bottomrule
  \end{tabular}
\end{table}

The later seed-42 K-FAC repeat, excluded from all cell counts, differs from
the original. At step 200 its extent is $\TstarReplicateStepTwoHundred$
rather than $\TstarOriginalStepTwoHundred$, exceeding a tenth; at step 600
it is $\TstarReplicateStepSixHundred$ rather than
$\TstarOriginalStepSixHundred$. At step $\TstarReplicateUndeterminedStep$
the first sampled point fails, leaving an undefined prefix rather than a
finite small extent. Later points can still pass.
Table~\ref{tab:validity-resolution} separates this
repeat from the nine original late K-FAC cells and reports first-failure
margins and absolute loss-change scales. An empty prefix alone does not
identify a rounding artefact.

Pointwise and realized curvature also differ. Along K-FAC, pointwise
curvature grows during training, while whole-step secant curvature passes
two by step 20 and lies between three and three and a half at probes 60,
200, 600 and 1500 (Table~\ref{tab:validity} and Figure~\ref{fig:validity}b).
At the final probe, the ratio of seed means is $\CurvRatioLast$, but per-seed
ratios range from $\CurvRatioLastLo$ to $\CurvRatioLastHi$.

The direction of this discrepancy varies across the measured optimizer trajectories. Across all
probes and seeds, secant curvature is below pointwise curvature in
$\SecantBelowKfac$ of $\SecantCellsArm$ K-FAC cells,
$\SecantBelowSgd$ gradient-descent cells and $\SecantBelowAdam$ Adam cells.
At probes 60, 200, 600 and 1500, median pointwise-to-secant ratios are
$\CurvRatioMedianKfac$ for K-FAC and $\CurvRatioMedianSgd$ for gradient
descent, spanning $\CurvRatioSpanLo$ to $\CurvRatioSpanHi$ across cells.
For Adam, the median is $\CurvRatioMedianAdam$, with range
$\CurvRatioAdamLo$ to $\CurvRatioAdamHi$ crossing one; at the median,
realized curvature exceeds pointwise curvature.

Removing the damping term $\tfrac12t^2\lambda\|\mathbf{d}_0\|^2$ from the stored
model values has different effects across arms. At probes 200, 600 and
1500, the undamped extent remains at most a tenth in
$\TstarUndampedOkKfac$ of $\TstarCellsPerArm$ K-FAC cells and
$\TstarUndampedOkSgd$ of $\TstarCellsPerArm$ gradient-descent cells.
For Adam it reaches half a step or further in $\TstarUndampedHalfAdam$ of
$\TstarCellsPerArm$ cells. There the damping term contributes
$\DampSharePctAdamLo$ to $\DampSharePctAdamHi$ per cent of the quadratic
coefficient $\mathbf{d}_0^\top \mathbf{H}\mathbf{d}_0+\lambda\|\mathbf{d}_0\|^2$.
Short damped agreement is therefore common to the original late arms, but
short undamped agreement is not. On the original K-FAC probes it persists
after damping is removed.

\FigureValidity

Lanczos estimates place $\lambda_{\min}$ below $-\lambda$ in
$\LambdaMinCellsBelowDamping$ of $\LambdaMinCellsTotal$ arm/seed/probe
cells, indicating directions along which the damped model is unbounded below.
The exceptions are the final probe of two K-FAC seeds. Along K-FAC, the
fraction of rectified-linear pre-activations changing sign between $\boldsymbol{\theta}$
and $\boldsymbol{\theta}+\mathbf{d}_0$ falls from $\KinkShareInit$ per cent at initialization to
$\KinkShareLate$ per cent at the final probe.

\subsection{Protocol and numerical interpretation}

\paragraph{Training arms and measurements}
The three arms follow separate ResNet-50 training trajectories on Stanford
Cars, rather than evaluating three directions on one trajectory. Each uses
batch 64, model damping $\lambda=1$ and unit outer displacement multiplier
$\eta=1$. K-FAC uses learning rate one and the Fisher factors specified in
Table~\ref{tab:stage-b-config}. Gradient descent uses $\mathbf{d}_0=-\mathbf{g}$ without momentum,
and Adam uses learning rate $10^{-3}$, included in $\mathbf{d}_0$. Fractions
$t$ scale the complete displacement. The latter two arms have no weight decay, clipping or schedule and are
distinct from the Stage B SGD and Adam baselines. Each original arm uses seeds
42--44, subject to the seed-44 worker-state limitation described in
\ref{app:protocol}. The seed-42 K-FAC repeat used the code then available.
It is excluded from all cell counts because it is a second realization of
an existing seed, not a fourth independent seed.

Measurements are made at steps $\{0,20,60,200,600,1500\}$ on the current
minibatch before the update. Trial evaluations copy BatchNorm buffers and
leave training weights unchanged; the ordinary $\mathbf{d}_0$ update follows. Each
probe records:
\begin{enumerate}[label=(\roman*),leftmargin=1.6em,itemsep=2pt,topsep=2pt]
  \item The damped-model components $\mathbf{g}^\top \mathbf{d}_0$, $\mathbf{d}_0^\top \mathbf{H}\mathbf{d}_0$ and
    $\lambda\|\mathbf{d}_0\|^2$, together with $\|\mathbf{d}_0\|$ and $\|\mathbf{g}\|$.
  \item The loss change $\Delta L(t)=L(\boldsymbol{\theta}+t\mathbf{d}_0)-L(\boldsymbol{\theta})$ and model value
    $m(t\mathbf{d}_0)$ on the grid
    \[
      t\in\{10^{-4},10^{-3},10^{-2},0.05,0.1,0.25,0.5,1\},
    \]
    with one forward pass per point; the secant curvature
    $\kappa_{\mathrm{sec}}(t)=2\,(\Delta L(t)-t\,\mathbf{g}^\top \mathbf{d}_0)/(t^2\|\mathbf{d}_0\|^2)$;
    and $t^*$ at relative tolerance $\mathrm{tol}=\TstarTol$.
  \item Extreme eigenvalues $\lambda_{\max}$ and $\lambda_{\min}$ from one
    sixty-step Lanczos run using selected-AD products, and Rayleigh quotients
    along $\mathbf{d}_0$, $\mathbf{g}$ and a reconstructed top direction. These are
    finite-iteration estimates; the historical refinement budget was not recorded.
  \item Endpoint branch changes: the fractions of rectified-linear
    pre-activations changing sign and max-pool windows changing winner between
    $\boldsymbol{\theta}$ and $\boldsymbol{\theta}+\mathbf{d}_0$. These compare endpoints rather than count all
    crossings along the ray. Table~\ref{tab:validity} reports the first fraction.
\end{enumerate}

\paragraph{Sampled-prefix interpretation}
The extent is calculated from signed loss and model values within each seed
and probe record. Figure~\ref{fig:validity} instead displays absolute values
of seed means, whose visual agreement cannot establish the per-seed test.
Definition~\ref{def:tstar} uses the initial prefix of sampled fractions. At zero loss change, the stored
relative error is zero if the model also vanishes and infinity otherwise.

Each measurement concerns one minibatch rather than the full training or
validation objective.

The undamped extent in \ref{sec:results-validity} is calculated from
the stored values as $m(t\mathbf{d}_0)-\tfrac12t^2\lambda\|\mathbf{d}_0\|^2$, using the same
signed relative-error test, prefix rule and tolerance. The calculation uses
$\Delta L(t)$ and $m(t\mathbf{d}_0)$ at every grid point and the single recorded value
of $\lambda\|\mathbf{d}_0\|^2$; it requires no new measurements.

\paragraph{Numerical resolution}
Table~\ref{tab:validity-resolution} separates the nine original late K-FAC
cells (seeds 42--44 at probes 200, 600 and 1500) from the later seed-42 repeat.
For either displayed model $\mu(t)$, the signed margin at the first failed
point is
\begin{equation}\label{eq:validity-margin}
  M(t)=|\Delta L(t)-\mu(t)|-0.1|\Delta L(t)|,
  \qquad \mu(t)\in\{m(t\mathbf{d}_0),q(t\mathbf{d}_0)\};
\end{equation}
a positive margin indicates failure. The original cells first fail at
fractions $0.05$, $0.1$ or $0.25$, at the native-loss scales shown in the table.
The repeat's empty prefix at step 1500 instead fails at $10^{-4}$ with
$|\Delta L|\simeq8.45\times10^{-8}$. Its margin is about 152 illustrative
binary32 loss-spacing units, rather than less than one.

The undamped values are obtained by subtraction from recorded damped values,
not by repeating the calculations at a known precision. Historical tensor
and reduction precision is unrecorded for these cohorts. Returned-loss
spacing provides a scale comparison but does not bound errors in the forward
pass, AD/HVP, parameter addition or loss reduction. It therefore cannot
certify a prefix or identify the repeat, or the original nine cases, as
rounding artefacts.

\begin{table}[htbp]
  \centering
  \caption{Retrospective validity-resolution diagnostic on K-FAC
    (ResNet-50, Stanford Cars, batch 64,
    $\lambda=1$), using each record's own pre-update training minibatch
    (sample identities not retained). Original: exactly nine cells, seeds
    42--44 crossed with steps 200, 600, 1500 from the original input bundle;
    repeat: the separate seed-42 K-FAC repeat, never an extra independent seed.
    Original rows give componentwise min--max ranges over those nine cells
    (scientific endpoints stacked: minimum above maximum), not paired extrema;
    repeat rows show each checkpoint separately.
    Grid: $\{0.0001, 0.001, 0.01, 0.05, 0.1, 0.25, 0.5, 1\}$. Sampled validity extent is the last successful point before the
    first failure $t_f$ of $|\Delta L-\mu|\le0.1|\Delta L|$.
    At $t_f$ the table shows native $\Delta L=L(\boldsymbol{\theta}+t_f\mathbf{d}_0)-L(\boldsymbol{\theta})$, the
    model value and signed margin $M=|\Delta L-\mu|-0.1|\Delta L|$ (positive fails).
    Damped uses the logged model; undamped subtracts
    $t^2\lambda\|\mathbf{d}_0\|^2/2$ from it, not from native loss.
    $\varnothing$ means an undefined prefix (the first point failed); ``none''
    in the failure columns means no sampled failure, not an undefined prefix.
    Zero/zero succeeds; nonzero model with zero loss change fails.
    $s_{32}=|\operatorname{spacing}(\operatorname{binary32}(L(\boldsymbol{\theta})))|$ is
    an illustrative subtraction-resolution scale at the rounded logged loss, not an
    error bound; a missing scale or overflowing ratio is shown as ``none''.
    Failure ends the prefix even if later points pass.}
  \label{tab:validity-resolution}
  \scriptsize
  \setlength{\tabcolsep}{3pt}
  \begin{tabular}{@{}lllcccccc@{}}
    \toprule
    Cohort / seed & Step & Model & Prefix & $t_f$ & $\Delta L(t_f)$ & $\mu(t_f)$ & $M(t_f)$ & $|M|/s_{32}$ \\
    \midrule
    Original 42--44 & 200/600/1500 & damped & $[0.01,\,0.1]$ & $[0.05,\,0.25]$ & \begin{tabular}[c]{@{}c@{}}$-2.75{\cdot}10^{-1}$\\$-9.78{\cdot}10^{-4}$\end{tabular} & \begin{tabular}[c]{@{}c@{}}$-1.77{\cdot}10^{-1}$\\$-8.49{\cdot}10^{-4}$\end{tabular} & \begin{tabular}[c]{@{}c@{}}$3.18{\cdot}10^{-5}$\\$2.19{\cdot}10^{-1}$\end{tabular} & \begin{tabular}[c]{@{}c@{}}$1.04{\cdot}10^{5}$\\$7.35{\cdot}10^{6}$\end{tabular} \\
    Original 42--44 & 200/600/1500 & undamped & $[0.01,\,0.1]$ & $[0.05,\,0.25]$ & \begin{tabular}[c]{@{}c@{}}$-2.75{\cdot}10^{-1}$\\$-1.65{\cdot}10^{-3}$\end{tabular} & \begin{tabular}[c]{@{}c@{}}$-8.71{\cdot}10^{-2}$\\$2.83{\cdot}10^{-1}$\end{tabular} & \begin{tabular}[c]{@{}c@{}}$8.73{\cdot}10^{-5}$\\$5.22{\cdot}10^{-1}$\end{tabular} & \begin{tabular}[c]{@{}c@{}}$1.87{\cdot}10^{5}$\\$1.75{\cdot}10^{7}$\end{tabular} \\
    \midrule
    Repeat 42 & 200 & damped & $0.25$ & $0.5$ & $-5.07{\cdot}10^{-1}$ & $4.24{\cdot}10^{-1}$ & $8.80{\cdot}10^{-1}$ & $1.48{\cdot}10^{7}$ \\
    Repeat 42 & 200 & undamped & $0.05$ & $0.1$ & $-3.25{\cdot}10^{-1}$ & $-3.68{\cdot}10^{-1}$ & $1.02{\cdot}10^{-2}$ & $1.72{\cdot}10^{5}$ \\
    Repeat 42 & 600 & damped & $0.05$ & $0.1$ & $-7.43{\cdot}10^{-3}$ & $-4.97{\cdot}10^{-3}$ & $1.72{\cdot}10^{-3}$ & $9.21{\cdot}10^{5}$ \\
    Repeat 42 & 600 & undamped & $0.05$ & $0.1$ & $-7.43{\cdot}10^{-3}$ & $-5.31{\cdot}10^{-3}$ & $1.37{\cdot}10^{-3}$ & $7.37{\cdot}10^{5}$ \\
    Repeat 42 & 1500 & damped & $\varnothing$ & $0.0001$ & $-8.45{\cdot}10^{-8}$ & $-1.11{\cdot}10^{-7}$ & $1.77{\cdot}10^{-8}$ & $1.52{\cdot}10^{2}$ \\
    Repeat 42 & 1500 & undamped & $\varnothing$ & $0.0001$ & $-8.45{\cdot}10^{-8}$ & $-1.11{\cdot}10^{-7}$ & $1.77{\cdot}10^{-8}$ & $1.52{\cdot}10^{2}$ \\
    \bottomrule
  \end{tabular}
\end{table}

\section{Further context for supporting measurements}
\label{app:related-context}

The structural measurements in \ref{app:extended-mass} compare
$\Phi$ with size and diagonal-profile references for the selected parameter
Hessian. Dong et al.~\citep{dong2025quantifying} give class-count dependence
$O(1/C)$ for hidden and $O(1/C^{2})$ for output blocks at random initialization
in linear and one-hidden-layer models, using per-neuron/per-class rather than
our group partitions. Their scaling therefore does not supply a reference
value for our measured groups. The normalized GGN quantity
$\rho_{\mathrm{cross}}$ of Hong et al.~\citep{hong2025better} has the form of
$\Phi$ after squaring, but their operator and partition differ; they also
study exact Hessians and the $\mathbf{H}-\mathrm{GGN}$ residual. Operator choice
also limits comparisons with preconditioning results: relations between
Fisher and GGN curvature \citep{martens2020new} and the power-iteration
interpretation of the square of Shampoo's matrix approximation
\citep{morwani2025shampoo} do not equate those approximations with our Hessian
projection. On transformers up to 150 million parameters, Abreu et
al.~\citep{abreu2025potential} find that precise layerwise GGN preconditioning
nearly matches the full GGN variant; this concerns preconditioning outcomes,
not the parameter-Hessian projection error measured here.

Within the structural measurements, the recurring value/output role of the
strongest measured transformer pairs provides a specific comparison with the
heterogeneous self-attention Hessian-block dependencies derived by Ormaniec
et al.~\citep{ormaniec2025transformer}. Rank, power-law spectra, outliers and negative tails
\citep{singh2021analytic,tang2025hessian,papyan2019measurements,
sagun2017eigenvalues} describe aspects of the operator distinct from
$\Phi$ and $S$. \ref{app:conditioning} separately estimates spectral
extrema for one network; neither those estimates nor the Frobenius ratios
measure the preconditioned Hessian studied for adaptive methods at the edge
of stability \citep{cohen2022adaptive}.

The sampled validity extent in \ref{app:validity} concerns a signed
one-ray full-Hessian minibatch diagnostic, rather than a frozen multi-step
training continuation. Gradient-descent sharpness and full-Hessian Taylor
continuations are studied by Cohen et al.~\citep{cohen2021gradient}.
Meterez et al.~\citep{meterez2026defense} use linearized cross-entropy and
convex GGN proxies for LLM pretraining, preserving optimizer state and
schedule and finding middle-to-late agreement over roughly 5--10 per cent
of training. Their horizon and evaluation objective differ from ours.

\section{Conditioning of the blocks}
\label{app:conditioning}

This appendix provides a separate single-seed ResNet-18/CIFAR-10 illustration
of block spectra and the negative-direction implication in
\ref{app:extended-methods}.

A full-Hessian Newton step solves with $\mathbf{H}+\lambda \mathbf{I}$, whereas a Hessian-block
Newton step would use $\mathbf{H}_{\boldsymbol{\theta}_v,\boldsymbol{\theta}_v}+\lambda \mathbf{I}$. These differ from the
cached Fisher-based baseline $\mathbf{B}$ in~\eqref{eq:residual} and from the
reduced correction system, which has matrix $|\hat{\mathbf{H}}|+\lambda \mathbf{I}$ and right-hand side $-\mathbf{U}^\top \mathbf{r}$.
For a symmetric matrix $\mathbf{A}$, Table~\ref{tab:conditioning-resnet} reports
\begin{equation}\label{eq:cond}
  \lambda_{\max}(\mathbf{A}),\qquad
  \lambda_{\min}(\mathbf{A}),\qquad
  \kappa_\lambda(\mathbf{A}):=
  \begin{cases}
    \dfrac{\lambda_{\max}(\mathbf{A})+\lambda}{\lambda_{\min}(\mathbf{A})+\lambda},
      &\lambda_{\min}(\mathbf{A})+\lambda>0,\\
    \infty,&\text{otherwise}.
  \end{cases}
\end{equation}
The finite ratio is the spectral condition number of the positive-definite
damped system. Infinity denotes failure of positive definiteness, rather
than the ordinary condition number of an invertible indefinite matrix.

Training uses seed 42 and the Stage A schedule: stochastic
gradient descent at $0.1$, momentum $0.9$, cosine annealing over 100 epochs,
and batch 128. At initialization and the final checkpoint, the network is
measured in evaluation mode on the same 32 validation examples. The full
Hessian and its six diagonal parameter blocks are accessed through
selected-AD Hessian-vector products, with both extreme eigenvalues estimated
from one sixty-step Lanczos run per matrix.

\begin{table}[htbp]
  \centering
  \caption{Conditioning~\eqref{eq:cond} of the parameter-space blocks
    $\mathbf{H}_{\boldsymbol{\theta}_v,\boldsymbol{\theta}_v}$ and the full $\mathbf{H}_{\boldsymbol{\theta}}$ of ResNet-18 (seed~42, at
    initialisation and after training). $\infty$ means the damped matrix
    $\mathbf{A}+\lambda \mathbf{I}$ is not positive definite
    ($\lambda_{\min}+\lambda\le0$): a non-PD diagnostic convention,
    not the general condition number of an invertible indefinite matrix.}
  \label{tab:conditioning-resnet}
  \footnotesize
  \begin{tabular}{@{}llcccc@{}}
    \toprule
    Model & Block & $\lambda_{\max}$ & $\lambda_{\min}$ & $\kappa_{10^{-2}}$ & $\kappa_{10^{-1}}$ \\
    \midrule
    ResNet-18 (init) & full & $24.827$ & $-15.099$ & $\infty$ & $\infty$ \\
     & stem & $1.422$ & $-0.196$ & $\infty$ & $\infty$ \\
     & layer1 & $2.002$ & $-1.650$ & $\infty$ & $\infty$ \\
     & layer2 & $3.077$ & $-2.213$ & $\infty$ & $\infty$ \\
     & layer3 & $6.966$ & $-2.211$ & $\infty$ & $\infty$ \\
     & layer4 & $3.156$ & $-1.125$ & $\infty$ & $\infty$ \\
     & head & $8.264$ & $3.10{\cdot}10^{-9}$ & $827.4$ & $83.6$ \\
    ResNet-18 (final) & full & $5.79{\cdot}10^{3}$ & $-117.831$ & $\infty$ & $\infty$ \\
     & stem & $540.209$ & $-8.612$ & $\infty$ & $\infty$ \\
     & layer1 & $2.02{\cdot}10^{3}$ & $-47.856$ & $\infty$ & $\infty$ \\
     & layer2 & $2.86{\cdot}10^{3}$ & $-47.384$ & $\infty$ & $\infty$ \\
     & layer3 & $349.440$ & $-17.467$ & $\infty$ & $\infty$ \\
     & layer4 & $34.505$ & $-6.147$ & $\infty$ & $\infty$ \\
     & head & $0.496$ & $2.00{\cdot}10^{-7}$ & $50.6$ & $6.0$ \\
    \bottomrule
  \end{tabular}
\end{table}

The estimated full Hessian is indefinite at both checkpoints. At
initialization its smallest eigenvalue is of the same order in magnitude as
the largest, as are those of the first two stage blocks. After training the
full Hessian's negative extreme is smaller relative to the positive extreme,
but remains far from zero. Every trained block except the head has a negative
smallest eigenvalue; the first two stages have the largest negative
magnitudes, well beyond the damping used in Stage B. The head has no resolved
negative eigenvalue at the estimator's precision, which does not certify
positive semidefiniteness. Accordingly, $\kappa_\lambda$ is infinite for every
row except the head at both reported damping values.

For the trained first two stages, positive-definite damping would require
values above the entire Stage B calibration range (\ref{app:protocol}).
At the selected $\lambda=1$, all blocks except the head have estimated
smallest eigenvalues below $-\lambda$. A full-Hessian or Hessian-block Newton
system on this ResNet-18 would therefore remain indefinite at that damping.
The non-positive search-direction curvature recorded in the separate
ResNet-50/Cars experiment (Section~\ref{sec:results-b}) is a different
observation: that stop includes zero and establishes neither strict negative
curvature nor any individual block's sign.

The correction instead uses a positive-definite reduced metric with condition
number at most $(\|\hat{\mathbf{H}}\|_2+\lambda)/\lambda$. This bound depends on the
reduced Hessian and damping, with no explicit factor of the parameter count
$P$.

For the separate ResNet-50/Cars solves, the per-solve column of
Table~\ref{tab:payoff-budget} counts conjugate-gradient iterations on the full
Hessian system; it does not measure
$\kappa_\lambda(\mathbf{H}_{\boldsymbol{\theta}})$. On the K-FAC trajectory, the most frequent stop is
non-positive curvature ($\OracleNegCurvShare$ per cent of solves). For an
exact symmetric product, such a search-direction quotient implies
$\lambda_{\min}(\mathbf{H}_{\boldsymbol{\theta}})+\lambda\le0$, where the convention
in~\eqref{eq:cond} assigns infinite $\kappa_\lambda$.

\section{Exploratory frozen-state comparisons}
\label{app:causal-pilot}

The exploratory study uses $\CausalPilotStates$ pre-update states from
$\CausalPilotSeeds$ new plain K-FAC trajectories under the Stage B protocol
(seeds 42, 43 and 44; zero-based steps 20, 60 and 200). Training applies
$\mathbf{d}_0$; corrections are evaluated only as diagnostics. Comparisons share saved
weights and the augmented minibatch, with BatchNorm in training mode. These
states are separate from the corrected training runs used for criterion~(b).

For $\boldsymbol{\delta}=\mathbf{d}_1-\mathbf{d}_0$, the undamped model's endpoint slope
$(\mathbf{g}+\mathbf{H}\mathbf{d}_0)^\top\boldsymbol{\delta}$ has the opposite sign to the measured slope
$\nabla L(\boldsymbol{\theta}+\mathbf{d}_0)^\top\boldsymbol{\delta}$ in $\CausalPilotSlopeSignFlips$ states.
This identifies a directional discrepancy at the endpoint.

With $\mathbf{d}_0$, $\mathbf{U}$, $\lambda$ and the batch fixed, the same reduced solve is
evaluated after replacing the predicted endpoint gradient by the measured
one, replacing $\mathbf{U}^\top \mathbf{H}\mathbf{U}$ by its endpoint value, or making both replacements.
The residual retains $\lambda \mathbf{U}^\top \mathbf{d}_0$, and the model gate uses the initial
damped model. Controls use $\mathbf{d}_0+\alpha\boldsymbol{\delta}$ for
$\alpha\in\{0,1/32,1/16,1/8,1/4,1/2,1\}$. Gradient refresh exceeds the best
sampled scaling in $\CausalPilotGradientScaleWins/\CausalPilotStates$ states.
At matched correction norms (the smaller of the two), the original loss
minus the gradient-refresh loss is $+0.03456$, $+0.02934$, $+0.004792$
and $+0.04330$ for (seed, step) $(42,60)$, $(43,60)$, $(43,200)$ and
$(44,60)$, respectively: all four favour refresh.
Neither curvature-only nor joint refresh exceeds the best sampled scaling.
Finite preferences within the closed band
$[-\CausalPilotGainFloor,\CausalPilotGainFloor]$ are reported as ties; wins and
sign reversals must exceed this band. The original norm, model and loss gates
remain unchanged.

A second comparison fixes either the ReLU masks or the max-pool winners at
$\boldsymbol{\theta}$, leaving the candidates unchanged. A sham control uses the same
instrumented forward path while preserving native rectifier and pooling
decisions; its candidate and sampled-ray losses agree with native losses
within absolute $10^{-6}$. At $\boldsymbol{\theta}$, loss and full parameter-gradient
differences satisfy
$\|\mathrm{difference}\|\le10^{-6}+10^{-4}\|\mathrm{reference}\|$.
Selected-AD Hessian-vector products along $\mathbf{d}_0$, $\boldsymbol{\delta}$ and a seeded random
vector satisfy the same test with relative tolerance $10^{-3}$. These checks
cover three directions, rather than the entire Hessian or classical
derivatives at every branch boundary.

Fixing ReLU masks changes the finite preference in $\CausalPilotReluFlips$
states; fixing pooling changes it in $\CausalPilotPoolFlips$. In
$\CausalPilotReluBothLossesRiseFlips$ of the ReLU reversals, both surrogate
losses increase, with the larger increase at the baseline
(Table~\ref{tab:causal-pilot}). The finite ranking can differ despite agreement
of the checked initial AD derivatives. These changes concern
surrogate losses rather than improvements in the native objective.


\begin{table}[htbp]
\centering
\setlength{\tabcolsep}{4pt}
\begin{tabular}{rrrrrrr}
\toprule
Seed & Step & Native gain & Grad. extra & ReLU gain & $\Delta L_0$ & $\Delta L_1$ \\
\midrule
42 & 20 & $0.9519$ & $-0.5778$ & $0.8288$ & $0.4195$ & $0.5427$ \\
42 & 60 & $0.0525$ & $0.0126$ & $1.2375$ & $1.1427$ & $-0.0423$ \\
42 & 200 & $-0.0603$ & $-0.0002$ & $-0.0320$ & $0.0191$ & $-0.0092$ \\
43 & 20 & $0.9767$ & $-0.4824$ & $0.6760$ & $0.6352$ & $0.9360$ \\
43 & 60 & $-0.3046$ & $0.0178$ & $2.6351$ & $6.0726$ & $3.1329$ \\
43 & 200 & $-0.0730$ & $0.0041$ & $0.1259$ & $0.1814$ & $-0.0175$ \\
44 & 20 & $0.6035$ & $-0.2593$ & $0.7989$ & $0.6128$ & $0.4174$ \\
44 & 60 & $-0.0926$ & $0.0360$ & $1.7213$ & $1.9390$ & $0.1251$ \\
44 & 200 & $-0.0381$ & $-0.0022$ & $-0.0178$ & $0.0129$ & $-0.0073$ \\
\bottomrule
\end{tabular}
\caption{Selected frozen-state one-step refresh and surrogate comparisons; every gain is $L(\boldsymbol{\theta}+\mathbf{d}_0)-L(\boldsymbol{\theta}+\mathbf{d})$ within the same loss function. Native gain uses the raw correction, not the accepted fallback. Grad. extra is the gradient-refresh native gain minus the best gain on the registered discrete scale grid. ReLU gain uses the fixed-ReLU surrogate. The last two columns are surrogate minus native losses at the baseline ($\Delta L_0$) and raw corrected ($\Delta L_1$) steps; positive values mean worse losses, not a native-loss improvement.}
\label{tab:causal-pilot}
\end{table}

Table~\ref{tab:causal-pilot} shows selected outcomes from the original records.
The public archive~\citep{bolshim2026curvblocksoftware} provides all nine states in
\path{data/reporting/causal_pilot/results.json}: each row reports the refresh
candidates and scale grid under \texttt{native}, and the fixed-ReLU and
fixed-pooling results under \texttt{interventions}. The norm-matched direction
comparison is recorded in the author-held primary measurements, not in that
public extract.

\section{Saved-state residual-input comparison}
\label{app:residual-input}

This comparison uses the same nine retained ResNet-50/Cars captures as
\ref{app:causal-pilot}: training seeds 42, 43 and 44 at steps 20,
60 and 200 on three plain K-FAC trajectories. It uses the saved weights,
buffers, module modes, prepared 64-image minibatch, runtime settings and base
displacement $\mathbf{d}_0$, without new training or reconstruction of K-FAC factors.
The gradient $\mathbf{g}$ and selected-AD parameter Hessian $\mathbf{H}$ come from the same
current loss graph; the stored gradient is a reference check, not a replacement
for $\mathbf{g}$. The captures are shared, but the residual-input interventions
differ from those in \ref{app:causal-pilot}; neither comparison replays
the corrected Stage B trajectories or their historical refresh proposals.

\paragraph{Residual inputs and common geometry}
Let $\mathbf{D}_{\mathbf{H}}=\operatorname{bd}_6(\mathbf{H})$ for the groups stem, layer1--layer4 and
head. These are measurement groups, not the module blocks of K-FAC. Define
\begin{equation}\label{eq:residual-input-split}
  \begin{aligned}
    \mathbf{r}_{\rm full}&=\mathbf{g}+\mathbf{H}\mathbf{d}_0+\lambda \mathbf{d}_0,\\
    \mathbf{r}_{\rm off}&=(\mathbf{H}-\mathbf{D}_{\mathbf{H}})\mathbf{d}_0,\\
    \mathbf{r}_{\rm rest}&=\mathbf{g}+\mathbf{D}_{\mathbf{H}}\mathbf{d}_0+\lambda \mathbf{d}_0.
  \end{aligned}
\end{equation}
Thus $\mathbf{r}_{\rm full}=\mathbf{r}_{\rm off}+\mathbf{r}_{\rm rest}$ without assuming that $\mathbf{d}_0$
solves an exact block-Hessian system. The remaining residual includes
within-group interactions and baseline operator/damping differences; it is
not the Fisher information retained by K-FAC. All six masked-source
Hessian-vector products enter the split, regardless of the selected pairs.

Within each state and geometry draw, the three corrections share the same
factored $\mathbf{U}$, reduced eigensystem and coefficient metric,
\begin{equation}\label{eq:residual-input-solve}
  \mathbf{A}=\operatorname{sym}(\mathbf{U}^\top \mathbf{H}\mathbf{U}),\qquad
  \mathbf{M}=|\mathbf{A}|+\lambda \mathbf{I},\qquad
  \boldsymbol{\delta}_j=-\mathbf{U}\mathbf{M}^{-1}\mathbf{U}^\top \mathbf{r}_j.
\end{equation}
The protocol fixes $\lambda=1$, ten coupling probes, threshold $0.1$, pair
rank four, eight power iterations, at most three pairs and norm-ratio gate
two. Primary coupling/basis seeds are training seed plus 100003/200003;
sensitivity seeds are plus 300003/400003. Before scaling or gates, numerical
checks compare the masked-source sum with $\mathbf{H}\mathbf{d}_0$, the residual and correction
sums with their full counterparts, and the full proposal with the native
correction. Component-scaled tolerances account for cancellation; native full
arithmetic is not replaced by a component sum. Component corrections need
not lower the full quadratic, and their finite loss changes need not add.

\paragraph{Endpoints and controls}
For each arm, the primary endpoint is the raw native-loss difference
\begin{equation}\label{eq:residual-input-loss}
  \Delta L_j(\alpha)=L(\boldsymbol{\theta}+\mathbf{d}_0+\alpha\boldsymbol{\delta}_j)-L(\boldsymbol{\theta}+\mathbf{d}_0)
\end{equation}
at $\alpha=1$, before any gate; negative means improvement. The gains in
Table~\ref{tab:causal-pilot} use the opposite sign. The complete
fixed scale grid is $0$, $1/32$, $1/16$, $1/8$, $1/4$, $1/2$ and $1$. Every finite candidate is
measured even if its norm or model gate rejects it. Applied outcomes are
separate: a valid fallback has zero applied gain, while unavailable technical
outcomes remain missing with a reason. Native equality acceptance is not a
strict loss improvement. For each full/off, full/rest and off/rest pair, both
corrections are scaled down to their smaller original norm; neither is
increased. A correction below $10^{-3}\|\mathbf{d}_0\|$, or exactly zero, cannot
establish a directional advantage in this control. Grid minima include zero
and are retrospective diagnostics, not an online line search or training
result.

For the retrospective standalone-versus-combined comparison, define
$x=\Delta L_{\rm off}(\alpha)$,
$y=\Delta L_{\rm full}(\alpha)-\Delta L_{\rm rest}(\alpha)$, and
$J=y-x$. The common baseline was checked across arms. At unit scale $J<0$
in all nine states in each draw. Its primary range is approximately
$[-5.62764,-0.0634367]$ and its sensitivity range is
$[-8.40099,-0.0417573]$, calculated from unrounded losses. Across the full
grid all 108 nonzero-scale values of $J$ are negative as recorded, while all
18 zero-scale contrasts are observed zeros. Some small-scale values are only
a few $10^{-6}$; strict signs do not imply 108 numerically resolved
interactions, and zero observed repeat spread is not a bound on total error.
There are 126 common-scale contrasts. Of the 54 norm-matched comparisons
(three pairs per state and draw), 18 compare full with rest; none defines a common-scale $J$.
Even in an exact quadratic with exactly additive corrections, $J$ can be
negative; its sign does not establish nonquadraticity or a new mechanism.

\paragraph{Numerical interpretation}
Functional loss trials preserve parameters, buffers, modes and random state.
Losses are evaluated three times; a separate preliminary run on the same
seed-42, step-60 capture also tests reverse candidate order. The operational
preference band uses the largest within-candidate repeat or order discrepancy
in that run: $\max(10^{-4},10\times\text{same-candidate spread})$, not from
arm differences. Raw values and strict gates are unchanged inside the band;
unresolved preference does not mean zero effect. Reference and independent
flat-coordinate AD checks use absolute tolerance $10^{-6}$ and relative
tolerances $10^{-4}$ for gradients and $10^{-3}$ for Hessian-vector products.
These checks do not certify exact derivatives at branch boundaries or total
floating-point error. The evaluations use native float32 arithmetic and saved backend settings.

The action diagnostic is $\eta_{\rm action}=\|\mathbf{U}^\top \mathbf{r}_{\rm off}\|^2/
\|\mathbf{r}_{\rm off}\|^2$ only after the factored basis passes the fixed $10^{-4}$
native screen: the maximum absolute entry of float32
$\operatorname{project}(\operatorname{lift}(\mathbf{I}))-\mathbf{I}$. Its coordinate numerator, denominator and rejection
reason are retained separately. A rejected basis yields no energy fraction;
coefficient damping $\lambda \mathbf{I}$ then must not be identified with the reduced
ambient damping $\lambda \mathbf{U}^\top \mathbf{U}$. The shared native linear-map comparison
still applies, but an orthonormal-basis guarantee is not certified. Zero or
numerically unresolved action denominators are distinguished from a resolved
zero projection. Even when available, this diagnostic concerns summed
off-group action on $\mathbf{d}_0$, not Frobenius coverage, individual-pair coverage
or subspace-selection utility.

Table~\ref{tab:residual-geometry} reports all 18 geometries. The native
max-entry errors range from $\BasisNativeErrorMin$ to $\BasisNativeErrorMax$;
eight pass the screen. Both draws at seed 42, step 60 fail it; their directly
evaluated losses do not depend on passing this screen. Separately, float64
products of the saved float32
factors representing $\mathbf{U}=\mathbf{C} \mathbf{R}_{\mathbf{U}}$ form $\mathbf{G}=\mathbf{R}_{\mathbf{U}}^\top(\mathbf{C}^\top \mathbf{C})\mathbf{R}_{\mathbf{U}}$ without a
dense ambient $\mathbf{U}$. Across draws, represented Gram eigenvalues range from
$\BasisGramEigenMin$ to $\BasisGramEigenMax$ and the largest per-draw
condition number is $\BasisGramConditionMax$. This represented-factor
calculation is not the native screen or a recomputation of recorded losses.

For a compact map sensitivity, let the saved reduced eigenfactors be
$\mathbf{V},\boldsymbol{\Lambda}$, set $\mathbf{A}_s=\mathbf{V}\boldsymbol{\Lambda} \mathbf{V}^\top$ and
$\mathbf{P}_s=\mathbf{V}\operatorname{diag}((|\Lambda_i|+\lambda)^{-1})\mathbf{V}^\top$ with
$\lambda=1$. Define $\mathbf{Q}=\mathbf{U}\mathbf{G}^{-1/2}$,
$\mathbf{B}=\mathbf{G}^{-1/2}\mathbf{A}_s\mathbf{G}^{-1/2}$, $\mathbf{K}_s=-\mathbf{U}\mathbf{P}_s\mathbf{U}^\top$ and
$\mathbf{K}_o=-\mathbf{Q}(|\mathbf{B}|+\lambda \mathbf{I})^{-1}\mathbf{Q}^\top$. The largest
$\rho=\|\mathbf{K}_s-\mathbf{K}_o\|_2/\|\mathbf{K}_s\|_2$ over the 18 geometries is
$\BasisMapDifferencePercentMax$ per cent. This compares maps represented by
the saved factors, not the full native float32 implementation or HVPs; it is
neither a bound on every individual relative correction error nor a
certificate of finite-loss ordering.

\begin{table*}[htbp]
  \centering
  \caption{Native float32 project/lift max-entry screen and separate float64
    represented-factor geometry for all registered state/draw cases (P/S).
    Gram extrema and condition describe the saved factors; the last column is
    the relative spectral difference between saved and whitened represented maps,
    in per cent, not an error bound on native arithmetic or loss.}
  \label{tab:residual-geometry}
  \scriptsize
  \setlength{\tabcolsep}{2pt}
  \begin{tabular}{@{}rrcllllrrrr@{}}
    \toprule
    Seed & Step & Draw & $k$/raw & Native error & Screen & $\|\mathbf{G}-\mathbf{I}\|_2$ & $\lambda_{\min}(\mathbf{G})$ & $\lambda_{\max}(\mathbf{G})$ & $\kappa(\mathbf{G})$ & Map diff. (\%) \\
    \midrule
    42 & 20 & P & 28/30 & $2.34{\cdot}10^{-5}$ & pass & $2.37{\cdot}10^{-5}$ & 1.000000 & 1.000024 & 1.000024 & 0.0010 \\ 
    42 & 20 & S & 28/30 & $3.61{\cdot}10^{-5}$ & pass & $4.01{\cdot}10^{-5}$ & 1.000000 & 1.000040 & 1.000040 & 0.0018 \\ 
    42 & 60 & P & 28/30 & $2.47{\cdot}10^{-4}$ & fail & $2.79{\cdot}10^{-4}$ & 1.000000 & 1.000279 & 1.000279 & 0.0148 \\ 
    42 & 60 & S & 28/30 & $2.67{\cdot}10^{-4}$ & fail & $2.69{\cdot}10^{-4}$ & 1.000000 & 1.000269 & 1.000269 & 0.0153 \\ 
    42 & 200 & P & 28/30 & $6.22{\cdot}10^{-4}$ & fail & $6.23{\cdot}10^{-4}$ & 1.000000 & 1.000623 & 1.000622 & 0.0248 \\ 
    42 & 200 & S & 29/30 & $1.47{\cdot}10^{-4}$ & fail & $1.78{\cdot}10^{-4}$ & 1.000000 & 1.000178 & 1.000179 & 0.0086 \\ 
    43 & 20 & P & 28/30 & $1.30{\cdot}10^{-5}$ & pass & $1.68{\cdot}10^{-5}$ & 1.000000 & 1.000017 & 1.000017 & 0.0002 \\ 
    43 & 20 & S & 28/30 & $1.51{\cdot}10^{-5}$ & pass & $1.67{\cdot}10^{-5}$ & 1.000000 & 1.000017 & 1.000017 & 0.0003 \\ 
    43 & 60 & P & 28/30 & $1.28{\cdot}10^{-4}$ & fail & $1.30{\cdot}10^{-4}$ & 0.999996 & 1.000130 & 1.000134 & 0.0071 \\ 
    43 & 60 & S & 28/30 & $2.68{\cdot}10^{-4}$ & fail & $2.69{\cdot}10^{-4}$ & 1.000000 & 1.000269 & 1.000269 & 0.0157 \\ 
    43 & 200 & P & 28/30 & $4.08{\cdot}10^{-4}$ & fail & $4.73{\cdot}10^{-4}$ & 0.999876 & 1.000473 & 1.000598 & 0.0232 \\ 
    43 & 200 & S & 28/30 & $3.88{\cdot}10^{-4}$ & fail & $3.89{\cdot}10^{-4}$ & 1.000000 & 1.000389 & 1.000389 & 0.0210 \\ 
    44 & 20 & P & 28/30 & $1.96{\cdot}10^{-5}$ & pass & $2.00{\cdot}10^{-5}$ & 1.000000 & 1.000020 & 1.000020 & 0.0007 \\ 
    44 & 20 & S & 28/30 & $2.75{\cdot}10^{-5}$ & pass & $3.27{\cdot}10^{-5}$ & 0.999981 & 1.000033 & 1.000052 & 0.0012 \\ 
    44 & 60 & P & 28/30 & $7.28{\cdot}10^{-5}$ & pass & $7.46{\cdot}10^{-5}$ & 1.000000 & 1.000075 & 1.000075 & 0.0027 \\ 
    44 & 60 & S & 28/30 & $2.39{\cdot}10^{-4}$ & fail & $2.47{\cdot}10^{-4}$ & 1.000000 & 1.000247 & 1.000247 & 0.0131 \\ 
    44 & 200 & P & 28/30 & $5.17{\cdot}10^{-5}$ & pass & $5.20{\cdot}10^{-5}$ & 0.999991 & 1.000052 & 1.000061 & 0.0029 \\ 
    44 & 200 & S & 28/30 & $2.26{\cdot}10^{-4}$ & fail & $3.93{\cdot}10^{-4}$ & 0.999885 & 1.000393 & 1.000507 & 0.0177 \\ 
    \bottomrule
  \end{tabular}
\end{table*}

All nine states were evaluated on the same fixed scale grid with the same
implementation and calibration.
The design has 486 candidate slots and 54 raw $\alpha=1$ endpoints. Both geometry draws are reported for each state. Checkpoints within a
trajectory are dependent; the second draw tests sensitivity to the subspace,
not to the training seed. No population
confidence interval or $p$-value is inferred from these rows.

\paragraph{Complete outcomes and controls}
All $\AblationValid/\AblationExpected$ planned candidate loss evaluations
returned finite values, with no technical failures or unattempted cases.
Table~\ref{tab:residual-ablation}
reports all nine states and both draws; Table~\ref{tab:residual-controls}
reports the sampled minima and all pairwise norm controls. The 63 raw
unresolved slots comprise 54 exact alpha-zero ties and nine nonzero values
inside the preference band.
Native acceptance includes equality and subfloor improvements. At unit scale, the strict recorded event
$\Delta L_{\rm off}>0$ and $\Delta L_{\rm full}<\Delta L_{\rm rest}$ occurs in nine
state/draw comparisons; in four, the safeguard returns both full and rest
to the same base step. At seed 42, step 60, full passes the safeguard in
both draws, whereas off and rest do not.

Sampled grid minima improve beyond the band for full/off/rest in $5/9$,
$1/9$, $8/9$ primary state grids and $6/9$, $2/9$, $7/9$ sensitivity grids.
Two negative sensitivity-rest minima, approximately $-4.08\times10^{-5}$
and $-5.41\times10^{-5}$, remain unresolved. Among the eight geometries that pass the basis screen,
the action-energy fraction ranges from 0.0865 to 0.5099. An off-only
candidate can still raise the loss in these cases.

\begin{table*}[htbp]
  \centering
  \caption{Residual-input ablation at $\alpha=1$, all nine states and both
    registered geometry draws (P/S). Raw $\Delta L$ precedes every gate; applied
    $\Delta L$ is separate and is zero only for an actual finite fallback/tie.
    Negative is improvement. A/R denotes native acceptance/rejection; U denotes
    the frozen operational preference band. Missing values are never zeros.
    $\eta_{\rm action}$ is the checked projection of summed off action, not
    Frobenius coverage; nonorthog. denotes failure of the native orthonormality screen,
    and undefined reasons remain distinct. These are descriptive
    within-seed observations, not independent draws or numerical certificates.}
  \label{tab:residual-ablation}
  \scriptsize
  \setlength{\tabcolsep}{3pt}
  \begin{tabular}{@{}rrclllll@{}}
    \toprule
    Seed & Step & Draw & Endpoint & Full $\Delta L$ & Off $\Delta L$ & Rest $\Delta L$ & $\eta_{\rm action}$ \\
    \midrule
    42 & 20 & P & raw & $-9.52{\cdot}10^{-1}$ (A) & $6.89{\cdot}10^{-3}$ (R) & $-8.95{\cdot}10^{-1}$ (A) & $1.21{\cdot}10^{-1}$ \\ 
     &  &  & applied & $-9.52{\cdot}10^{-1}$ & $0$ (U) & $-8.95{\cdot}10^{-1}$ &  \\
    42 & 20 & S & raw & $-9.35{\cdot}10^{-1}$ (A) & $1.58{\cdot}10^{-2}$ (R) & $-8.83{\cdot}10^{-1}$ (A) & $1.25{\cdot}10^{-1}$ \\ 
     &  &  & applied & $-9.35{\cdot}10^{-1}$ & $0$ (U) & $-8.83{\cdot}10^{-1}$ &  \\
    42 & 60 & P & raw & $-5.25{\cdot}10^{-2}$ (A) & $2.12{\cdot}10^{0}$ (R) & $8.94{\cdot}10^{-2}$ (R) & nonorthog. \\ 
     &  &  & applied & $-5.25{\cdot}10^{-2}$ & $0$ (U) & $0$ (U) &  \\
    42 & 60 & S & raw & $-4.30{\cdot}10^{-2}$ (A) & $2.26{\cdot}10^{0}$ (R) & $1.35{\cdot}10^{-1}$ (R) & nonorthog. \\ 
     &  &  & applied & $-4.30{\cdot}10^{-2}$ & $0$ (U) & $0$ (U) &  \\
    42 & 200 & P & raw & $6.03{\cdot}10^{-2}$ (R) & $1.12{\cdot}10^{0}$ (R) & $2.46{\cdot}10^{-1}$ (R) & nonorthog. \\ 
     &  &  & applied & $0$ (U) & $0$ (U) & $0$ (U) &  \\
    42 & 200 & S & raw & $5.95{\cdot}10^{-2}$ (R) & $1.47{\cdot}10^{0}$ (R) & $5.26{\cdot}10^{-1}$ (R) & nonorthog. \\ 
     &  &  & applied & $0$ (U) & $0$ (U) & $0$ (U) &  \\
    43 & 20 & P & raw & $-9.77{\cdot}10^{-1}$ (A) & $6.43{\cdot}10^{-2}$ (R) & $-9.66{\cdot}10^{-1}$ (A) & $8.65{\cdot}10^{-2}$ \\ 
     &  &  & applied & $-9.77{\cdot}10^{-1}$ & $0$ (U) & $-9.66{\cdot}10^{-1}$ &  \\
    43 & 20 & S & raw & $-1.02{\cdot}10^{0}$ (A) & $-4.06{\cdot}10^{-2}$ (A) & $-9.36{\cdot}10^{-1}$ (A) & $1.24{\cdot}10^{-1}$ \\ 
     &  &  & applied & $-1.02{\cdot}10^{0}$ & $-4.06{\cdot}10^{-2}$ & $-9.36{\cdot}10^{-1}$ &  \\
    43 & 60 & P & raw & $3.05{\cdot}10^{-1}$ (R) & $9.68{\cdot}10^{-1}$ (R) & $-1.93{\cdot}10^{-1}$ (A) & nonorthog. \\ 
     &  &  & applied & $0$ (U) & $0$ (U) & $-1.93{\cdot}10^{-1}$ &  \\
    43 & 60 & S & raw & $3.12{\cdot}10^{-1}$ (R) & $8.63{\cdot}10^{-1}$ (R) & $-2.01{\cdot}10^{-1}$ (A) & nonorthog. \\ 
     &  &  & applied & $0$ (U) & $0$ (U) & $-2.01{\cdot}10^{-1}$ &  \\
    43 & 200 & P & raw & $7.30{\cdot}10^{-2}$ (R) & $3.85{\cdot}10^{0}$ (R) & $1.85{\cdot}10^{0}$ (R) & nonorthog. \\ 
     &  &  & applied & $0$ (U) & $0$ (U) & $0$ (U) &  \\
    43 & 200 & S & raw & $6.33{\cdot}10^{-2}$ (R) & $5.58{\cdot}10^{0}$ (R) & $2.88{\cdot}10^{0}$ (R) & nonorthog. \\ 
     &  &  & applied & $0$ (U) & $0$ (U) & $0$ (U) &  \\
    44 & 20 & P & raw & $-6.04{\cdot}10^{-1}$ (A) & $1.49{\cdot}10^{-1}$ (R) & $-6.86{\cdot}10^{-1}$ (A) & $1.30{\cdot}10^{-1}$ \\ 
     &  &  & applied & $-6.04{\cdot}10^{-1}$ & $0$ (U) & $-6.86{\cdot}10^{-1}$ &  \\
    44 & 20 & S & raw & $-6.15{\cdot}10^{-1}$ (A) & $1.44{\cdot}10^{-1}$ (R) & $-6.93{\cdot}10^{-1}$ (A) & $1.46{\cdot}10^{-1}$ \\ 
     &  &  & applied & $-6.15{\cdot}10^{-1}$ & $0$ (U) & $-6.93{\cdot}10^{-1}$ &  \\
    44 & 60 & P & raw & $9.26{\cdot}10^{-2}$ (R) & $1.64{\cdot}10^{0}$ (R) & $-3.23{\cdot}10^{-1}$ (A) & $3.49{\cdot}10^{-1}$ \\ 
     &  &  & applied & $0$ (U) & $0$ (U) & $-3.23{\cdot}10^{-1}$ &  \\
    44 & 60 & S & raw & $7.69{\cdot}10^{-2}$ (R) & $1.53{\cdot}10^{0}$ (R) & $-3.55{\cdot}10^{-1}$ (A) & nonorthog. \\ 
     &  &  & applied & $0$ (U) & $0$ (U) & $-3.55{\cdot}10^{-1}$ &  \\
    44 & 200 & P & raw & $3.81{\cdot}10^{-2}$ (R) & $1.42{\cdot}10^{-1}$ (R) & $3.35{\cdot}10^{-2}$ (R) & $5.10{\cdot}10^{-1}$ \\ 
     &  &  & applied & $0$ (U) & $0$ (U) & $0$ (U) &  \\
    44 & 200 & S & raw & $4.99{\cdot}10^{-2}$ (R) & $2.13{\cdot}10^{-1}$ (R) & $4.37{\cdot}10^{-2}$ (R) & nonorthog. \\ 
     &  &  & applied & $0$ (U) & $0$ (U) & $0$ (U) &  \\
    \bottomrule
  \end{tabular}
\end{table*}

\begin{table*}[htbp]
  \centering
  \caption{Scale and pairwise down-only norm diagnostics in fixed geometry.
    Minima are taken retrospectively over the seven-point grid, including zero,
    rather than by an online line search. Pair columns give left-minus-right
    raw $\Delta L$ at matched correction norms.}
  \label{tab:residual-controls}
  \scriptsize
  \setlength{\tabcolsep}{2pt}
  \begin{tabular}{@{}rrcrrrrrr@{}}
    \toprule
    Seed & Step & Draw & min full & min off & min rest & full--off & full--rest & off--rest \\
    \midrule
    42 & 20 & P & $-9.52{\cdot}10^{-1}$ & $-8.59{\cdot}10^{-3}$ & $-8.95{\cdot}10^{-1}$ & $-5.95{\cdot}10^{-1}$ & $2.57{\cdot}10^{-2}$ & $6.21{\cdot}10^{-1}$ \\
    42 & 20 & S & $-9.35{\cdot}10^{-1}$ & $-3.89{\cdot}10^{-3}$ & $-8.83{\cdot}10^{-1}$ & $-5.97{\cdot}10^{-1}$ & $2.41{\cdot}10^{-2}$ & $6.22{\cdot}10^{-1}$ \\
    42 & 60 & P & $-8.95{\cdot}10^{-2}$ & $0$ & $-1.59{\cdot}10^{-1}$ & $-6.74{\cdot}10^{-1}$ & $1.06{\cdot}10^{-1}$ & $2.07{\cdot}10^{0}$ \\
    42 & 60 & S & $-7.97{\cdot}10^{-2}$ & $0$ & $-1.43{\cdot}10^{-1}$ & $-6.36{\cdot}10^{-1}$ & $9.83{\cdot}10^{-2}$ & $2.17{\cdot}10^{0}$ \\
    42 & 200 & P & $0$ & $0$ & $-8.46{\cdot}10^{-4}$ & $4.45{\cdot}10^{-2}$ & $5.92{\cdot}10^{-2}$ & $8.90{\cdot}10^{-1}$ \\
    42 & 200 & S & $0$ & $0$ & $-4.08{\cdot}10^{-5}$ & $5.08{\cdot}10^{-2}$ & $5.62{\cdot}10^{-2}$ & $9.94{\cdot}10^{-1}$ \\
    43 & 20 & P & $-9.77{\cdot}10^{-1}$ & $0$ & $-9.66{\cdot}10^{-1}$ & $-6.11{\cdot}10^{-1}$ & $5.01{\cdot}10^{-3}$ & $6.23{\cdot}10^{-1}$ \\
    43 & 20 & S & $-1.02{\cdot}10^{0}$ & $-4.06{\cdot}10^{-2}$ & $-9.36{\cdot}10^{-1}$ & $-5.99{\cdot}10^{-1}$ & $-4.64{\cdot}10^{-2}$ & $5.72{\cdot}10^{-1}$ \\
    43 & 60 & P & $0$ & $0$ & $-1.93{\cdot}10^{-1}$ & $-4.33{\cdot}10^{-1}$ & $4.20{\cdot}10^{-1}$ & $8.22{\cdot}10^{-1}$ \\
    43 & 60 & S & $0$ & $0$ & $-2.01{\cdot}10^{-1}$ & $-3.84{\cdot}10^{-1}$ & $4.08{\cdot}10^{-1}$ & $7.61{\cdot}10^{-1}$ \\
    43 & 200 & P & $0$ & $0$ & $-2.18{\cdot}10^{-3}$ & $-2.45{\cdot}10^{-1}$ & $-1.47{\cdot}10^{-2}$ & $1.69{\cdot}10^{0}$ \\
    43 & 200 & S & $-6.87{\cdot}10^{-4}$ & $0$ & $-4.96{\cdot}10^{-3}$ & $-3.44{\cdot}10^{-1}$ & $-1.59{\cdot}10^{-2}$ & $2.83{\cdot}10^{0}$ \\
    44 & 20 & P & $-6.04{\cdot}10^{-1}$ & $0$ & $-6.86{\cdot}10^{-1}$ & $-6.34{\cdot}10^{-1}$ & $1.81{\cdot}10^{-1}$ & $8.07{\cdot}10^{-1}$ \\
    44 & 20 & S & $-6.15{\cdot}10^{-1}$ & $0$ & $-6.93{\cdot}10^{-1}$ & $-6.33{\cdot}10^{-1}$ & $1.71{\cdot}10^{-1}$ & $7.91{\cdot}10^{-1}$ \\
    44 & 60 & P & $-2.48{\cdot}10^{-2}$ & $0$ & $-3.23{\cdot}10^{-1}$ & $-7.70{\cdot}10^{-1}$ & $3.74{\cdot}10^{-1}$ & $1.86{\cdot}10^{0}$ \\
    44 & 60 & S & $-3.41{\cdot}10^{-2}$ & $0$ & $-3.55{\cdot}10^{-1}$ & $-6.42{\cdot}10^{-1}$ & $3.67{\cdot}10^{-1}$ & $1.77{\cdot}10^{0}$ \\
    44 & 200 & P & $0$ & $0$ & $0$ & $2.34{\cdot}10^{-2}$ & $3.22{\cdot}10^{-2}$ & $5.20{\cdot}10^{-2}$ \\
    44 & 200 & S & $0$ & $0$ & $-5.41{\cdot}10^{-5}$ & $3.38{\cdot}10^{-2}$ & $4.80{\cdot}10^{-2}$ & $1.71{\cdot}10^{-1}$ \\
    \bottomrule
  \end{tabular}
\end{table*}
\section{Saved-state crossing and disjoint-input check}
\label{app:crossing-check}

This study assesses conditional Monte Carlo uncertainty in the stage/layer
contrast at six saved checkpoints on two fixed disjoint inputs.
Table~\ref{tab:crossing-check} reports all twelve cells, including unresolved
threshold comparisons. The metrics specified for this follow-up are separate
from the four original pre-specified criteria in \ref{app:criteria-verdicts}.

\paragraph{States, inputs and arithmetic}
The six epoch-300 checkpoints are ResNet-18 and Plain-18 at training seeds
42, 43 and 44 from the long schedule. Checkpoint hashes in the primary records identify the states used here,
not their historical measurement-time identity. The public archive~\citep{bolshim2026curvblocksoftware}
lists the states under \texttt{protocol.checkpoints} in
\path{data/reporting/exp8_crossing_check/results.json}.
B0 consists of offline CIFAR-100 test rows 0--31, matching the historical
selection inferred from source; B1 consists of rows 32--63. Both use channel
means $(0.5071,0.4865,0.4409)$ and standard deviations
$(0.2673,0.2564,0.2762)$, without augmentation or download, in evaluation mode
with mean cross-entropy. Within each replicate, the partitions share the same
checkpoint and input but use separate PRNG streams.

Parameters, floating buffers, inputs, forward/HVP calculations and component
reductions use float32; targets and BatchNorm counters use int64.
Deterministic algorithms and cuDNN deterministic mode are enabled, while
autocast, matmul/cuDNN TF32 and cuDNN benchmarking are disabled. The cuBLAS
workspace setting is \texttt{:4096:8}. Scalar pooling and resampling use CPU
float64. The resulting inference is conditional on this finite-precision
procedure, without a total floating-point/AD/HVP error bound or an
exact-Hessian spectral certificate. Internal chunk and out-of-memory
adaptation is enabled, but retry counts and chunk history were not recorded.
Probe budgets count successful observations rather than all attempted kernels.

\paragraph{Estimator and fixed budget}
Each seed/model/input cell has thirty paired replicates, each with 100
successful Rademacher probes per diagonal squared component and per full-norm
component. Both partitions estimate the denominator from separate full-Hessian
probes. This estimator has the same ideal target as the original stage
all-block reconstruction, but a different Monte Carlo construction.
Its conditional intervals therefore
do not serve as error bars for the historical scalars in
Table~\ref{tab:two-partitions}.

For partition $p$, let $d_{prg}$ estimate $\|\mathbf{H}_{gg}\|_F^2$ and $f_{pr}$ estimate
$\|\mathbf{H}\|_F^2$ in replicate $r$. The squared components are pooled as
$\bar d_{pg}=30^{-1}\sum_r d_{prg}$ and $\bar f_p=30^{-1}\sum_r f_{pr}$ before
forming
\begin{equation}\label{eq:crossing-pooling}
  A_p=\Bigl(\sum_g\sqrt{\bar d_{pg}}\Bigr)^2,\qquad
  S_p=\frac{A_p}{\bar f_p},\qquad
  \Phi_p=1-\frac{\sum_g\bar d_{pg}}{\bar f_p},\qquad
  \Phi_{\mathrm{sep},p}=1-\frac{\sum_g\bar d_{pg}}{A_p}.
\end{equation}
The paired contrast is $\Delta S=S_{\mathrm{stage}}-S_{\mathrm{layer}}$.
Pooling precedes the nonlinear transformation, rather than averaging thirty
replicate ratios. The stage partition has six groups; the layer partitions
have 22 for ResNet-18 and 19 for Plain-18, each covering the trainable
parameters once. Probe replicates describe Monte Carlo variation conditional
on three training seeds and two fixed inputs.

\paragraph{Conditional uncertainty and joint screen}
The analysis uses 10000 bootstrap resamples from PCG64 seed 20261020.
Within each cell, a resample draws thirty indices with replacement and keeps
the complete paired stage/layer vector, including all diagonal and full-norm
components. Cells use independent index arrays, and
\eqref{eq:crossing-pooling} is recomputed after resampled pooling. Marginal
95 per cent intervals use the 2.5 and 97.5 percentiles, with linear
interpolation at order position $(n-1)p$.

For the 36 prespecified metrics---two $S$ values and $\Delta S$ per cell---let
$\hat\theta_j$ denote the observed estimate and $s_j$ the fixed bootstrap
sample SD (\texttt{ddof=1}). The joint screen is
\begin{equation}\label{eq:crossing-screen}
  T_b^*=\max_{j=1,\ldots,36}
    \frac{|\hat\theta_{bj}^*-\hat\theta_j|}{s_j},\qquad
  I_j=\bigl[\hat\theta_j-c_{.95}s_j,\ \hat\theta_j+c_{.95}s_j\bigr],
\end{equation}
where $c_{.95}=\CrossCriticalValue$ is the linearly interpolated 95th percentile
of $T_b^*$. This gives an approximate 95 per cent family screen for conditional
Monte Carlo variation, distinct from the marginal percentile intervals and
the original across-seed intervals. A non-finite or non-regular pooled domain,
or a non-positive denominator, makes the affected marginal unavailable.
Any invalid family member or zero/non-finite family SD makes the entire joint
screen unavailable; draws are not deleted, clipped or replaced. All observed
and resampled domains were regular. Delete-one paired jackknife bias and
standard-error estimates are diagnostic and are not used for bias correction.

\begin{table}[htbp]
  \centering
  \caption{Saved-state crossing check on the two fixed disjoint CIFAR-100
    test inputs (B0: indices 0--31; B1: 32--63), with all twelve cells retained.
    Endpoints are the approximate 95 per cent conditional family screen over all 36
    prespecified $S_{\mathrm{stage}}$, $S_{\mathrm{layer}}$ and $\Delta S$
    metrics, not pointwise percentile intervals or across-seed intervals.
    Resolved means unrounded stage lower endpoint $>1$ and layer upper endpoint
    $<1$; Reverse means the opposite exclusions; otherwise Unresolved.
    Positive $\Delta S$ alone does not resolve a crossing.}
  \label{tab:crossing-check}
  \scriptsize
  \setlength{\tabcolsep}{3pt}
  \begin{tabular}{@{}lrc c c c c l@{}}
    \toprule
    Model & Seed & Input & $S_{\mathrm{stage}}$ & Family endpoints & $S_{\mathrm{layer}}$ & Family endpoints & Screen \\
    \midrule
    ResNet-18 & 42 & B0 & $1.1251$ & $[1.1072,\,1.1431]$ & $0.8394$ & $[0.8269,\,0.8519]$ & Resolved \\
    ResNet-18 & 42 & B1 & $1.0566$ & $[1.0417,\,1.0715]$ & $0.7037$ & $[0.6916,\,0.7158]$ & Resolved \\
    Plain-18 & 42 & B0 & $1.1771$ & $[1.1515,\,1.2026]$ & $0.8468$ & $[0.8260,\,0.8675]$ & Resolved \\
    Plain-18 & 42 & B1 & $1.1145$ & $[1.0433,\,1.1857]$ & $1.0113$ & $[0.9585,\,1.0641]$ & Unresolved \\
    ResNet-18 & 43 & B0 & $1.0988$ & $[1.0666,\,1.1309]$ & $0.9097$ & $[0.8890,\,0.9304]$ & Resolved \\
    ResNet-18 & 43 & B1 & $1.0537$ & $[1.0114,\,1.0961]$ & $0.8534$ & $[0.8232,\,0.8837]$ & Resolved \\
    Plain-18 & 43 & B0 & $1.1601$ & $[1.1118,\,1.2085]$ & $0.9286$ & $[0.8877,\,0.9696]$ & Resolved \\
    Plain-18 & 43 & B1 & $1.1396$ & $[1.0958,\,1.1835]$ & $1.0161$ & $[0.9824,\,1.0498]$ & Unresolved \\
    ResNet-18 & 44 & B0 & $1.1477$ & $[1.1273,\,1.1681]$ & $0.9021$ & $[0.8881,\,0.9162]$ & Resolved \\
    ResNet-18 & 44 & B1 & $1.0873$ & $[1.0557,\,1.1188]$ & $0.9758$ & $[0.9533,\,0.9984]$ & Resolved \\
    Plain-18 & 44 & B0 & $1.1908$ & $[1.1277,\,1.2539]$ & $1.0041$ & $[0.9603,\,1.0480]$ & Unresolved \\
    Plain-18 & 44 & B1 & $1.1238$ & $[1.0550,\,1.1925]$ & $0.9746$ & $[0.9030,\,1.0463]$ & Unresolved \\
    \bottomrule
  \end{tabular}
\end{table}

\paragraph{Threshold classifications}
All \CrossResnetResolved/\CrossCellsPerModel\ ResNet-18 cells have an unrounded
stage lower endpoint above one and layer upper endpoint below one. The
seed-44/B1 layer upper endpoint, \CrossNarrowLayerUpper, is close to the
threshold. Plain-18 has \CrossPlainResolved\ resolved crossings, at seeds
42 and 43 on B0, and \CrossPlainUnresolved\ unresolved comparisons: seed 44/B0
and all three B1 cells have layer intervals containing one. Every $\Delta S$
is positive, including in the four unresolved cells, but a
positive contrast alone does not resolve the two threshold exclusions.

Marginal $S$ half-widths range from \CrossSHalfWidthLo\ to \CrossSHalfWidthHi;
\CrossSGoalsUnmet/24 exceed the descriptive goal of 0.03, all in Plain-18.
Counting $\Delta S$ gives \CrossFamilyGoalsUnmet/36, including ResNet-18 seed 43/B1.
Neither width nor classification changed the probe budget. In
\path{data/tables/crossing-summary.jsonl}, each cell's \texttt{observation}
records marginal \path{half_width} under \path{partitions.stage.S},
\path{partitions.layer.S} and \path{DeltaS}; Table~\ref{tab:crossing-check}
uses joint endpoints from \path{screen_95}.